%% file: FSAF.tex
\documentclass{article}

\PassOptionsToPackage{numbers, compress}{natbib}

\usepackage[preprint]{neurips_2021}
\usepackage{xcolor}

\usepackage[utf8]{inputenc} 
\usepackage[T1]{fontenc}    
\usepackage{hyperref}       
\usepackage{url}            
\usepackage{booktabs}       
\usepackage{amsfonts}       
\usepackage{nicefrac}       
\usepackage{microtype}      
\usepackage{enumitem}       
\usepackage[normalem]{ulem}
\usepackage{graphicx}
\usepackage{amsmath,amssymb}
\usepackage{mathtools}
\usepackage{ntheorem}
\usepackage{comment}
\usepackage{algorithm}
\usepackage{algpseudocode}
\usepackage{setspace}
\usepackage{graphicx}
\usepackage{subfigure}
\usepackage{amssymb, amsmath, graphicx, latexsym, amsfonts}
\usepackage{enumitem}
\usepackage{ragged2e}
\usepackage{mathtools}
\usepackage{tabu}
\usepackage{epstopdf}
\usepackage{multirow}
\usepackage{verbatim}
\usepackage{color}
\usepackage{lipsum}
\usepackage{mdframed}
\usepackage{bm}
\usepackage{siunitx}
\usepackage{xpatch}
\usepackage{listings}
\usepackage{wrapfig}
\usepackage{multibib}
\newcites{app}{Additional References for the Appendix}

\DeclareMathOperator*{\argmin}{argmin} 
\DeclareMathOperator*{\argmax}{argmax} 

\newtheorem{remark}{Remark}

\DeclareMathAlphabet{\mathpzc}{OT1}{pzc}{m}{it}
\DeclarePairedDelimiterX{\bkt}[1]{(}{)}{ #1}
\DeclarePairedDelimiterX{\sbkt}[1]{[}{]}{ #1}
\DeclarePairedDelimiterX{\lbkt}[1]{\{}{\}}{ #1}
\DeclareMathOperator{\E}{\mathbb{E}}
\DeclareMathOperator{\X}{\mathbb{X}}
\DeclareMathOperator{\V}{\mathbb{V}}

\DeclareMathOperator{\cL}{\mathcal{L}}
\DeclareMathOperator{\cM}{\mathcal{M}}
\DeclareMathOperator{\cD}{\mathcal{D}}
\DeclareMathOperator{\cT}{\mathcal{T}}

\DeclareMathOperator{\cS}{\mathcal{S}}
\DeclareMathOperator{\cA}{\mathcal{A}}

\DeclareMathOperator{\cB}{\mathcal{B}}
\DeclareMathOperator{\cF}{\mathcal{F}}
\DeclareMathOperator{\cR}{\mathcal{R}}

\DeclarePairedDelimiterX{\infdivx}[2]{(}{)}{%
  #1\;\delimsize\|\;#2%
}
\newcommand{\kldiv}{D_{\text{KL}}\infdivx}
\DeclarePairedDelimiter{\norm}{\lVert}{\rVert}
\newtheorem*{proof-non}{Proof}

\newcolumntype{x}[1]{>{\raggedright\arraybackslash}p{#1}}
\newmdtheoremenv[linewidth=0.8pt, skipabove=6pt, skipbelow=6pt]{theorem_md}{Theorem}
\newmdtheoremenv[linewidth=0.8pt, skipabove=6pt, skipbelow=6pt]{lemma_md}{Lemma}
\newmdtheoremenv[linewidth=0.8pt, skipabove=6pt, skipbelow=6pt]{proposition_md}{Proposition}


\title{Reinforced Few-Shot Acquisition Function Learning for Bayesian Optimization}
  \author{%
  \textsuperscript{1}Bing-Jing Hsieh, \textsuperscript{1}Ping-Chun Hsieh, \textsuperscript{2}Xi Liu\\
  
  \textsuperscript{1}Department of Computer Science, National Chiao Tung University, Hsinchu, Taiwan\\
  \textsuperscript{2}Applied Machine Learning, Facebook AI, Menlo Park, CA, USA\\
  \{bingjing2000.cs08g,pinghsieh\}@nctu.edu.tw, xliu1@fb.com  
  
}

\begin{document}

\maketitle

\begin{abstract}
Bayesian optimization (BO) conventionally relies on handcrafted acquisition functions (AFs) to sequentially determine the sample points. However, it has been widely observed in practice that the best-performing AF in terms of regret can vary significantly under different types of black-box functions. It has remained a challenge to design one AF that can attain the best performance over a wide variety of black-box functions. This paper aims to attack this challenge through the perspective of reinforced few-shot AF learning (FSAF). Specifically, we first connect the notion of AFs with Q-functions and view a deep Q-network (DQN) as a surrogate differentiable AF. While it serves as a natural idea to combine DQN and an existing few-shot learning method, we identify that such a direct combination does not perform well due to severe overfitting, which is particularly critical in BO due to the need of a versatile sampling policy. To address this, we present a Bayesian variant of DQN with the following three features: (i) It learns a distribution of Q-networks as AFs based on the Kullback-Leibler regularization framework. This inherently provides the uncertainty required in sampling for BO and mitigates overfitting. (ii) For the prior of the Bayesian DQN, we propose to use a demo policy induced by an off-the-shelf AF for better training stability. (iii) On the meta-level, we leverage the meta-loss of Bayesian model-agnostic meta-learning, which serves as a natural companion to the proposed FSAF. Moreover, with the proper design of the Q-networks, FSAF is general-purpose in that it is agnostic to the dimension and the cardinality of the input domain. Through extensive experiments, we demonstrate that the FSAF achieves comparable or better regrets than the state-of-the-art benchmarks on a wide variety of synthetic and real-world test functions.
\end{abstract}

\input{1-intro}
\input{2-preliminaries}
\input{4-algorithm}
\input{5-experiments}
\input{6-related}
\input{7-conclusion}
{
\small
\bibliographystyle{unsrtnat}
\bibliography{reference}
}

\input{8-Appendix}

\end{document}

%% file: 1-intro.tex
\section{Introduction}
\label{section:intro}
Bayesian optimization (BO) has served as a powerful and popular framework for global optimization in many real-world tasks, such as hyperparameter tuning \cite{Snoek2012practical,wu2019hyperparameter,thornton2013auto,zhang2015improving}, robot control \cite{calandra2016bayesian}, automatic material design \cite{frazier2015bayesian,griffiths2020constrained,shahriari2015taking}, etc. 
To search for global optima under a small sampling budget and potentially noisy observations, BO imposes a Gaussian process (GP) prior on the unknown black-box function and continually updates the posterior as more samples are collected. BO relies on \textit{acquisition functions} (AFs) to determine the sample location, i.e., those with larger AF values are prioritized than those with smaller ones. AFs are often designed to capture the trade-off between exploration and exploitation of the global optima. The design of AFs has been extensively studied from various perspectives, such as optimism in the face of uncertainty (e.g., GP-UCB \cite{srinivas2010gaussian}), optimizing information-theoretic metrics (e.g., entropy search methods \cite{hennig2012entropy,hernandez2014predictive,wang2017max}), and maximizing one-step improvement (e.g., expected improvement or EI \cite{movckus1975bayesian,jones1998efficient}). 
As a result, AFs are often handcrafted according to different perspectives of the trade-off, and the best-performing AFs can vary significantly under different types of black-box functions \cite{hernandez2014predictive}.
This phenomenon is repeatedly observed in our experiments in Section \ref{section:exp}.
Therefore, one critical issue in BO is to design an AF that can adapt to a variety of black-box functions.

To achieve better adaptability to new tasks in BO, recent works propose to leverage \textit{meta-data}, the data previously collected from similar tasks \cite{wistuba2021fewshot,wistuba2018scalable,feurer2018scalable}.
For example, in the context of hyperparameter optimization under a specific dataset, the meta-data could come from the evaluation of previous hyperparameter configurations for the same learning model over any other related dataset.
In \cite{wistuba2021fewshot}, meta-data is used to fine-tune the initialization of the GP parameters and thereby achieves better GP model selection for each specific task. 
However, the potential benefit of using meta-data for more efficient exploration via AFs is not explored.
On the other hand, in \cite{wistuba2018scalable,feurer2018scalable}, meta-data is split into multiple subsets and then used to construct a transferable acquisition function based on some off-the-shelf AF (e.g. EI) and an ensemble of GP models. Each of the GP models is learned over a separate subset of the meta-data.
However, to achieve effective knowledge transfer, this approach would require a sufficiently large amount of meta-data, which significantly limits its practical use.
As a result, there remains a critical unexplored challenge in BO: \textit{how to design an AF that can effectively adapt to a wide variety of back-box functions given only a small amount of meta-data?}

\vspace{-2mm}
\begin{figure*}[!htbp]
\centering 
  \begin{minipage}{\textwidth}
    \centering 
    \subfigure[]{
    \label{Fig:Training_curve_1particle}
    \includegraphics[width=0.24\textwidth]{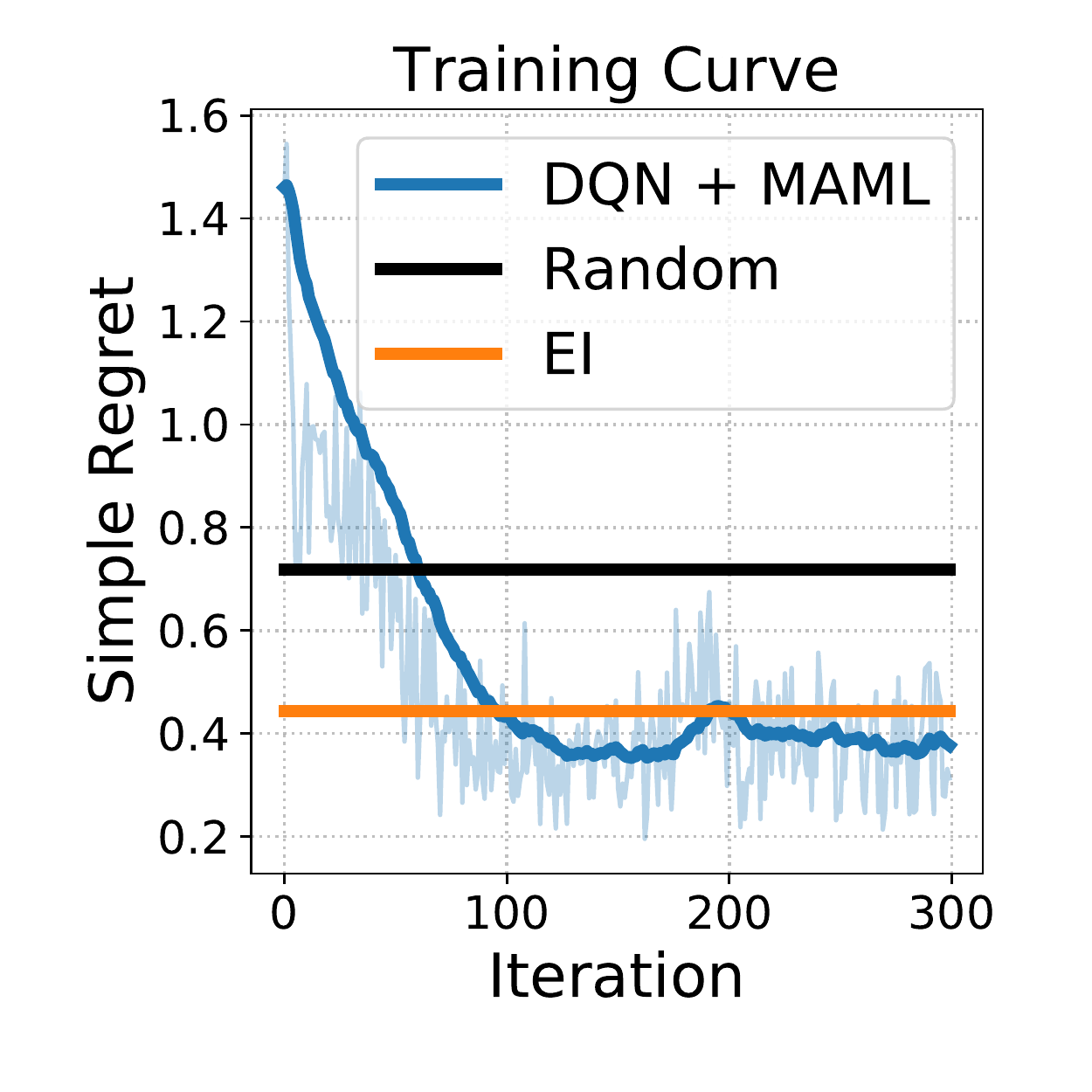}}
    \hspace{-5mm}
    \subfigure[]{
    \label{Fig:1particle_EI.0}
    \includegraphics[width=0.24\textwidth]{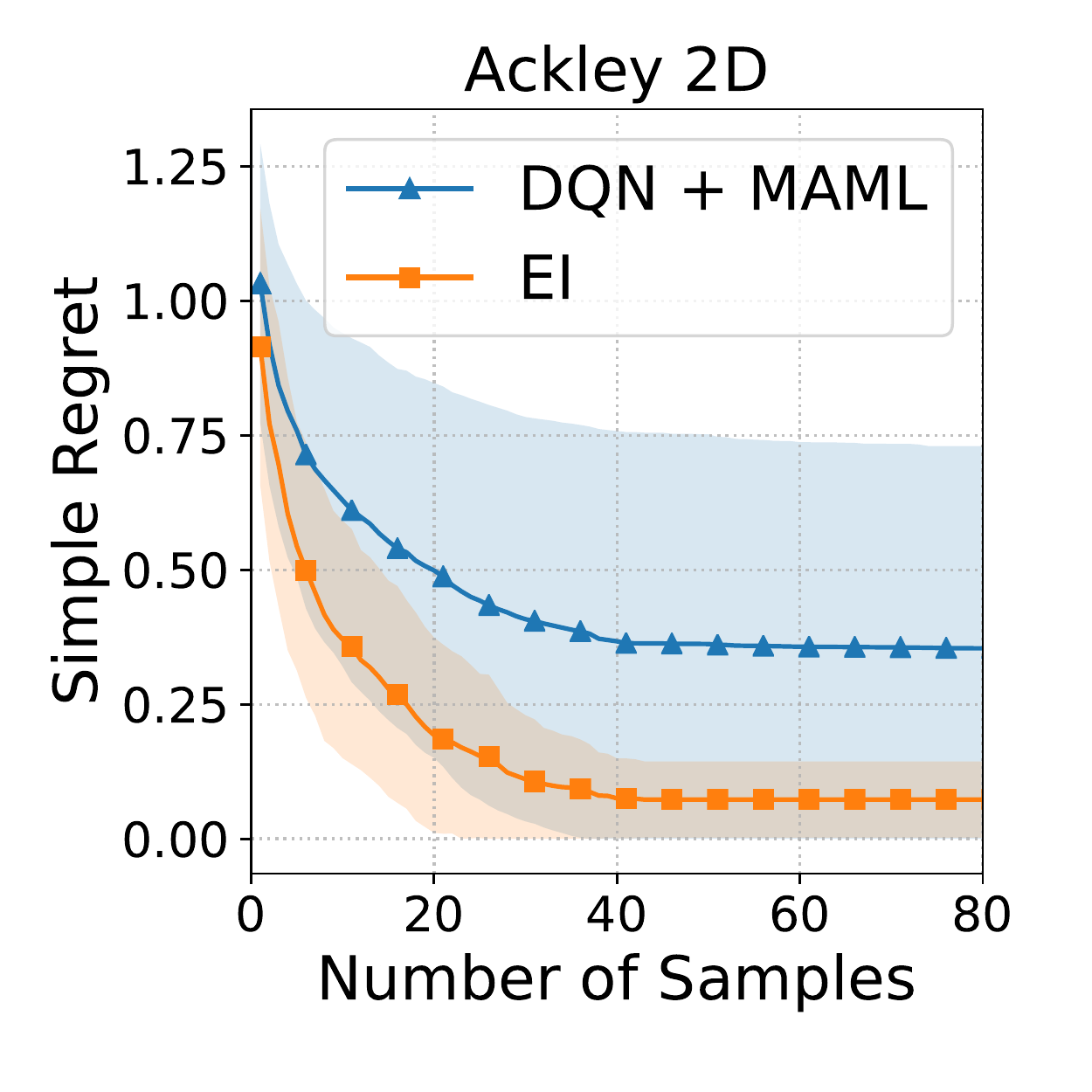}}
    \hspace{-5mm}
    \subfigure[]{
    \label{Fig:1particle_EI.1}
    \includegraphics[width=0.24\textwidth]{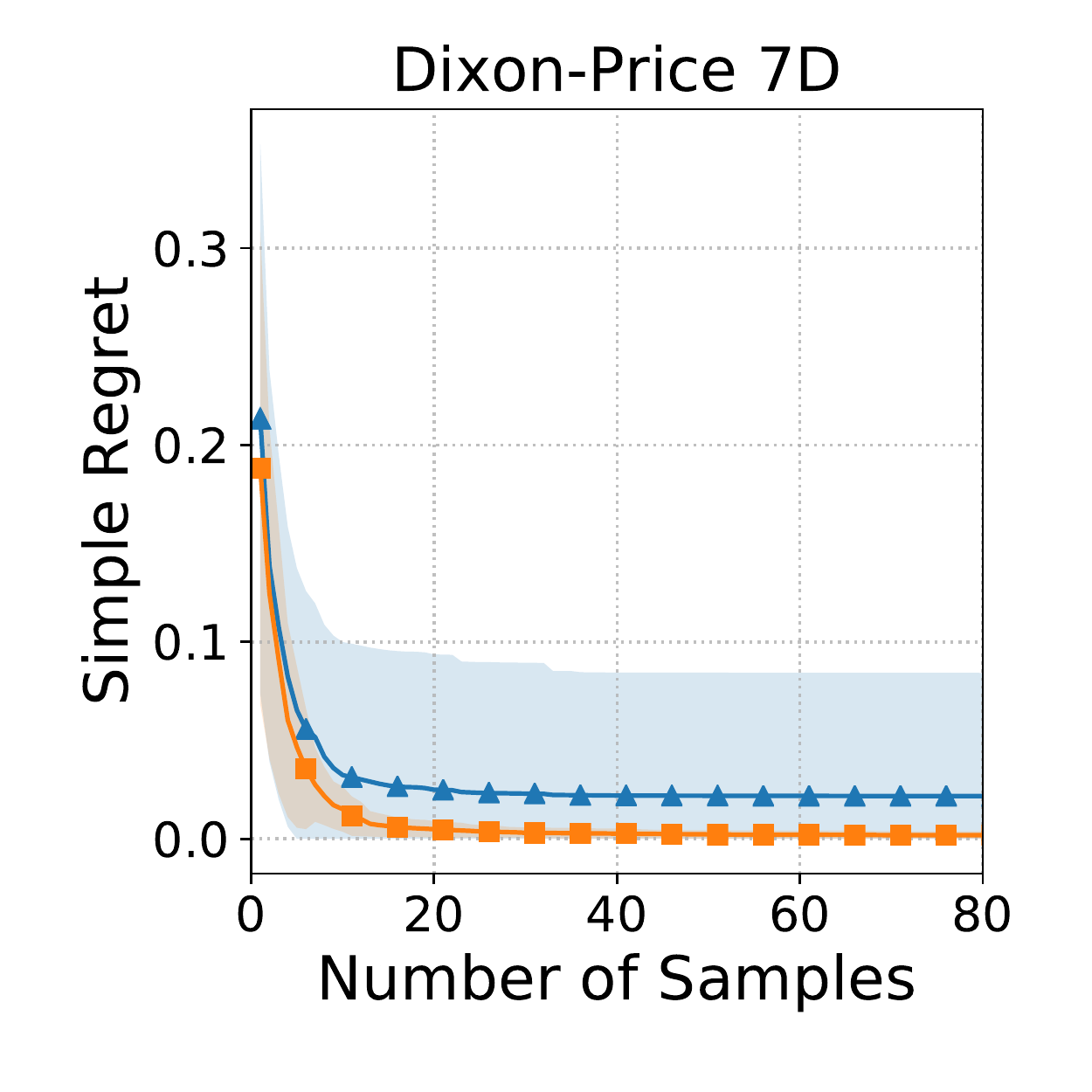}}
    \hspace{-5mm}
    \subfigure[]{
    \label{Fig:1particle_EI.2}
    \includegraphics[width=0.24\textwidth]{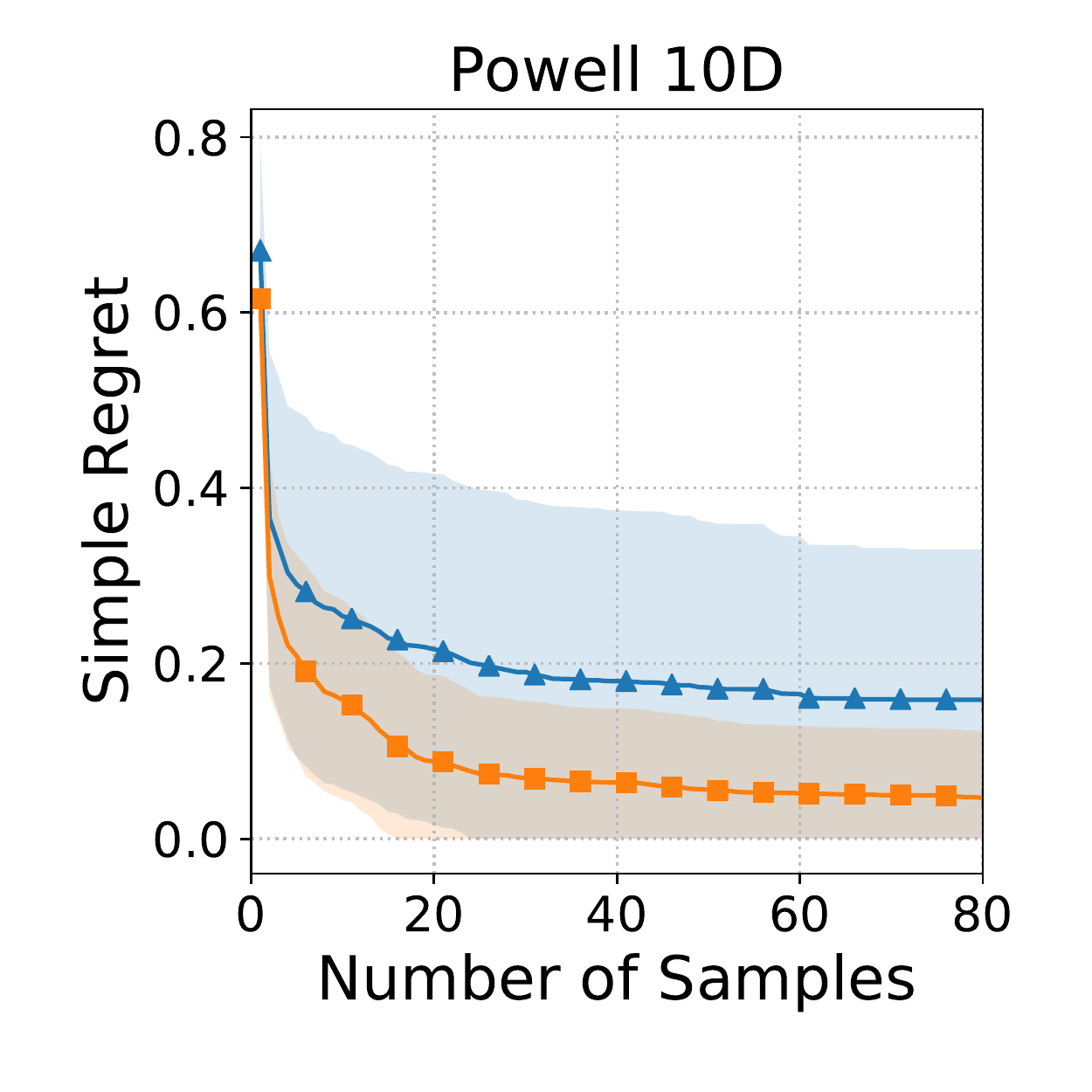}}
    \vspace{-3mm}
    \caption{An illustration of the overfitting issue of DQN+MAML trained with GP functions: (a) Training curves of DQN+MAML (As EI and Random do not require any training, their lines are flat); (b)-(d) Average simple regrets of DQN+MAML and EI in testing under benchmark functions.}
    \label{Fig:overfitting}
  \end{minipage}
\end{figure*}





To tackle this challenge, we propose to rethink the use of meta-data in BO through the lens of \textit{few-shot acquisition function (FSAF) learning}.
Specifically, our goal is to learn an initial AF model that allows few-shot fast adaptation to each specific task during evaluation.
Inspired by the similarity between AFs and Q-functions, we use a deep Q-network (DQN) as a surrogate differentiable AF, i.e., the Q-network would output an indicator for each candidate sample point given its posterior mean and variance as well as other related information.
Given the parametric nature and the differentiability of a Q-network, it is natural to leverage optimization-based few-shot learning approaches, such as model-agnostic meta-learning (MAML) \cite{finn2017model}, for the training of few-shot AFs.
Despite this natural connection, we find that a direct combination of standard DQN and MAML (DQN+MAML) is prone to overfitting, as illustrated by the comparison of training and testing results.
Note that in Figure \ref{Fig:overfitting} (with detailed configuration in Appendix \ref{app:exp details}), DQN+MAML achieves better regret than EI on the training set but suffers from much higher regret during testing. We hypothesize this is because both DQN and MAML are prone to overfitting \cite{mishra2018simple,yoon2018bayesian,fu2019diagnosing,zintgraf2019fast}.
This issue could be particularly critical in BO due to the uncertainty required in adaptive sampling. 
Based on our findings and inspired by \cite{liu2017stein}, we propose a Bayesian variant of DQN with the following three salient features: (i) The Bayesian DQN learns a distribution of parameters of DQN based on the Kullback-Leibler (KL) regularization framework. Through this, the problem of minimizing the Q-learning temporal-difference (TD) error in DQN is converted into a Bayesian inference problem. (ii) For the prior of the Bayesian DQN, we propose to use a demonstration policy induced by an off-the-shelf AF to further stabilize the training process;
(iii) We then use the chaser meta-loss in \cite{yoon2018bayesian}, which serves as a natural companion to the proposed Bayesian DQN.
As shown by the experimental results in Section \ref{section:exp}, the proposed design effectively mitigates overfitting and achieves good generalization under various black-box functions.
Moreover, with the proper design of the Q-networks, the proposed FSAF is general-purpose in the sense that it is agnostic to both the dimension and the cardinality of the input domain.

The main contributions of this paper can be summarized as follows:
\vspace{-2mm}
\begin{itemize}[leftmargin=*]
    \item We consider a novel setting of few-shot acquisition function learning for BO and present the first few-shot acquisition function that can use a small amount of meta-data to achieve better task-specific exploration and thereby effectively adapt to a wide variety of black-box functions.
    \vspace{-1mm}
    \item 
    Inspired by the similarity between AFs and Q-functions, we view DQN as a parametric and differentiable AF and use it as the base of our FSAF.
    We identify the important overfitting issue in the direct combination of DQN and MAML and thereafter present a Bayesian variant of DQN that mitigates overfitting and enjoys stable training through a demo-based prior.
    \vspace{-1mm}
    \item We extensively evaluate the proposed FSAF in a variety of tasks, including optimization benchmark functions, real-world datasets, and synthetic GP functions. We show that the proposed FSAF can indeed effectively adapt to a variety of tasks and outperform both the conventional benchmark AFs as well as the recent state-of-the-art meta-learning BO methods.
\end{itemize}

%% file: 2-preliminaries.tex
\vspace{-3mm}
\section{Preliminaries}
\label{section:prelim}
\vspace{-1mm}

Our goal is to design a sampling policy to optimize a black-box function $f:\X \rightarrow \mathbb{R}$, where $\X\subset \mathbb{R}^d$ denotes the compact domain of $f$.
$f$ is black-box in the sense that there are no special structural properties (e.g., concavity or linearity) or derivatives (e.g., gradient or Hessian) about $f$ available to the sampling policy.
In each step $t$, the policy selects $x_t\in \X$ and obtains a noisy observation $y_t=f(x_t)+\varepsilon_t$, where $\varepsilon_t$ are i.i.d. zero-mean Gaussian noises.
To evaluate a sampling policy, we define the \textit{simple regret} as
$\text{Regret}(t):=\max_{x\in \X}f(x^*)-\max_{1\leq s \leq t}f(x_s)$, which quantifies the best sample up to $t$.
For convenience, we let $\cF_t:=\{(x_i,y_i)\}_{i=1}^{t-1}$ denote the observations up to step $t$.

\textbf{Bayesian optimization.} To optimize $f$ in a sample-efficient manner, BO first imposes on the space of objective functions a GP prior, which is fully characterized by a mean function and a covariance function, and then determines the next sample based on the resulting posterior \cite{frazier2018tutorial,brochu2010tutorial}.
In each step $t$, given $\cF_t$, the posterior predictive distribution of each $x\in\X$ is $\mathcal{N}(\mu_t(x),\sigma^2_t(x))$, where $\mu_t(x):=\E[f(x)\rvert \cF_t]$ and $\sigma_t(x):=(\V[f(x)\rvert \cF_t])^{\frac{1}{2}}$ can be derived in closed form.
In this way, BO can be viewed as a sequential decision making problem.
However, it is typically difficult to obtain the exact optimal policy due to the curse of dimensionality \cite{frazier2018tutorial}.
To obtain tractable policies, BO algorithms construct AFs $\Psi(x; \cF_t)$, which resort to maximizing one-step look-ahead objectives based on $\cF_t$ and the posterior \cite{frazier2018tutorial}. 
For example, EI chooses the sample location based on the improvement made by the immediate next sample in expectation, i.e., 
$\Psi_{\text{EI}}(x;\cF_t)=\E[f(x) - \max_{1\leq i\leq t-1}f(x_i)\rvert \cF_t]$, which enjoys a closed form in $\mu_t(x)$, $\sigma_t(x)$, and $\max_{1\leq i\leq t-1}f(x_i)$.

\textbf{Meta-learning with few-shot fast adaptation.}
Meta-learning is a generic paradigm for generalizing the knowledge acquired during training to solving unseen tasks in the testing phase. 
In the few-shot setting, meta-learning is typically achieved via a bi-level framework: (i) On the upper level, the training algorithm is meant to determine a proper initial model with an aim to facilitating subsequent task-specific adaptation;
(ii) On the lower level, given the initial model and the task of interest, a fast adaptation subroutine is configured to fine-tune the initial model based on a small amount of task-specific data.
Specifically, during training, the learner finds a model parameterized by $\theta$ based on a collection of tasks $\cT$, where each task $\tau\in\cT$ is associated with a training set $\cD^{\text{tr}}_\tau$ and a validation set $\cD^{\text{val}}_\tau$.
For any initial model parameters $\theta$ and training set $\cD^{\text{tr}}$ of a task $\tau$, let $\cM(\theta, \cD^{\text{tr}}_{\tau})$ be an algorithm that outputs the adapted model parameters by applying few-shot fast adaptation to $\theta$ based on $\cD^{\text{tr}}_\tau$.
The performance of the adapted model is evaluated on $\cD^{\text{val}}_\tau$ by a meta-loss function $\cL(\cM(\theta, \cD^{\text{tr}}_\tau), \cD^{\text{val}}_\tau)$.
Accordingly, the overall training can be viewed as solving the following optimization problem:

\begin{equation}
    \theta^*:=\argmin_{\theta}\sum_{\tau\in \cT} \cL(\cM(\theta, \cD_{\tau}^{\text{tr}}), \cD_{\tau}^{\text{val}}).
    \label{eq:meta-learning obj}
\end{equation}

By properly configuring the loss function, the formulation in (\ref{eq:meta-learning obj}) is readily applicable to various learning problems, including supervised learning and RL.
Note that the adaptation subroutine is typically chosen as taking one or a few gradient steps with respect to $\cL(\cdot,\cdot)$ or some relevant loss function.
For example, under the celebrated MAML \cite{finn2017model}, the adaptation subroutine is $\cM(\theta,\cD_{\tau}^{\text{tr}})\equiv \theta - \eta \nabla_{\theta}\cL(\theta,\cD_{\tau}^{\text{tr}})$, where $\eta$ denotes the learning rate.

\textbf{Reinforcement learning and Q-function.}
\label{section:prelim:RL}
Following the conventions of RL, we use $s_t$, $a_t$, and $r_t$ to denote the state, action, and reward obtained at each step $t$. Let $R$ and $\gamma$ be the reward function and the discount factor.
The goal is to find a stationary randomized policy that maximizes the total expected discounted reward $\E[\sum_{t=0}^{\infty}\gamma^t r_t]$.
To achieve this, given a policy $\pi$, a helper function termed Q-function is defined as $Q^{\pi}(s,a):=\mathbb{E}[\sum_{t=0}^{\infty}\gamma^t R(s_t,a_t)\rvert s_o=s,a_0=a;\pi]$.
The optimal Q-function can then be defined as $Q^*(s,a):=\max_{\pi}Q^{\pi}(s,a)$, for each state-action pair.
One fundamental property of the optimal Q-function is the \textit{Bellman optimality equation}, i.e., $Q^*(s,a)=\E\big[r_t+\gamma \max_{a'}Q^*(s_{t+1},a')\big\rvert s_t=a,a_t=a\big]$.
Then, the Bellman optimality operator can be defined by $[\cB Q] (s,a):= R(s,a)+\gamma \E_{s'}[\max_{a'\in\mathcal{A}}Q(s',a')]$.
It is known that $Q^*$ is the unique fixed point of $\cB$.
We leverage this fact to describe the proposed FSAF in Section \ref{section:alg:DQN}.

%% file: 4-algorithm.tex
\section{Few-Shot Acquisition Function}
\label{section:alg}


\vspace{-1mm}
\subsection{Deep Q-Network as a Differentiable Parametric Acquisition Function}
\label{section:alg:DQN as AF}
\vspace{-2mm}
%
Based on the conceptual similarity between acquisition functions and Q-functions, in this section we present how to cast a deep Q-network as a parametric and differentiable instance of acquisition function.
To begin with, we consider the Q-network architecture with state and action representations as the input, as typically adopted by Q-learning for large action spaces \cite{van2020q}.

\textbf{State-action representation.} In BO, an action corresponds to choosing one location to sample from the input domain $\X$, and the state at each step $t$ can be fully captured by the collection of sampled points $\{(x_i,y_i)\}_{i=1}^{t-1}$.
However, this raw state representation appears problematic as its dimension depends on the number of observed sample points.
Inspired by the acquisition functions, we leverage the posterior mean and variance as the \textit{joint state-action representation} for each candidate sample location.
In addition, we include the best observation so far (defined as $y_t^*:=\argmax_{1\leq i\leq t-1}y_i$) and the ratio between the current timestamp and total sampling budget $T$, which reflects the sampling progress in BO.
In summary, at each step $t$, the state-action representation of each $x\in\X$ is designed to be a 4-tuple $(\mu_t(x),\sigma_t(x),y_t^*,\frac{t}{T})$, which is agnostic to the dimension and cardinality of $\X$.

\textbf{Reward signal.} To reflect the sampling progress, we define the reward $r_t$ as a function of the simple regret, i.e., $r_t=g(\text{Regret}(t))$, where $g:\mathbb{R}_{+}\rightarrow \mathbb{R}$ is a strictly decreasing function. Practical examples include $g(z)=-z$ and $g(z)=-\log z$. 

%
\begin{remark}
\normalfont One popular approach to construct a representation of fixed dimension is through embedding. From this viewpoint, the posterior mean and variance can be viewed as a natural embedding generated by GP inference in the context of BO.
It is an interesting direction to extend the proposed design by constructing more general state and action representations via embedding techniques.
\end{remark}



\vspace{-1mm}
\subsection{A Bayesian Perspective of Deep Q-Learning for Bayesian Optimization}
\label{section:alg:DQN}
\vspace{-1mm}

One major challenge in designing an acquisition function for BO is to address the wide variability of black-box functions and accordingly achieve a favorable explore-exploit trade-off for each task.
To address this by deep Q-learning as described in Section \ref{section:alg:DQN}, instead of learning a single Q-network as in the standard DQN \cite{mnih2015human}, we propose to learn a distribution of Q-network parameters from a Bayesian inference perspective to achieve more robust exploration and more stable training.
Inspired by \cite{liu2017stein}, we adapt the regularized minimization problem to Q-learning in order to connect Q-learning and Bayesian inference as follows.
Let $C(\theta)$ be the cost function that depends on the model parameter $\theta$.
Instead of finding a single model, the Bayesian approach finds a distribution $q(\theta)$ over $\theta$ that minimizes the cost function augmented with a KL-regularization penalty, i.e.,
\begin{equation}
    \min_{q(\theta)} \Big\{\E_{\theta\sim q(\theta)}[C(\theta)]+\alpha \kldiv{q}{q_0}\Big\},\label{eq:ERM}
\end{equation}
where $q_0$ denotes a prior distribution over $\theta$ and $\alpha$ is a weight factor of the penalty and $\kldiv{\cdot}{\cdot}$ denotes the Kullback-Leibler (KL) divergence between two distributions.
Note that $q(\theta)$ essentially induces a distribution over the sampling policies $\pi_\theta$, and accordingly $q_0$ can be interpreted as constructing a prior over the policies.
By setting the derivative of the objective in (\ref{eq:ERM}) with respect to the measure induced by $q$ to be zero, one can verify that the optimal solution to (\ref{eq:ERM}) is
\begin{equation}
    q^*(\theta)=\frac{1}{Z}\exp\Big(\frac{-C(\theta)}{\alpha}\Big)q_0(\theta),\label{eq:q theta as a posterior}
\end{equation}
where $Z$ is the normalizing factor.
One immediate interpretation of (\ref{eq:q theta as a posterior}) is that $q^*(\theta)$ can be viewed as the posterior distribution under the prior distribution $q_0(\theta)$ and the likelihood $\exp(-C(\theta)/\alpha)$. 
To adapt the KL-regularized minimization framework to value-based RL for BO, the proposed FSAF algorithm is built on the following design principles for $C(\theta)$ and $q_0(\theta)$:
\begin{itemize}[leftmargin=*]
    \item \textbf{Use mean-squared TD error as the cost function:} 
    Recall from Section \ref{section:prelim} that the optimal Q-function is the fixed point of the Bellman optimality backup operation. Hence, we have $\mathcal{B}Q=Q$ if and only if $Q$ is the optimal Q-function. 
    Based on this observation, one principled choice of $C(\theta)$ is the squared TD error under the operator $\cB$, i.e., $\norm{\cB Q - Q}_2^2$.
    Moreover, in practice, DQN typically incorporates a replay buffer $\cR_{Q}$ (termed the Q-replay buffer) as well as a target Q-network to achieve better training stability \cite{mnih2015human}.
    Therefore, we choose the cost function as
    \begin{equation}
        C(\theta)=\E_{(s,a,s',r)\sim\rho}\Big[\Big(\big(r+\gamma \max_{a'\in\cA}Q(s',a';\theta^{-})\big)-Q(s,a;\theta)\Big)^2\Big],\label{eq:C theta}
    \end{equation}
    where $\rho$ denotes the underlying sample distribution of the replay buffer and $\theta^-$ is the parameter of the target Q-network\footnote{In (\ref{eq:C theta}), the cost function depends implicitly on the target network parameterized by $\theta^-$. Despite this, as the target network is updated periodically from $Q(\cdot,\cdot;\theta)$, for notational convenience we do not make explicit the dependence of the cost function on $\theta^-$ in the notation $C(\theta)$.}.
    In practice, the cost $C(\theta)$ is estimated by the empirical average over a mini-batch $\cD$ of samples $(s,a,r,s')$ drawn from the replay buffer, i.e.,
    $C(\theta)\approx \hat{C}(\theta)=\frac{1}{\lvert \cD\rvert}\sum_{(s,a,r,s')\in \cD}((r+\gamma \max_{a'\in\cA}Q(s',a';\theta^{-}))-Q(s,a;\theta))^2$.
    \vspace{1mm}
    \item \textbf{Construct an informative prior with the help of the existing acquisition functions:} In (\ref{eq:ERM}), the KL-penalty with respect to a prior distribution is meant to encode prior domain knowledge as well as provide regularization that prevents the learned parameter from collapsing into a point estimate.
    One commonly-used choice is a uniform prior (i.e., $q(\theta)=c$ for some positive constant $c$), under which the KL-penalty reduces to the negative entropy of $q$.  
    Given that BO is designed to optimize expensive-to-evaluate functions, it is therefore preferred to use a more informative prior for better sample efficiency.
    \indent Based on the above, we propose to construct a prior with the help of a demo policy $\pi_{\text{D}}$ induced by existing popular AFs (e.g., EI or PI), which inherently capture critical information structure of the GP posterior. 
    Define a similarity indicator $\delta(\pi_{\theta},\pi_{\text{D}})$ of $\pi_{\theta}$ and $\pi_{\text{D}}$ as 
    \begin{equation}
        \delta(\pi_\theta, \tilde{\pi}):=\E_{s\sim \rho,a\sim \pi_{\text{D}}(\cdot|s)}\big[\log (\pi_{\theta}(s,a))\big],\label{eq:policy similarity indicator}
    \end{equation}
    where we slightly abuse the notation and let $\rho$ denote the state distribution induced by the replay buffers.
    Since the term $\log (\pi_{\theta}(s,a))$ in (\ref{eq:policy similarity indicator}) is the log-likelihood of that the action of $\pi_\theta$ matches that of $\pi_{\text{D}}$ at a state $s$, $\delta(\pi_\theta, \pi_{\text{D}})$ reflects how similar the two policies are on average (with respect to the state visitation distribution of $\pi_\theta$).
    Accordingly, we propose to design the prior $q_0(\theta)$ to be
    \begin{align}
        q_o(\theta)\propto \exp\big(\delta(\pi_\theta, \pi_{\text{D}})\big).\label{eq:L between pi theta and pi tilde}
    \end{align}
    As it is typically difficult to directly evaluate $\delta(\pi_\theta, {\pi_{\text{D}}})$ in practice, we construct another replay buffer $\cR_{\text{D}}$ (termed the \textit{demo} replay buffer), which stores the state-action pairs produced under the state distribution $\rho$ and the demo policy ${\pi_{\text{D}}}$, and estimate $\delta(\pi_\theta, \pi_{\text{D}})$ by the empirical average over a mini-batch $\cD'$ of state-action pairs drawn from the demo replay buffer, i.e., $\delta(\pi_\theta, \pi_{\text{D}})\approx \hat{\delta}(\pi_\theta, \pi_{\text{D}})=\frac{1}{\lvert \cD'\rvert}\sum_{(s,a)\in \cD'}\log (\pi_{\theta}(s,a))$.
    
    \item \textbf{Update the Q-networks by Stein variational gradient descent:} The solution in (\ref{eq:q theta as a posterior}) is typically intractable to evaluate and hence approximation is required. 
    We leverage the Stein variational gradient descent (SVGD), which is a general-purpose approach for Bayesian inference. Specifically, we build up $N$ instances of Q-networks (also called \textit{particles} in the context of the variational methods) and update the parameters via Bayesian inference.
    Let $\theta^{(n)}$ denote the parameters of the $n$-th Q-network and use $\Theta$ as a shorthand of $\{\theta^{(n)}\}_{n=1}^{N}$.
    Under SVGD \cite{liu2016stein} and the prior described in (\ref{eq:L between pi theta and pi tilde}), the Stein variational gradient of each particle can be derived as
    \begin{equation}
        {g}^{(n)}(\Theta)=\frac{1}{N}\sum_{i=1}^{N} \nabla_{\theta^{(i)}} \Big(\frac{-1}{\alpha}C(\theta^{(i)})+ \delta(\pi_{\theta^{(i)}},\pi_{\text{D}})\Big)k(\theta^{(i)},\theta^{(n)})+\nabla_{\theta^{(i)}}k(\theta^{(i)},\theta^{(n)}),
    \end{equation}
    where $k(\cdot,\cdot)$ is a kernel function.
    As mentioned above, in practice $C(\theta)$ and ${\delta}(\pi_\theta, \pi_{\text{D}})$ are estimated by the corresponding empirical $\hat{C}(\theta)$ and $\hat{\delta}(\pi_\theta, \pi_{\text{D}})$, respectively.
    Let $\hat{g}^{(n)}(\Theta)$ denote the estimated Stein variational gradient based on $\hat{C}(\theta)$ and $\hat{\delta}(\pi_\theta, \pi_{\text{D}})$.
    Accordingly, the parameters of each Q-network are updated iteratively by SVGD as
    \begin{equation}
        \theta^{(n)}\leftarrow \theta^{(n)}+\eta\cdot \hat{g}^{(n)}(\Theta),\label{eq:empirical SVGD}
    \end{equation}
    where $\eta$ is the learning rate. 
    For ease of notation, we use $\cM_{\text{SVGD}}(\Theta, \cD)$ to denote the subroutine that applies one SVGD update of (\ref{eq:empirical SVGD}) to all Q-networks parameterized by $\Theta$ based on some dataset $\cD$. 
    Note that the update scheme in (\ref{eq:empirical SVGD}) serves as a natural candidate for the few-shot adaptation subroutine of the meta-learning framework described in Section \ref{section:prelim}.
    In Section \ref{section:alg:BMAML}, we will put everything together and describe the full meta-learning algorithm of FSAF.
\end{itemize}





\begin{remark}
\normalfont The presented Bayesian DQN bears some high-level resemblance
to the prior works \cite{osband2016generalization,azizzadenesheli2018efficient,osband2016deep,tang2017variational}, which are inspired by the classic principle of Thompson sampling for exploration.
In \cite{azizzadenesheli2018efficient,osband2016deep}, the Q-function is assumed to be linearly parameterized with a Gaussian prior on the parameters such that the posterior can be computed in closed form.
On the other hand, without imposing the linearity assumption, \cite{osband2016deep} approximates the intractable posterior by maintaining an ensemble of Q-networks as a practical heuristic of Thompson sampling to DQN.
Different from \cite{osband2016generalization,azizzadenesheli2018efficient,osband2016deep}, we approach Bayesian DQN through the principled framework of KL regularization for parametric Bayesian inference and solve it via SVGD, without any linearity assumption.
\cite{tang2017variational} starts from an entropy-regularized formulation for Q-learning and assumes that the target Q-value is drawn from a Gaussian model to obtain a tractable posterior from the perspective of variational inference.
By contrast, we do not rely on the Gaussian assumption and directly find the posterior by SVGD, and for more efficient training we consider a prior induced by an acquisition function.
More importantly, the presented Bayesian DQN naturally helps substantiate the few-shot learning framework described in Section \ref{section:prelim} for BO.
\end{remark}

\begin{remark}
\normalfont The regularized formulation in (\ref{eq:ERM}) has been extensively applied in the class of policy-based methods in RL. 
For example, the entropy-regularized policy optimization \cite{ziebart2010modeling} has been applied to enable a ``soft version'' of policy iteration, which gives rise to the popular soft Q-learning \cite{haarnoja2017reinforcement} and soft actor-critic algorithms \cite{haarnoja2018soft}.
Another example is the Stein variational policy gradient method \cite{liu2017stein}, which connects the policy gradient methods with Bayesian inference.
Different from the prior works, we take a different path to connect the value-based RL approach with Bayesian inference.
\end{remark}

\begin{remark}
\normalfont 
The similarity indicator defined in (\ref{eq:policy similarity indicator}) has a similar form as the loss term in some of the classic imitation learning algorithms. For example, given an expert policy $\pi_e$, DAgger \cite{ross2011reduction} is designed to find a policy $\pi'$ that minimizes a surrogate loss $\E_{s}[\ell(s,\pi_{e})]$, where $\ell(\cdot,\cdot)$ is some loss function (e.g., 0-1 loss or hinge loss) that reflects the dissimilarity between $\pi'$ and $\pi_e$.
Despite this high-level resemblance, one fundamental difference between (\ref{eq:policy similarity indicator}) and imitation learning is that the demo policy ${\pi_{\text{D}}}$ is not a true expert in the sense that mimicking the behavior of ${\pi_{\text{D}}}$ is not the ultimate goal of the FSAF learning process.
Instead, the goal of FSAF is to learn a policy that can better adapt to new tasks and thereby outperform the existing AFs in various domains.
The penalty defined in (\ref{eq:L between pi theta and pi tilde}) is only meant to provide some prior domain knowledge to achieve more sample-efficient training. We further validate this design by providing training curves in Section \ref{section:exp}.
\end{remark}


\subsection{Meta-Learning via Bayesian MAML}
\label{section:alg:BMAML}
Based on the Bayesian DQN design in Section \ref{section:alg:DQN}, the natural way to substantiate the meta-learning framework in (\ref{eq:meta-learning obj}) is to leverage the Bayesian variant of MAML \cite{yoon2018bayesian}.
In the context of BO, each task typically corresponds to optimizing some type of black-box functions (e.g., GP functions from an RBF kernel with some lengthscale).
FSAF implements the bi-level framework of (\ref{eq:meta-learning obj}) as follows: (i) On the lower level, for each task $\tau$, FSAF enforces few-shot fast adaptation by taking $K$ steps of SVGD as described in (\ref{eq:empirical SVGD}) and thereafter obtains the fast-adapted parameters denoted by $\Theta_{\tau,K}$; (ii) On the upper level, for each task $\tau$, FSAF computes a meta-loss that reflects the dissimilarity between the approximated posterior induced by $\Theta_{\tau,K}$ and the true posterior distribution in (\ref{eq:L between pi theta and pi tilde}).
As the true posterior is not available, one practical solution is to approximate the true posterior by taking $S$ additional SVGD gradient steps based on $\Theta_{\tau,K}$ and obtaining a surrogate denoted by $\Theta^{*}_{\tau,S}$ \cite{yoon2018bayesian}.
This design can be justified by the fact that $\Theta^{*}_{\tau,S}$ becomes a better approximation for the true posterior as $S$ increases due to the nature of SVGD.  
For any two collections of particles $\Theta'\equiv \{\theta'^{(n)}\}$ and $\Theta''\equiv \{\theta''^{(n)}\}$, define $D(\Theta',\Theta''):=\sum_{n=1}^{N}\lVert \theta'^{(n)}- \theta''^{(n)} \rVert_2^2$ (called \textit{chaser loss} in \cite{yoon2018bayesian}).
Then, for any task $\tau$, the meta-loss of FSAF is computed as
\begin{align}
    \cL_{\text{meta}}(\Theta;{\tau})= D\big(\Theta_{\tau,K}, \text{stopgrad}(\Theta^{*}_{\tau,S})\big),
\end{align}
As will be shown by the experiments in Section \ref{section:exp}, using small $K$ and $S$ empirically leads to favorable performance.
The pseudo code of the training procedure of FSAF is provided in Appendix \ref{app:pseudo code}.

\begin{remark}
\normalfont This paper focuses on value-based methods for training a few-shot acquisition function. Based on the proposed training framework, it is also possible to extend the idea to design an actor-critic counterpart of our FSAF. We believe this is an interesting direction for future work. 
\end{remark}

%% file: 5-experiments.tex
\section{Experimental Results}
\label{section:exp}
\vspace{-1mm}

We demonstrate the effectiveness of FSAF on a wide variety of classes of black-box functions and discuss how FSAF addresses the critical challenges described in Section \ref{section:intro}.
Unless stated otherwise, we report the median of simple regrets over 100 evaluation trials along with the 25\% and 75\% percentiles (shown via shaded areas or tables in Appendix \ref{app:results} for clarity). 

\textbf{Popular benchmark methods.} We evaluate FSAF against various popular benchmark methods, including EI \cite{movckus1975bayesian}, PI \cite{Kushner1964ANM}, GP-UCB \cite{srinivas2010gaussian},
MES \cite{wang2017max}, and MetaBO \cite{volpp2020metalearning}.
The configuration and hyperparameters of the above methods are as follows.
GP-UCB, EI, and PI are classic general-purpose AFs that have been shown to achieve good regret performance for black-box functions drawn from GP. 
For GP-UCB, we tune its exploration parameter $\delta$ by a grid search between $10^{-1}$ and $10^{-6}$.
Among the family of entropy search methods, MES is a strong and computationally efficient benchmark method that achieves superior performance for several global optimization benchmark functions and GP functions \cite{wang2017max}.
For MES, we use Gumbel sampling and take one sample for $y_*$, as suggested by the original paper \cite{wang2017max}.
MetaBO is a neural AF trained via policy-based RL and recently achieves superior performance in BO. 
For fair and reproducible evaluations, we use the pre-trained model of MetaBO provided by \cite{volpp2020metalearning}.
As the original MetaBO does not address the use of meta-data, for a more comprehensive comparison, we further consider a few-shot variant of MetaBO (termed MetaBO-T), which is obtained by performing 100 more training iterations on the pre-trained MetaBO model using the meta-data for few-shot fine-tuning.

\textbf{Configuration of FSAF.} For training, we construct a collection of training tasks, each of which is a class of GP functions with either an RBF, Matern-3/2, or a spectral mixture kernel with different parameters (e.g., lengthscale and periods).
We take $N=5$, $K=5$, and $S=1$ given the memory limitation of GPUs.
Despite the small values of $N,K,S$, these choices already provide superior performance.
For the reward design of FSAF, we use $g(z)=-\log z$ to encourage high-accuracy solutions.
For testing, we use the model with the best average total return during training as our initial model, which is later fine-tuned via few-shot fast adaptation for each task.
For a fair comparison, we ensure that FSAF and MetaBO-T use the same amount of meta-data in each experiment.




\vspace{-3mm}
\begin{figure*}[!htbp]
\centering 
  \begin{minipage}{\textwidth}
    \centering 
    \hspace{-3mm}
    \subfigure[]{
    \label{Fig.ob.0}
    \includegraphics[width=0.2\textwidth]{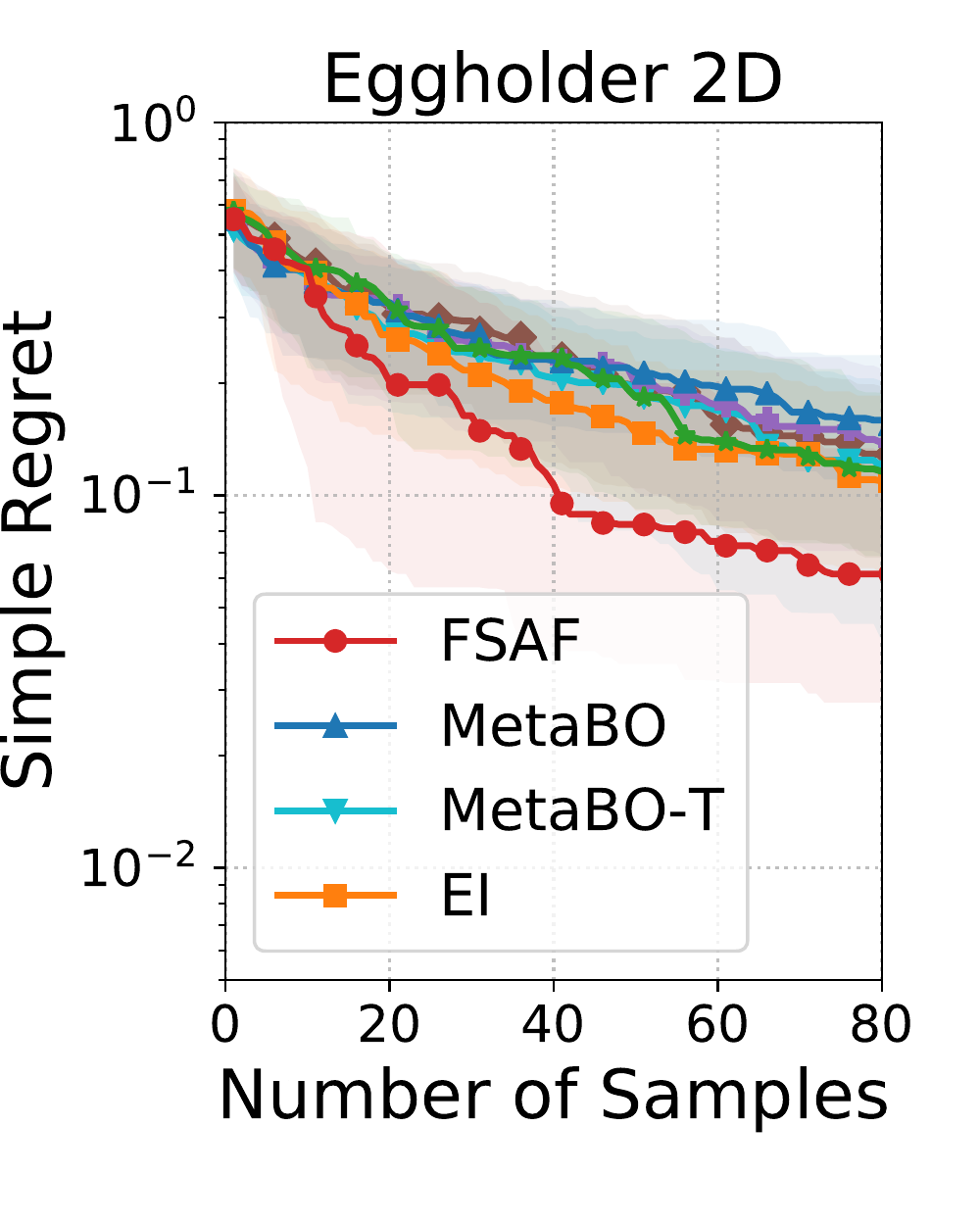}}
    \hspace{-4mm}
    \subfigure[]{
    \label{Fig.ob.1}
    \includegraphics[width=0.2\textwidth]{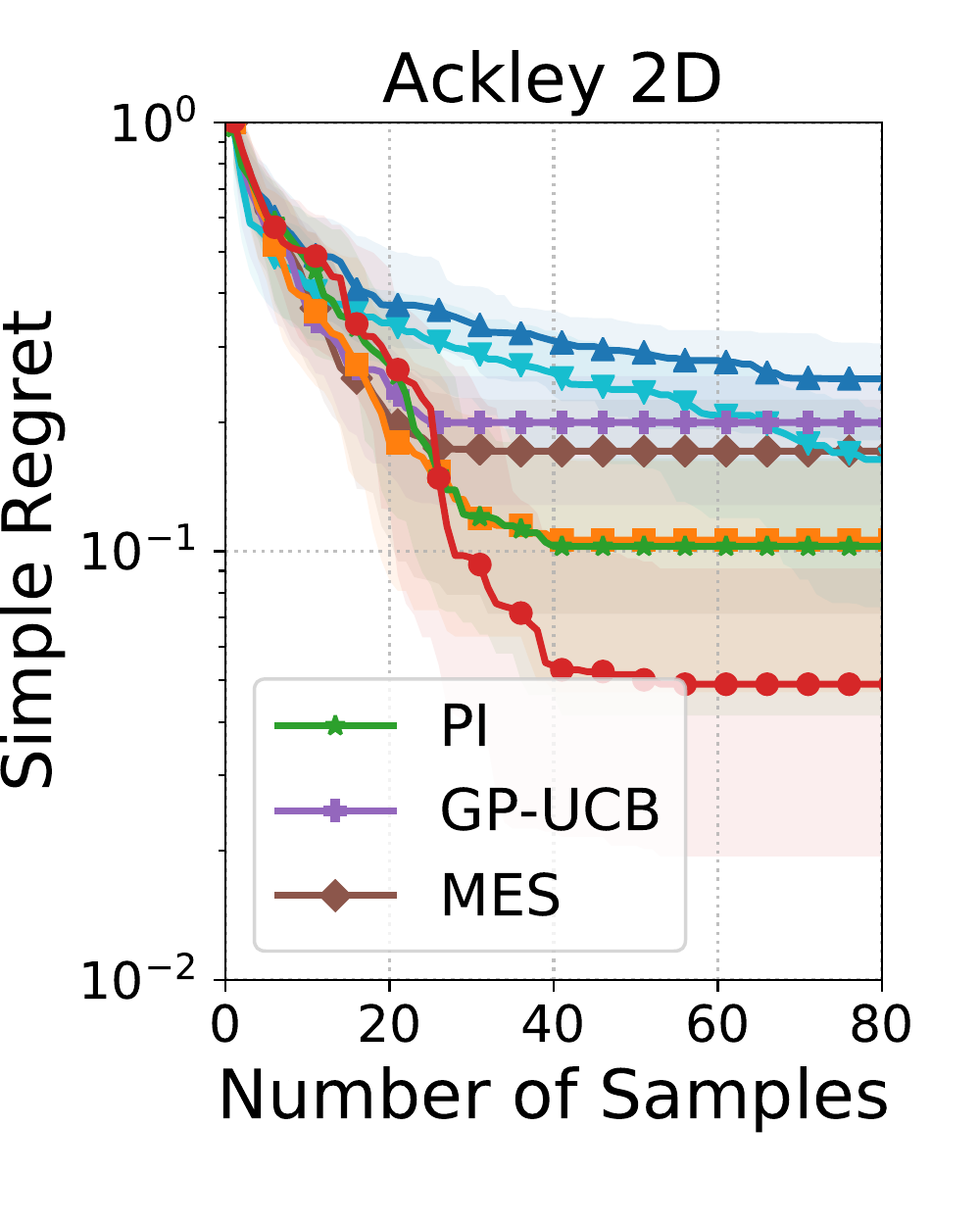}}
    \hspace{-4mm}
    \subfigure[]{
    \label{Fig.ob.2}
    \includegraphics[width=0.2\textwidth]{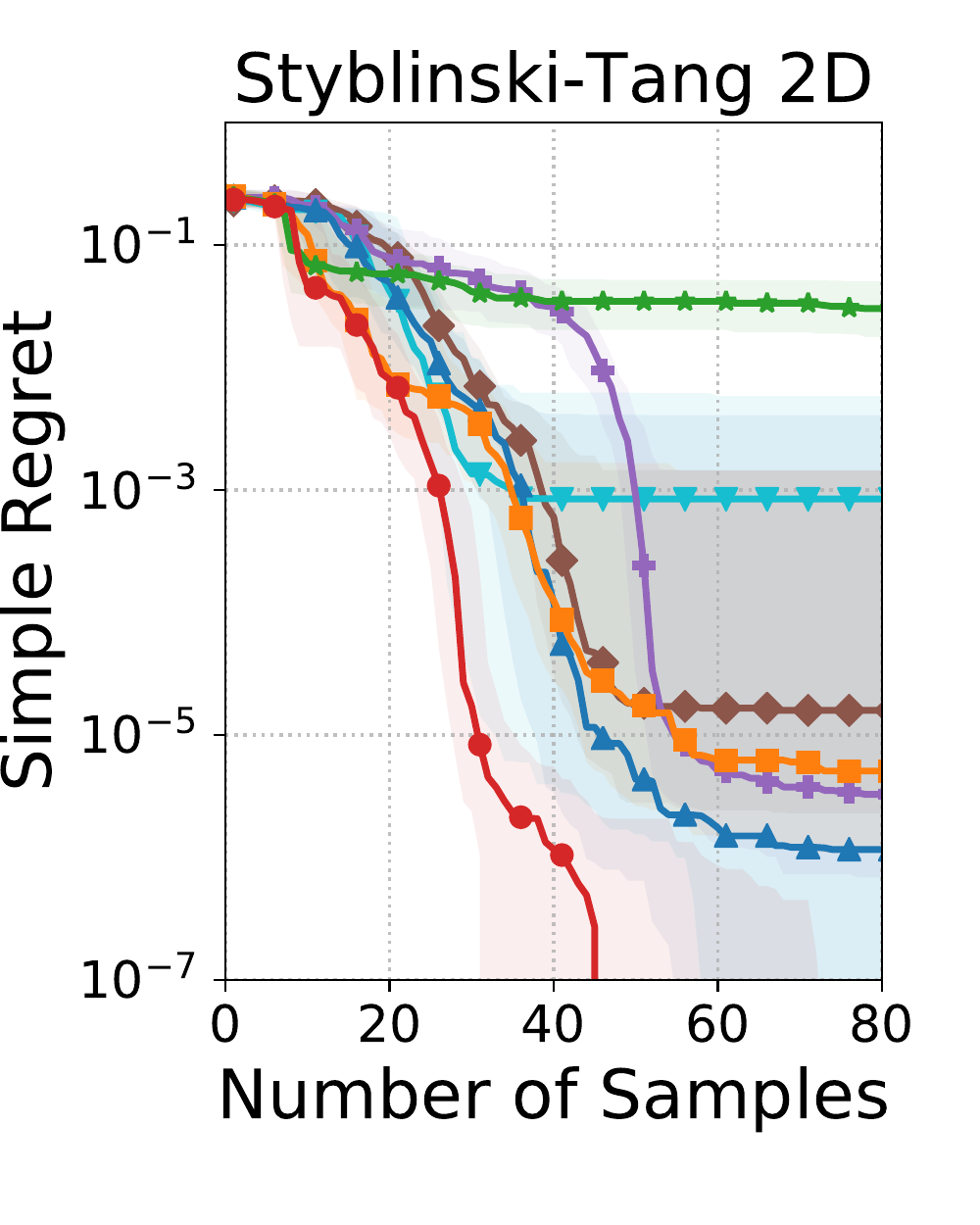}}
    \hspace{-4mm}
    \subfigure[]{
    \label{Fig.ob.4}
    \includegraphics[width=0.2\textwidth]{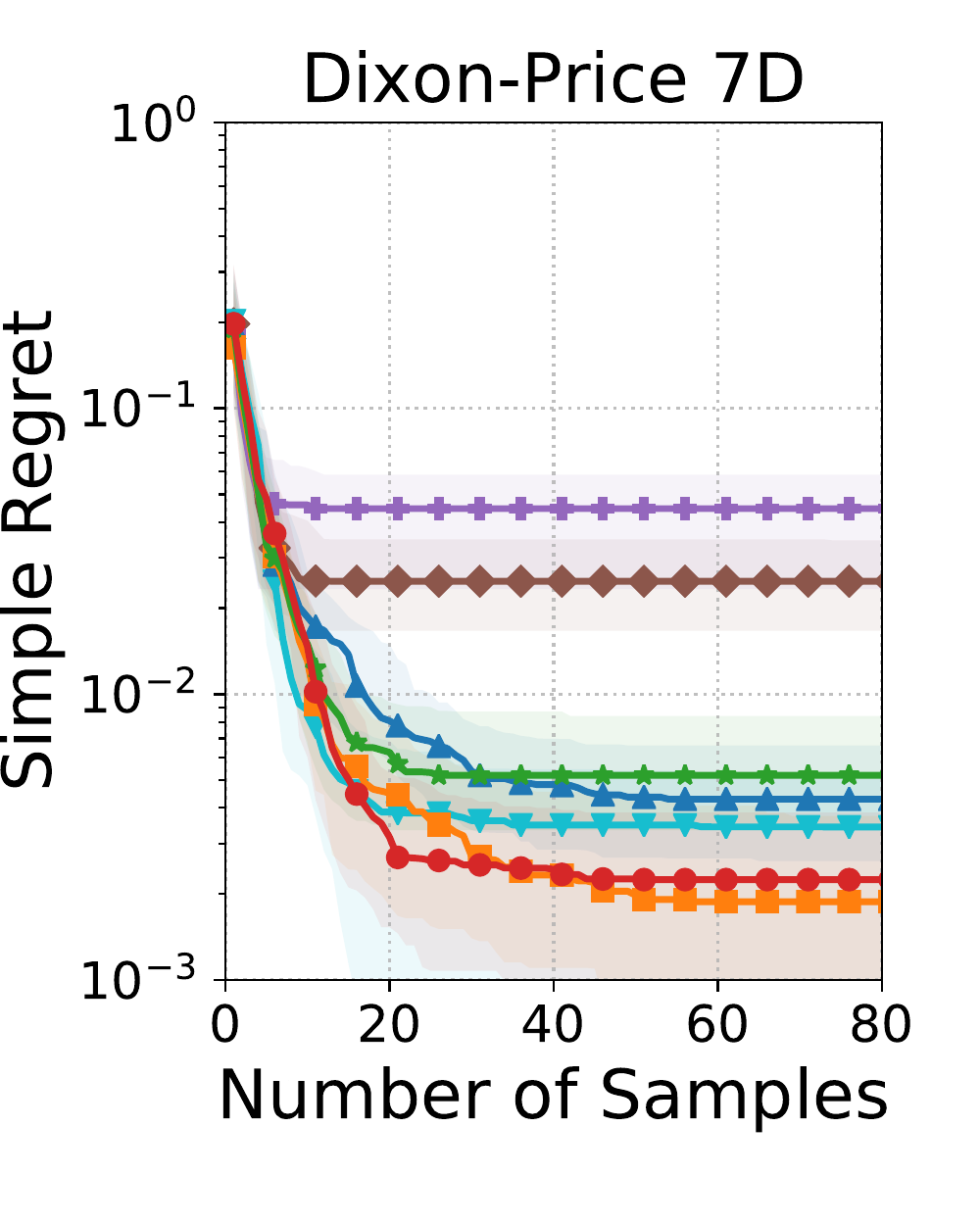}}
    \hspace{-4mm}
    \subfigure[]{
    \label{Fig.ob.5}
    \includegraphics[width=0.2\textwidth]{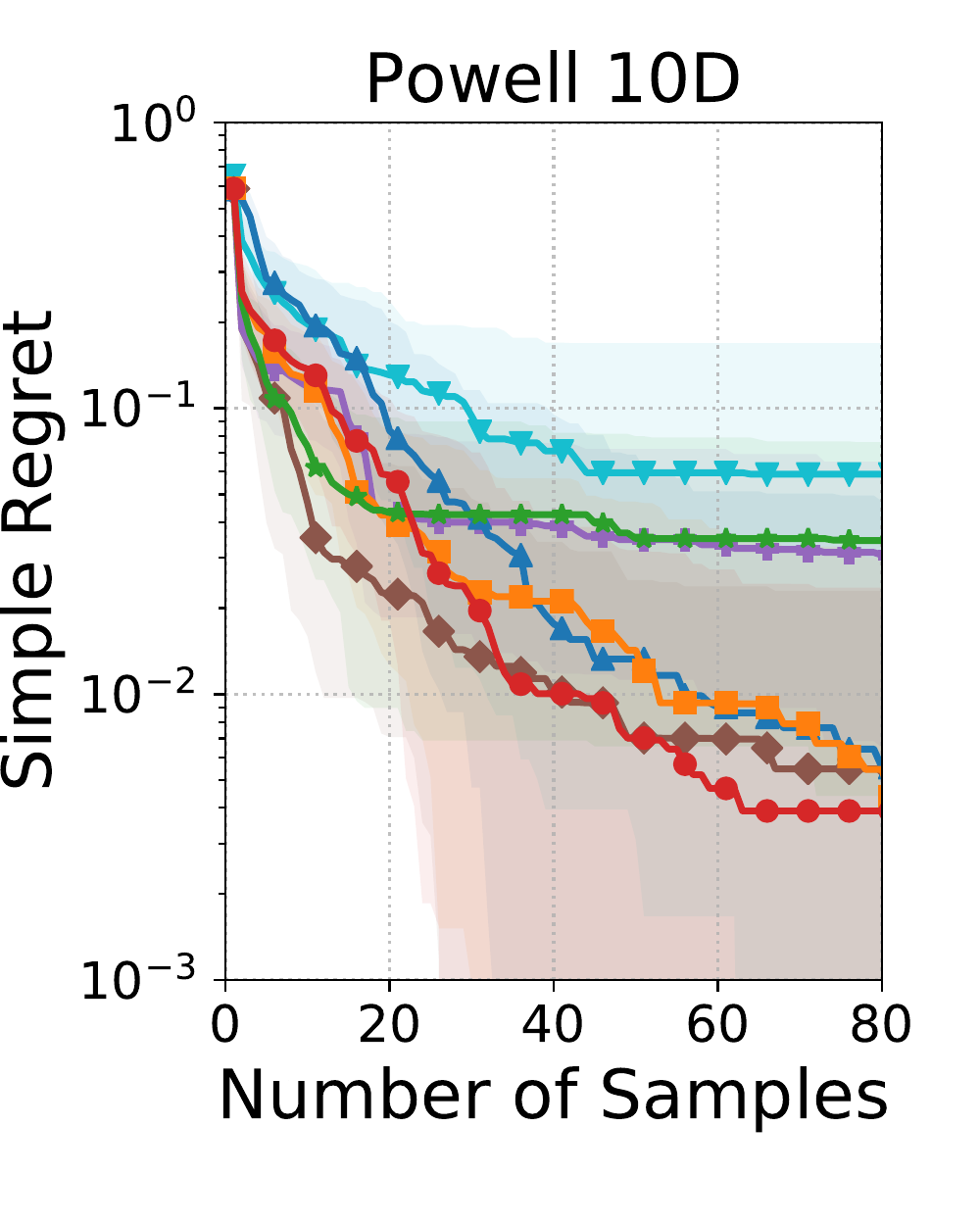}}
    \vspace{-3mm}
    \caption{Simple regrets of FSAF and the benchmark methods for optimization benchmark functions. The line and the shaded areas show the median and the 25\%/75\% percentiles, respectively.}
    \label{Fig:blackbox}
  \end{minipage}
\end{figure*}

\vspace{-4mm}
\begin{figure*}[!htbp]
\centering 
  \begin{minipage}{\textwidth}
    \centering 
    \hspace{-3mm}
    \subfigure[]{
    \label{Fig:real1.0}
    \includegraphics[width=0.2\textwidth]{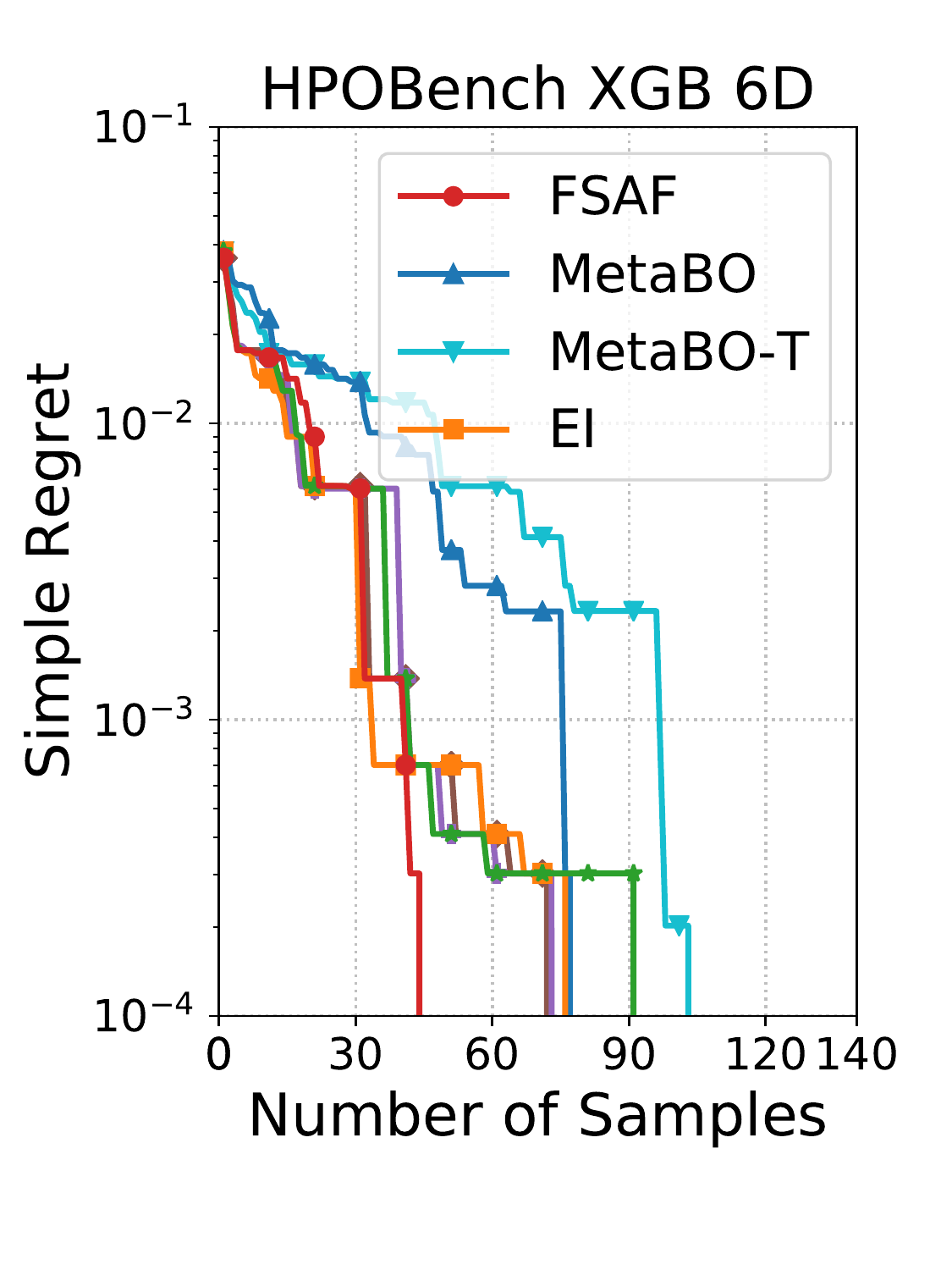}}
    \hspace{-4mm}
    \subfigure[]{
    \label{Fig:real1.2}
    \includegraphics[width=0.2\textwidth]{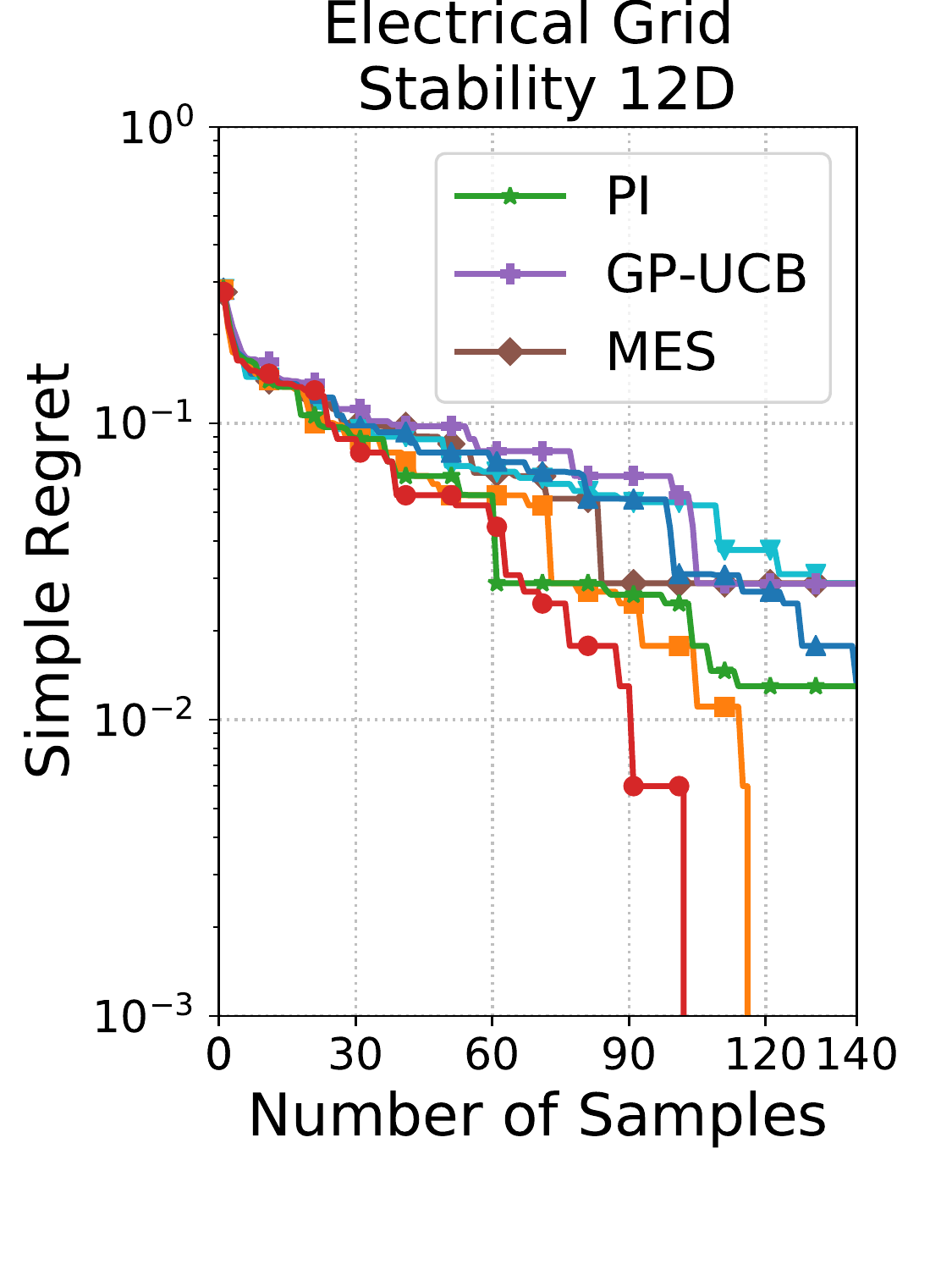}}
    \hspace{-4mm}
    \subfigure[]{
    \label{Fig:real1.3}
    \includegraphics[width=0.2\textwidth]{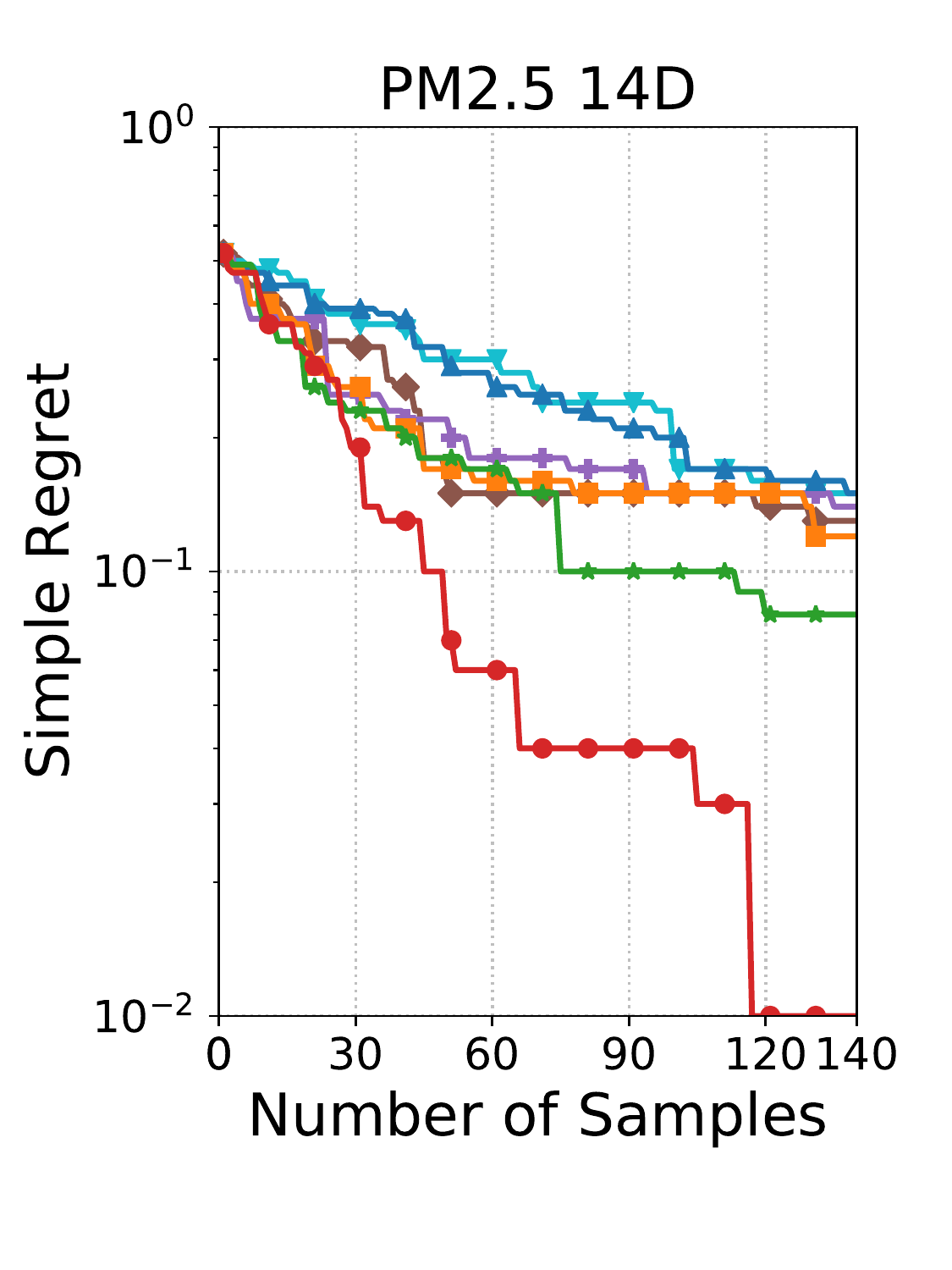}}
    \hspace{-4mm}
    \subfigure[]{
    \label{Fig:real2.0}
    \includegraphics[width=0.2\textwidth]{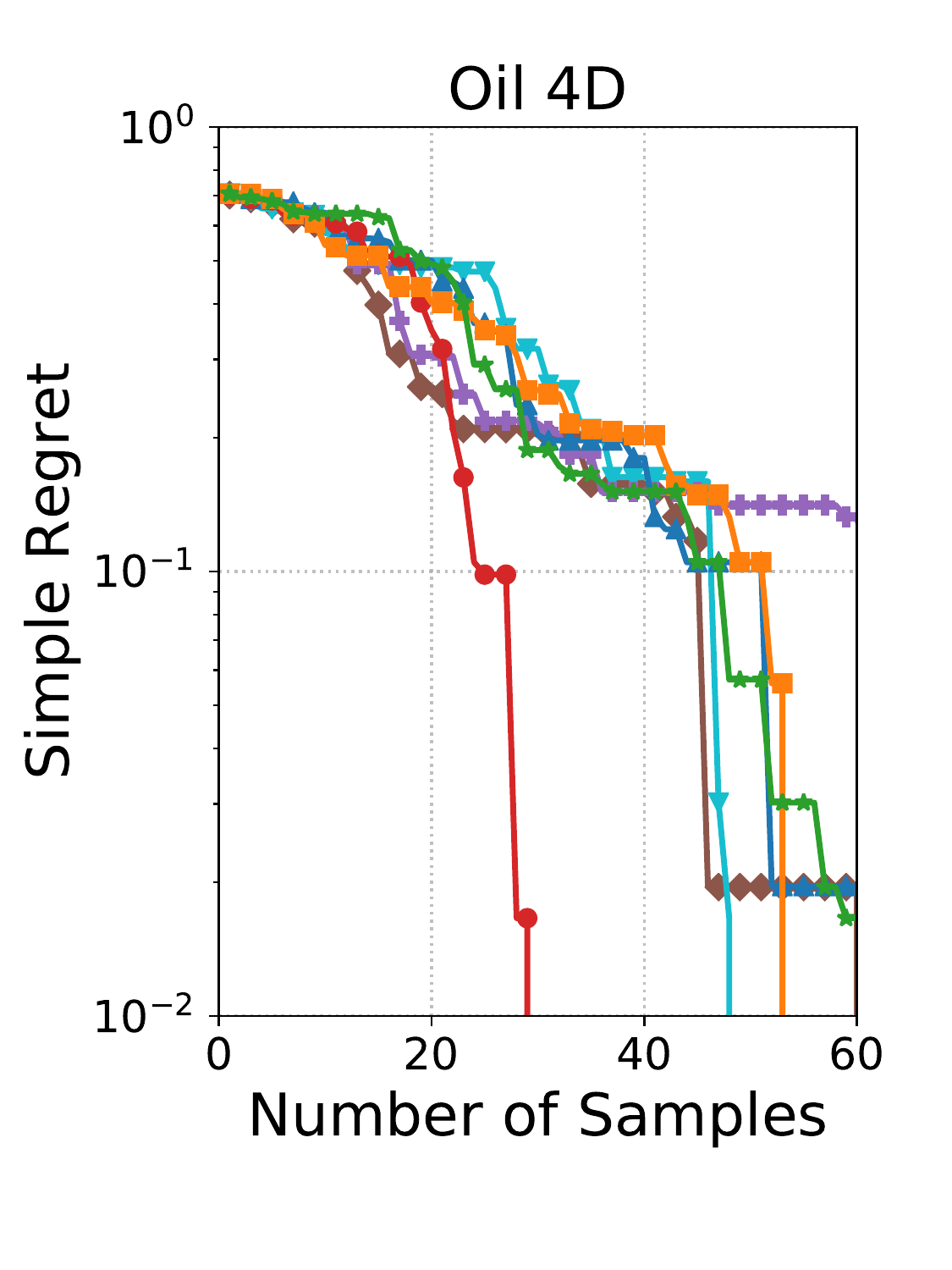}}
    \hspace{-4mm}
    \subfigure[]{
    \label{Fig:real2.1}
    \includegraphics[width=0.2\textwidth]{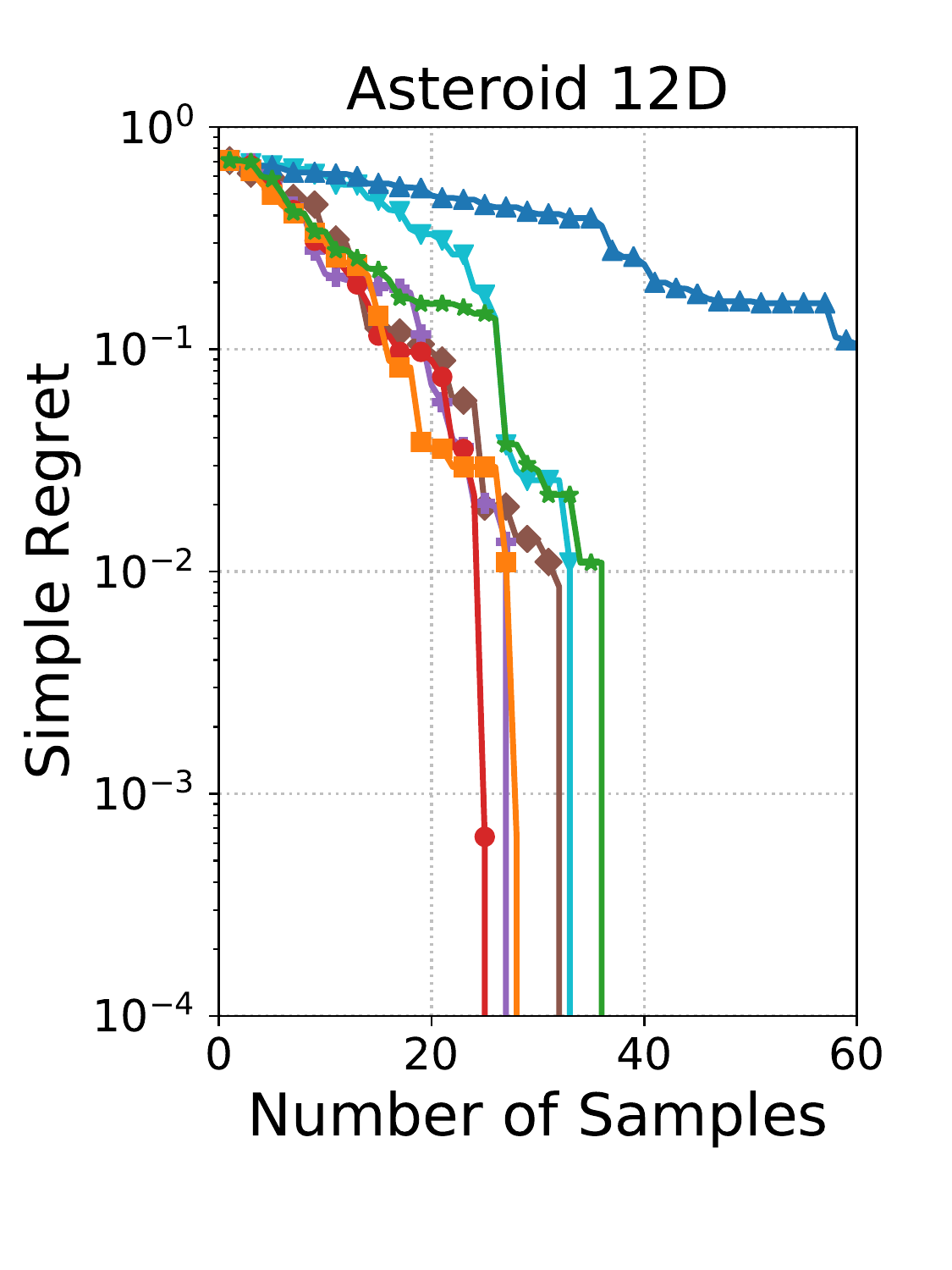}}
    \vspace{-3mm}
    \caption{Median simple regrets of FSAF and other benchmark methods for real-world test functions. Other percentiles are provided in Appendix \ref{app:results} for visual clarity.}
    \label{Fig:real}
  \end{minipage}

\end{figure*}

\textbf{Does FSAF adapt effectively to a wide variety of black-box functions?} 
To answer this, we first evaluate FSAF and the benchmark methods on several types of  standard optimization benchmark functions, including: (i) \textit{Ackley} and \textit{Eggholder}, which are functions with a large number of local optima but with different symmetry structures; (ii) \textit{Dixon-Price}, a valley-shaped function; (iii) \textit{Styblinski-Tang}, a smooth function with a couple of local optima; (iv) \textit{Powell}, a 10-dimensional function (the highest input dimension among the five functions).
As the $y$ values of these functions can be one or more orders of magnitude different from each other, for ease of comparison, we scale all the values of the functions to $[-2,2]$. 
Such scaling still preserves the salient structure and variations of each test function.
To construct the training and testing sets, we apply random translations and re-scalings of up to +/- 10\% to $x$ and $y$ values, respectively.
In this case, we consider 5-shot adaptation for FSAF and use the same amount of meta-data for MetaBO-T.
From Figure \ref{Fig:blackbox}, we observe that FSAF is constantly the best or among the best of all the methods under all the test functions.
We observe that MetaBO performs poorly under functions with many local optima but performs better under smooth functions (e.g., Styblinski-Tang and Powell).
This might be due to the fact that MetaBO was trained with smooth GP functions and lacks the ability to adapt to functions with more local variations.
We also find that MetaBO-T benefits from meta-data and improves upon MetaBO in some functions.
While this manifests the potential benefits of using meta-data for MetaBO, this also suggests that brute-force fine-tuning is not effective and a more careful design like FSAF is needed. 
Moreover, among the 6 benchmark methods, the best-performing AF indeed varies under different types of functions.
This corroborates the commonly-seen phenomenon and our motivation.



We proceed to evaluate FSAF on test functions obtained from five open-source real-world tasks in different application domains.
Based on the smoothness characteristics\footnote{To better understand the characteristics of each real-world dataset, we extract the smoothness information through marginal likelihood maximization on a surrogate GP model with an RBF kernel.}, the datasets can be categorized as:
(i) Smooth in all dimensions: Asteroid size prediction ($d=12$); (ii) Smooth in all but one dimension: hyperparameter optimization for XGBoost ($d=6$) and air quality prediction in PM 2.5 ($d=14$); (iii) Smooth in about half of the dimensions: maximization of electric grid stability ($d=12$); (iv) Non-smooth in all dimensions: location selection for oil wells ($d=4$).
The detailed description of the datasets is in Appendix \ref{app:exp details}.
In this setting, we consider 1-shot adaptation for FSAF, a rather sample-efficient scenario of few-shot learning.
From Figure \ref{Fig:real}, we observe that FSAF remains the best or among the best for all the five real-world test functions, despite the salient structural differences of the datasets.
For a more comprehensive comparison, we also evaluate all the AFs on synthetic functions from GP and observe the similar behavior. 
Due to space limitation, the results of GP functions are provided in Appendix \ref{app:results}.
Based on the above discussion, we confirm that FSAF indeed achieves favorable adaptability.

\vspace{-1mm}
\begin{figure*}[!htbp]
\centering 
  \begin{minipage}{\textwidth}
    \subfigure[]{
    \label{Fig.Training_curve}
    \includegraphics[width=0.44\textwidth]{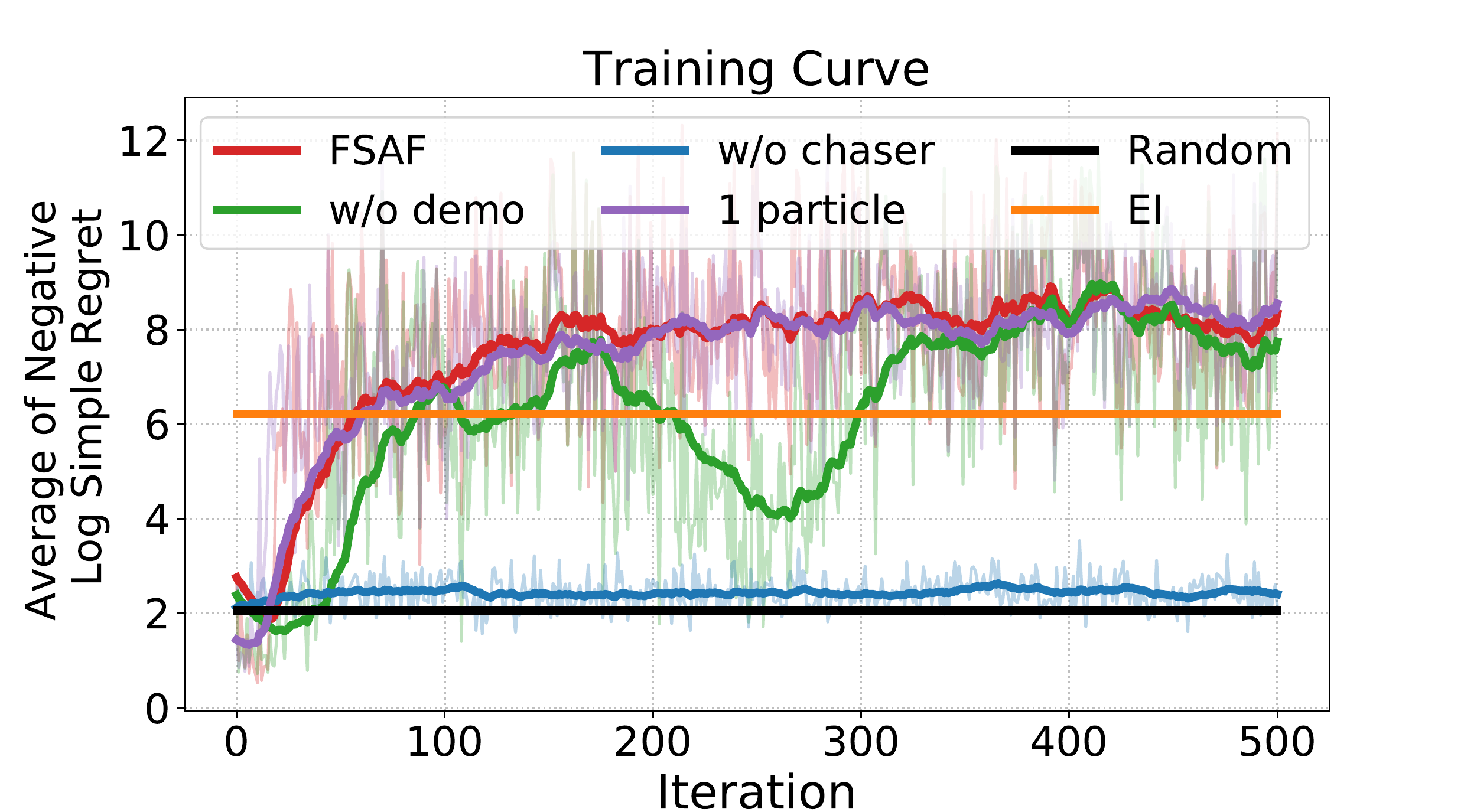}}
    \hspace{-7mm}
    \subfigure[]{
    \label{Fig.particle_compare.0}
    \includegraphics[width=0.195\textwidth]{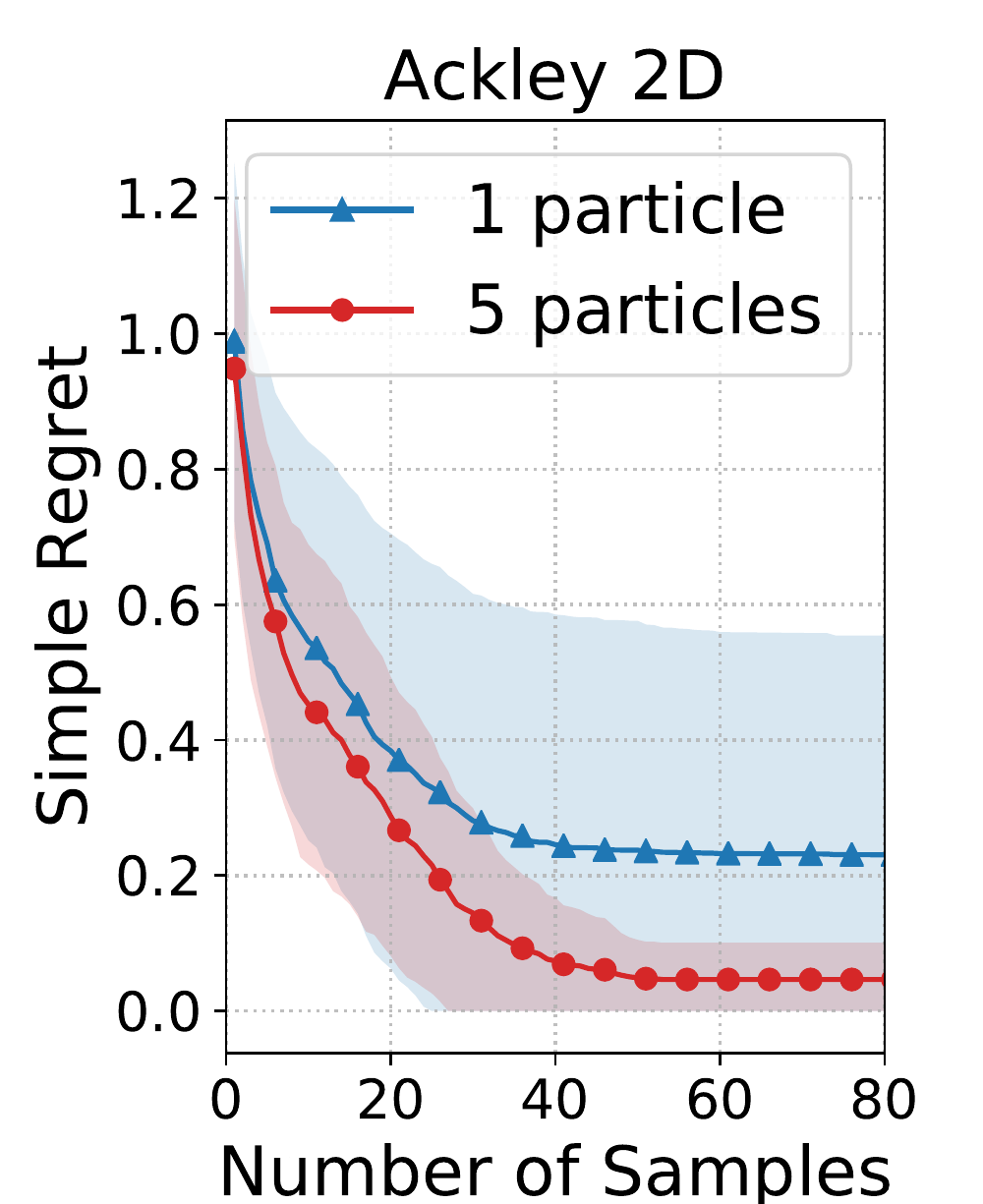}}
    \hspace{-4mm}
    \subfigure[]{
    \label{Fig.particle_compare.1}
    \includegraphics[width=0.195\textwidth]{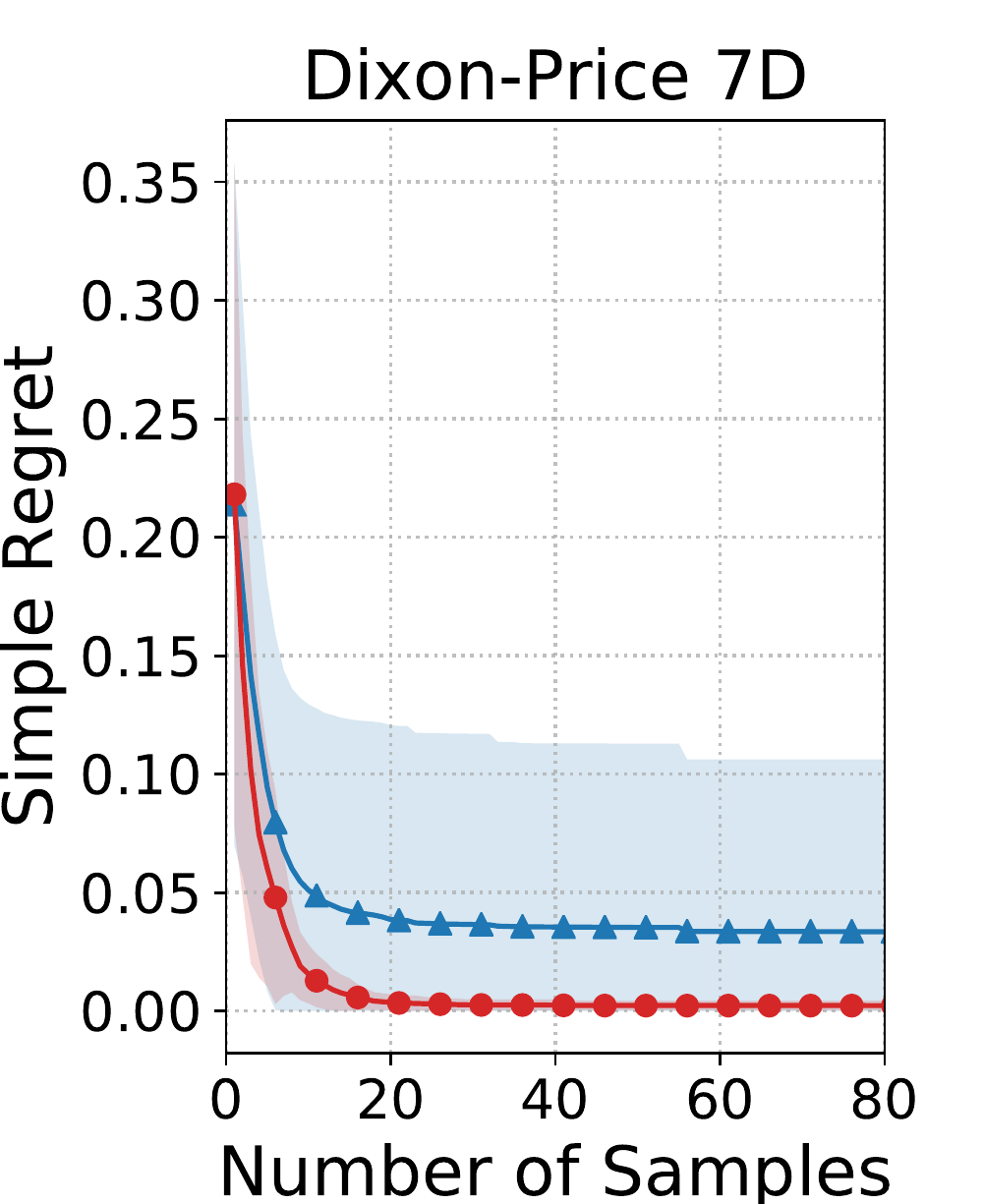}}
    \hspace{-4mm}
    \subfigure[]{
    \label{Fig.particle_compare.2}
    \includegraphics[width=0.195\textwidth]{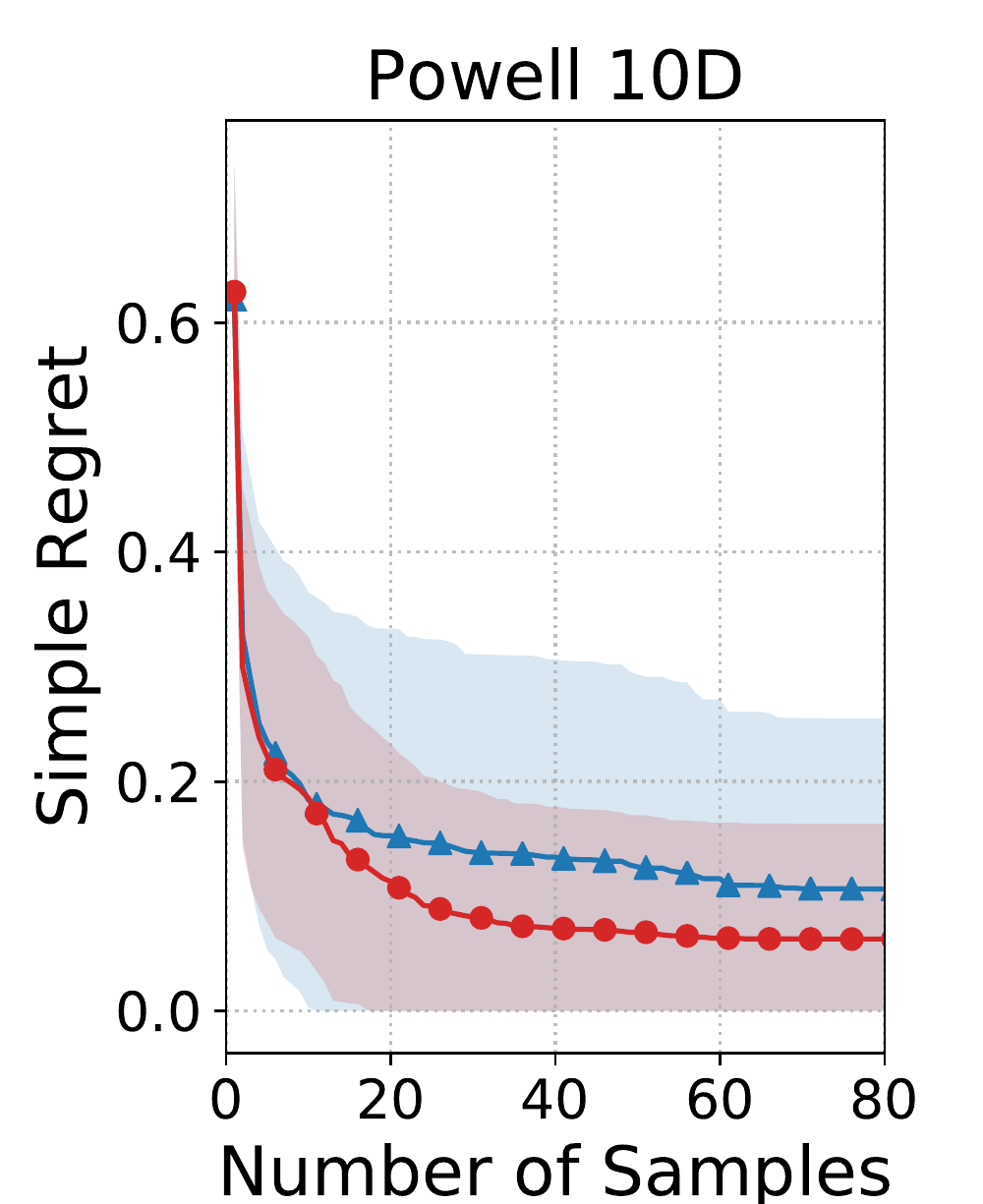}}
    \vspace{-3mm}
    \caption{Ablation study for FSAF: (a) Training curves of FSAF and its ablations (EI and Random as baselines); (b)-(d) Testing performance of FSAF with 1 and 5 particles of Q-networks.}
    \label{Fig.real2}
  \end{minipage}
\end{figure*}

\textbf{Does FSAF mitigate the overfitting issue?}
To answer this, we show the training curves of FSAF and its ablations, including: (i) FSAF without using the demo replay buffer (termed ``w/o demo''); (ii) FSAF with the chaser meta-loss replaced by TD loss (termed ``w/o chaser''); (iii) FSAF with only 1 particle (termed ``1 particle''), which is equivalent to DQN+MAML with an additional demo replay buffer.
The training metric plotted is the negative logarithm of simple regret at $t=30$ (averaged over episodes in each iteration), which is consistent with the rewards of FSAF and can better demonstrate the differences in training. 
The results of EI and random sampling are given as references.

\vspace{-2mm}
\begin{itemize}[leftmargin=*]
    \item \textbf{The Bayesian variant of DQN does mitigate overfitting.} This is confirmed by the fact that the training curves of FSAF with 1 and 5 particles are quite close, while in Figures \ref{Fig.particle_compare.0}-\ref{Fig.particle_compare.2} the testing performance of FSAF with 5 particles appears much better than that of 1 particle.
    \vspace{-1mm}
    \item \textbf{The demo-based prior helps improve the training stability.} This is verified by that without the demo replay buffer, the training progress becomes apparently slower initially and is subject to more variations throughout the training.
    \vspace{-1mm}
    \item \textbf{The chaser meta-loss appears effective in FSAF.} Interestingly, we find that chaser meta-loss results in stable and effective training, while the TD meta-loss can barely make any progress.
\end{itemize}

\begin{figure*}[!htbp]
\centering 
  \begin{minipage}{\textwidth}
    \centering 
    \subfigure[]{
    \label{Fig.growth.0}
    \includegraphics[width=0.205\textwidth]{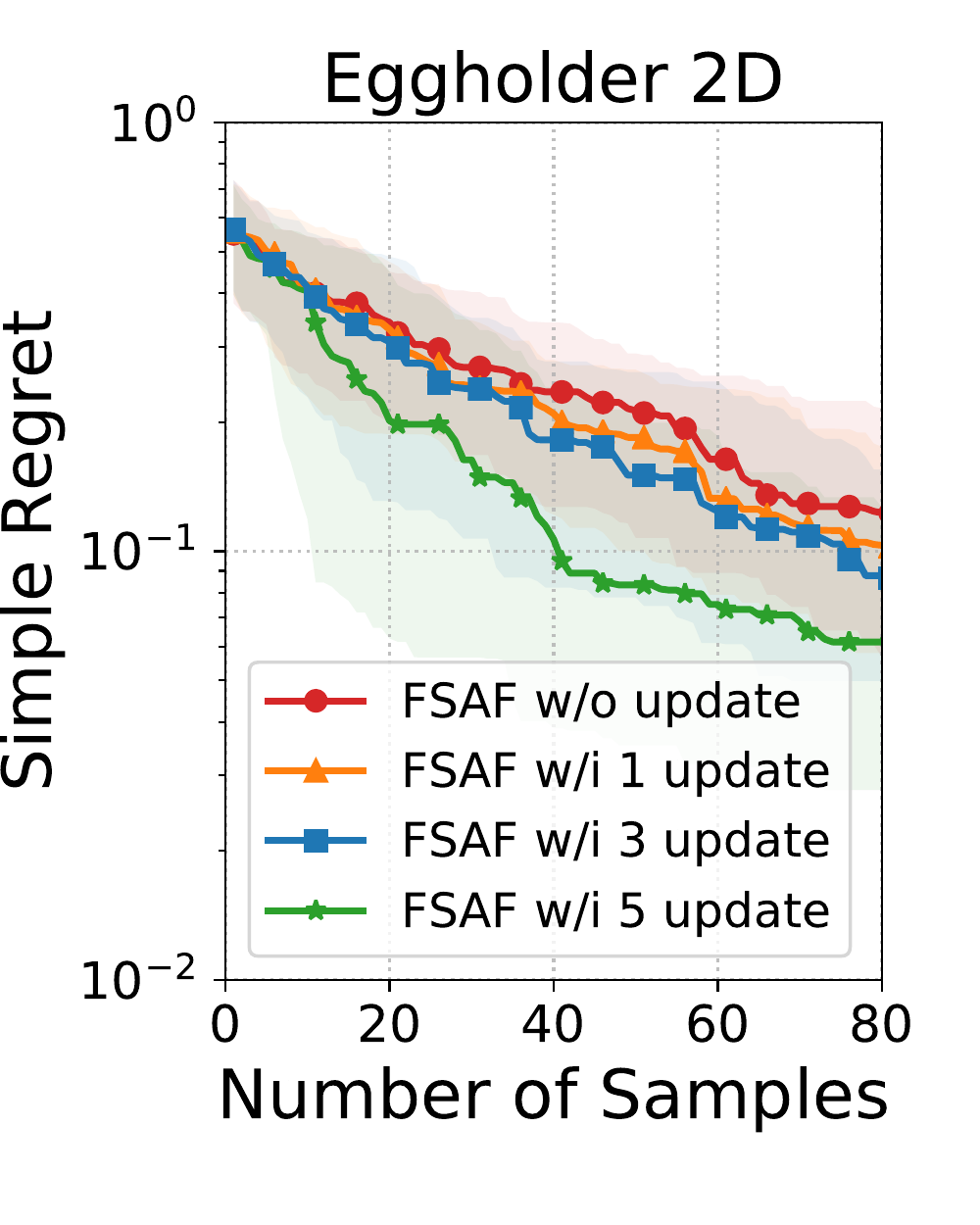}}
    \hspace{-4mm}
    \subfigure[]{
    \label{Fig.growth.1}
    \includegraphics[width=0.205\textwidth]{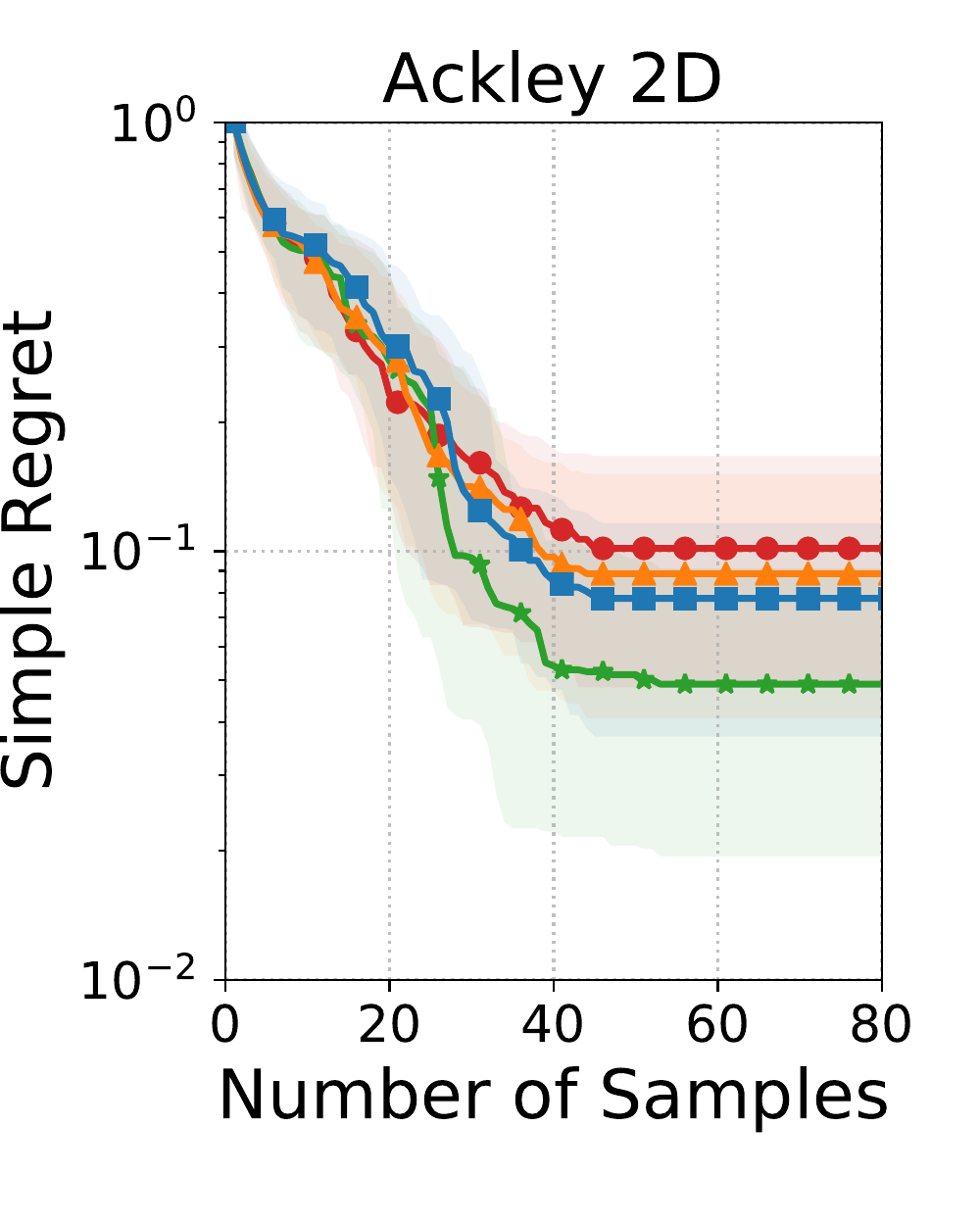}}
    \hspace{-4mm}
    \subfigure[]{
    \label{Fig.growth2.0}
    \includegraphics[width=0.205\textwidth]{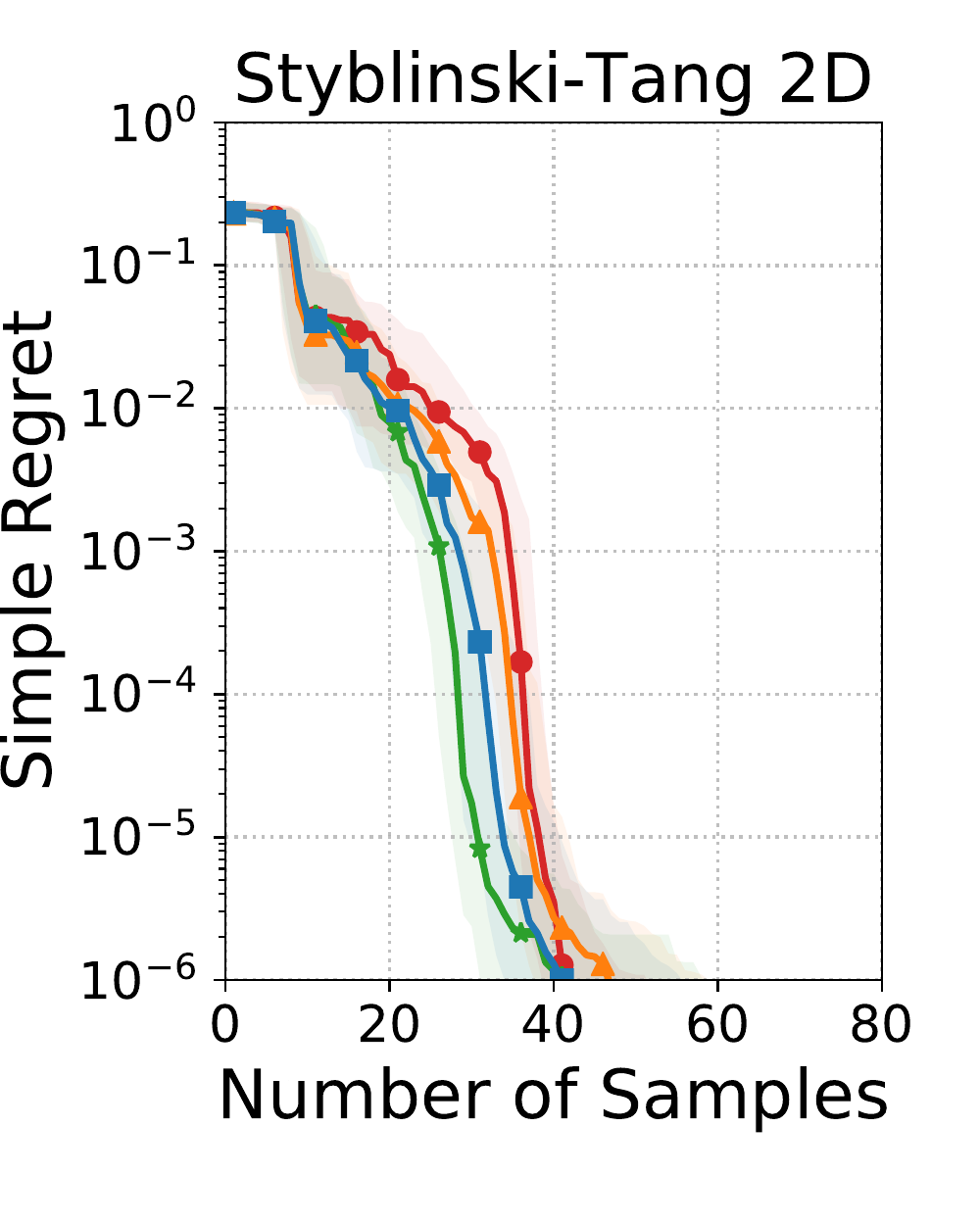}}
    \hspace{-4mm}
    \subfigure[]{
    \label{Fig.growth.2}
    \includegraphics[width=0.205\textwidth]{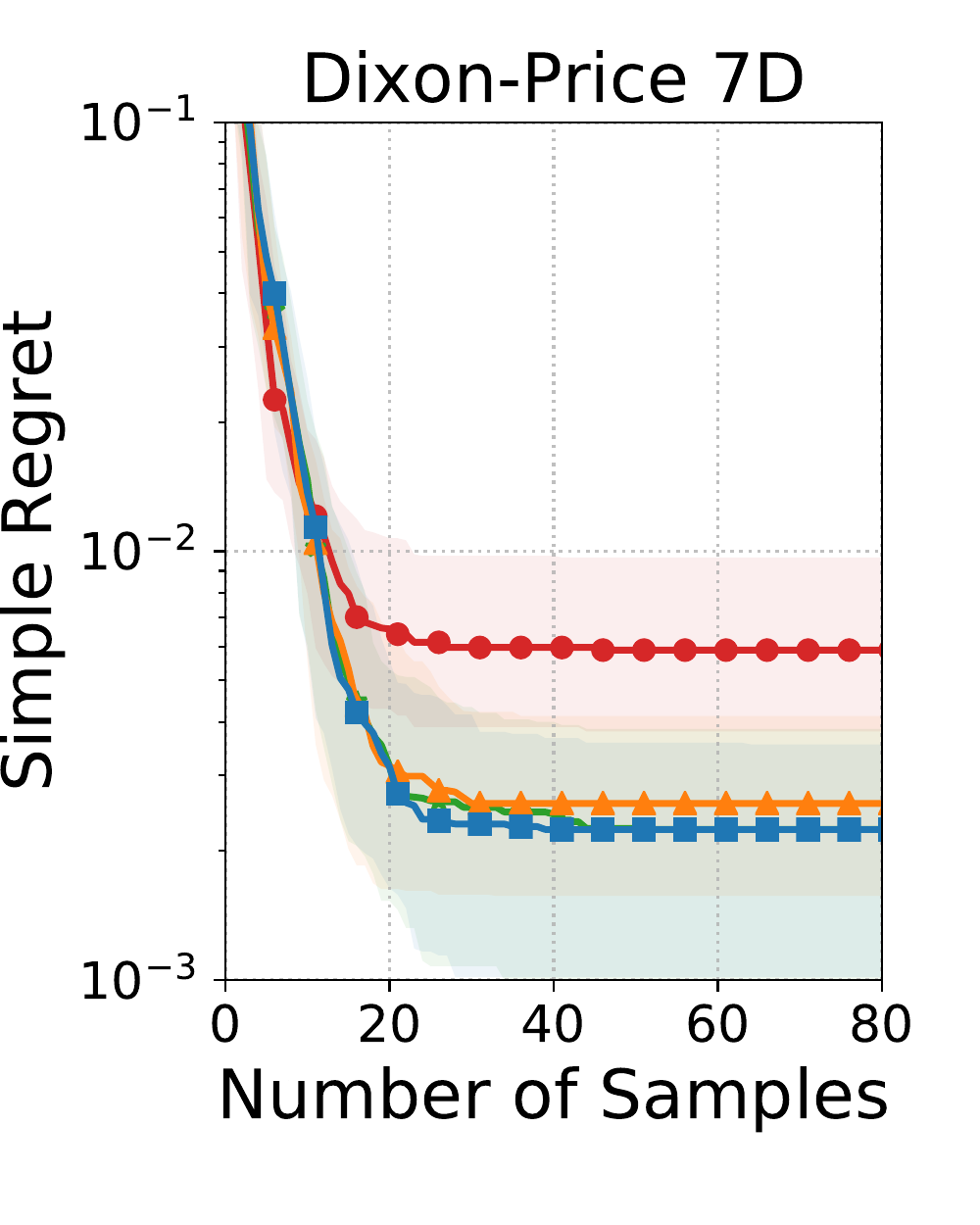}}
    \hspace{-4mm}
    \subfigure[]{
    \label{Fig.growth2.2}
    \includegraphics[width=0.205\textwidth]{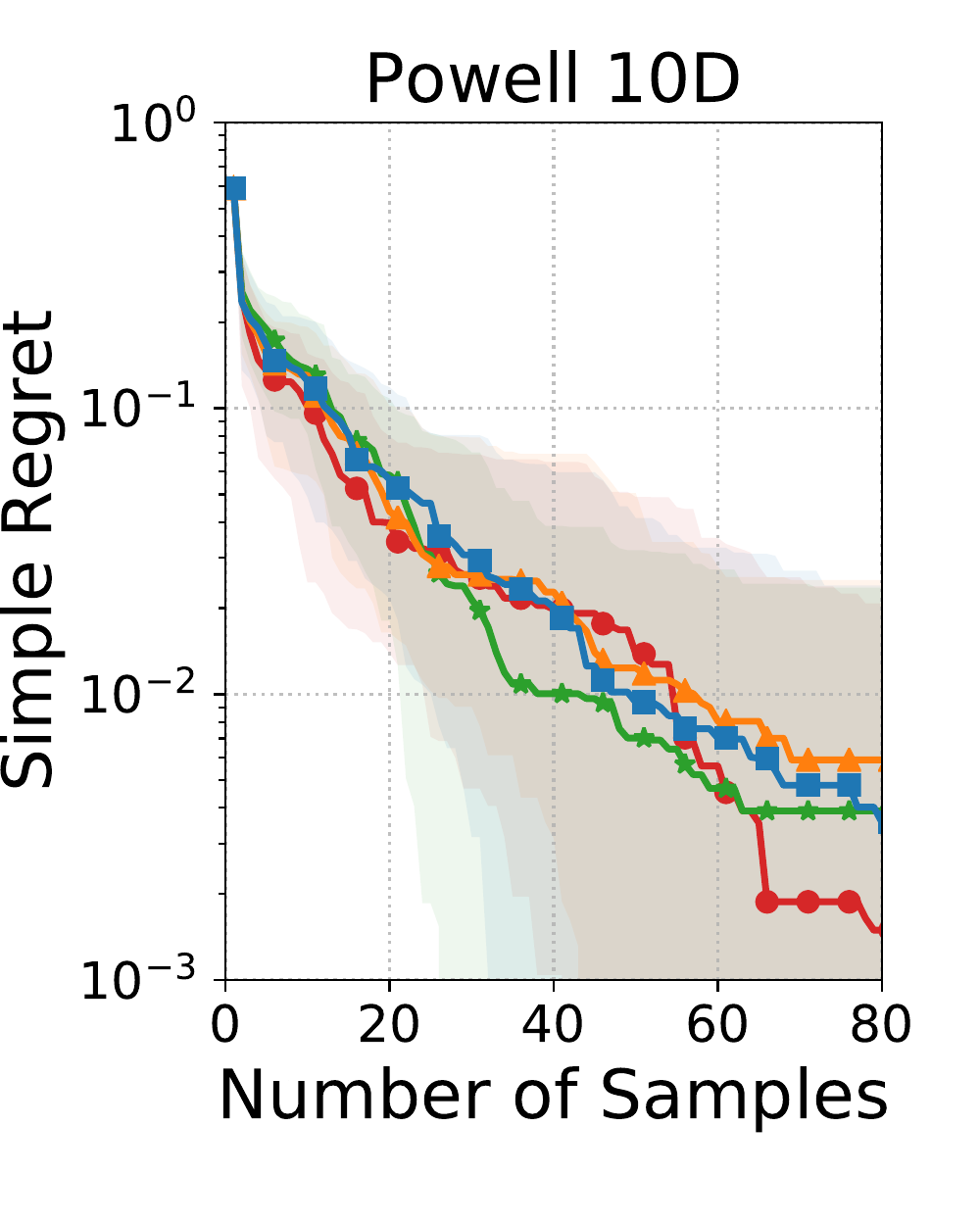}}
    \vspace{-3mm}
    \caption{Simple regrets of FSAF vs number of gradient steps for optimization benchmark functions.}
    \label{Fig:gradient updates}
  \end{minipage}
\end{figure*}

\textbf{Does FSAF benefit from the few-shot gradient updates?}
To better understand the effect of few-shot gradient updates, Figure \ref{Fig:gradient updates} shows the simple regrets of FSAF under different number of gradient updates (i.e., $K$ defined in Section \ref{section:alg:BMAML}) for the optimization benchmark functions.
We find that the adaptation effect does increase with $K$ for small $K$'s for most of the cases. 
This appears consistent with the general observations of MAML-like algorithms \cite{finn2017model}.

\vspace{-3mm}

\begin{remark}[FSAF and \cite{wistuba2021fewshot}]
\label{remark:FSAF and FSBO}
\normalfont As mentioned in Section \ref{section:intro}, \cite{wistuba2021fewshot} proposes to leverage meta-data to fine-tune the initialization of the GP kernel parameters for the off-the-shelf AFs (called FSBO in \cite{wistuba2021fewshot}). By contrast, FSAF uses meta-data for fast adaptation of an AF, which is a direction orthogonal to FSBO. Moreover, it is possible for our FSAF and FSBO to complement each other. We provide experimental results to verify this argument in Appendix \ref{app:results}.
\end{remark}

%% file: 6-related.tex
\vspace{-2mm}
\section{Related Work}
\label{section:related}
\vspace{-2mm}

\textbf{Bayesian optimization via myopic AFs.} 
In BO, AFs are usually designed to reflect the trade-off between exploration and exploitation of the global optima through maximizing some cheap-to-evaluate \textit{myopic} objective, which typically corresponds an one-step look-ahead strategy. 
Such AFs have been designed from various perspectives on how to address the trade-off.
For instance, under GP-UCB, AF is designed from the perspective of optimism in the face of uncertainty \cite{srinivas2010gaussian}. 
Under EI, AF incorporates the expected one-step improvement in terms of the best observation made so far by design \cite{movckus1975bayesian,jones1998efficient}. 
Similarly, under the knowledge gradient method \cite{frazier2009knowledge}, AF is chosen to be the expected improvement in the largest posterior mean over the input domain.
Instead of using the expected improvement, PI \cite{Kushner1964ANM} uses the tail probability of improvement as the AF for BO. 
Under entropy search methods, AF is the expected reduction in the uncertainty of the global optima \cite{hennig2012entropy,hernandez2014predictive,wang2017max}. 
Under Thompson sampling, AF values are directly drawn from the posterior predictive distribution \cite{chowdhury2017kernelized}.
As the AFs are often handcrafted according to different perspectives of the exploration-exploitation trade-off, the performance of an AF can vary significantly under different types of black-box functions \cite{hernandez2014predictive}.
Thus, designing an AF that can adapt and demonstrate outstanding performance to a wide variety of black-box functions remains a critical challenge. 
This is the goal of our paper: we propose a reinforced few-shot learning framework, in which an initial AF is learned with fast adaption to various black-box functions given a limited number of samples (e.g., 1-shot).

\textbf{Bayesian optimization via non-myopic methods.}
To go beyond myopic AFs, recently there has been some interest in designing non-myopic strategies for BO from the perspective of dynamic programming (DP).
For example, \cite{osborne2009gaussian,ginsbourger2010towards} characterized the optimal non-myopic strategies for BO. 
To tackle the intractable DP problem, \cite{gonzalez2016glasses,lam2016bayesian} proposed to approximately solve the DP problem for BO by different simulation techniques.
\cite{wu2019practical} focused on the two-step lookahead AFs for BO and proposed an efficient Monte-Carlo method to find such AFs.
Instead of fixing the horizon in advance, \cite{yue2020non} proposed a principled way to select the rolling horizon for approximate dynamic programming in BO. 
\cite{jiang2020binoculars} proposed to achieve non-myopia by maximizing a lower bound on the multi-step expected utility.
\cite{jiang2020efficient} proposed a tree-based AF to enable approximate multi-step lookahead in a one-shot manner by leveraging the reparameterization trick.
The proposed FSAF is also non-myopic by nature as it implicitly takes multi-step effect into account by using Q-networks as AFs.
Different from the above multi-step solutions, FSAF is meant to serve as an AF that can adapt to a wide variety of black-box functions given only a small amount of meta-data.

\textbf{Few-shot learning.} Few-shot learning has recently attracted much attention since the data scarcity has become a bottleneck to many real-world machine learning tasks \cite{thrun2012learning}. 
Using prior knowledge, few-shot learning can rapidly generalize a pre-trained model to new tasks given only a few examples. 
One of the most representative frameworks for few-shot learning is MAML \cite{finn2017model}, where an initialization of parameters is pre-trained to be close to the tasks drawn from the task distribution, and thus a few gradient steps are sufficient to adapt it to a specific task. 
There are several follow-up studies for MAML. For instance, in \cite{antoniou2018train}, vanilla MAML is shown to be sensitive to the network hyperparameters, often leading to training instability. To overcome that, learning the majority of hyperparameters end to end is proposed. In \cite{yoon2018bayesian}, a Bayesian version of MAML is proposed to learn complex uncertainty structure beyond a simple Gaussian approximation. In \cite{vuorio2019multimodal}, MAML is revisited under multimodel task distributions. To better adapt to different modes, a modulation network is proposed to identify the mode of the task distribution and then customize the meta-learned prior for the identified mode. 


\textbf{Meta Bayesian optimization.} Meta-learning for BO has been discussed in many prior studies. Some of them \cite{swersky2013multi,yogatama2014efficient} use a single GP model as the prior of all the tasks. However, in real-world applications, different tasks may have different task-specific features that fail to be captured by a single model. To mitigate the issue, several studies propose to learn an ensemble of GPs as the prior. For instance, \citet{wistuba2018scalable} proposed to use a transferable AF that uses several GPs to evaluate the target dataset then weights expected improvement by the result of the evaluation for fitting in the current task.
Feurer et al. \cite{feurer2018scalable} proposed to use an ensemble of GP models obtained from each subset of meta-data and use all of them for inference in the new task.
But these works still need a sufficiently large amount of data for knowledge transfer to new tasks.
MetaBO \cite{volpp2020metalearning} used policy-based RL to train a neural AF that learns structural properties of a set of source tasks to enable knowledge transfer to related new tasks.
FSBO \cite{wistuba2021fewshot} provides a few-shot deep kernel network for a GP surrogate that can quickly adapt its kernel parameters to unseen tasks. 
However, the benefits of using meta-data for more efficient exploration via few-shot fast adaptation of AFs were not explored by \cite{volpp2020metalearning,wistuba2021fewshot}.





%% file: 7-conclusion.tex
\vspace{-1mm}
\section{Concluding Remarks}
\label{section:conclusion}
\vspace{-1mm}

\begin{singlespace}
This paper tackles the critical challenge of how to effectively adapt an AF for BO to a wide variety of black-box functions through the lens of few-shot acquisition function (FSAF) learning.
One potential limitation of FSAF is the need of a small amount of meta-data. Without any few-shot adaptation, the performance may not always be satisfactory, as shown in Figure \ref{Fig:gradient updates}. 
Despite this, through extensive experiments, we show that FSAF is indeed a promising general-purpose approach for BO.
\end{singlespace}

%% file: 8-Appendix.tex
\newpage
\section*{Appendix}
\appendix
\label{section:Appendix}
\vspace{-2mm}
\section{Detailed Training Configuration of FSAF}
\label{app:training details}

\vspace{-2mm}
\textbf{Training tasks of FSAF.} 
To construct a diverse collection of tasks for the training of FSAF, we leverage GP functions with RBF, Mat\`ern-3/2, and spectral mixture kernels to capture smooth functions, functions with abrupt local variations, and functions of periodic nature, respectively.
For both the RBF and the Mat\`ern-3/2 kernels, we consider three possible ranges of lengthscales, including $[0.07,0.13], [0.17,0.23], [0.27,0.33]$.
For the spectral mixture kernels, we consider mixtures of two Gaussian components of periods $0.3$ and $0.6$ and three possible ranges of the lengthscales, including $[0.27,0.33], [0.47,0.53], [0.57,0.63]$.
As a result, there are nine candidate tasks in the task collection $\cT$. 
In each training iteration $i$, three out of the nine tasks are selected uniformly at random from the above task collection (and hence $\lvert \cT_i\rvert=3$ in Algorithm \ref{alg:FSAF}).
The input domain of these GP functions is configured to be $[0,1]^{3}$.
To facilitate the training procedure, we discretize the continuous input domain by using the Sobol sequence to generate a grid on which the GP functions and the AFs are evaluated, {as typically done in BO}.
\vspace{-1mm}

\textbf{Demo policy for the prior of Bayesian DQN.} 
Recall from Section \ref{section:alg:DQN} that we leverage a Bayesian variant of DQN and construct an informative prior using a demo policy induced by an off-the-shelf AF.
For the demo policy, we use EI, which is computationally efficient and achieves moderate regrets in most of the cases.
To control how much the demo policy involves in the training, we use a hyperparameter termed \textit{demonstration ratio} $\kappa$, which is implemented by randomly choosing whether to use a mini-batch of transitions from the demo replay buffer with probability $\kappa$ at each SVGD step.
In our experiments, we find that a small demonstration ratio of $\frac{1}{128}$ is sufficiently effective.
\vspace{-1mm}

\textbf{Network architecture of each DQN particle.}
For all the DQN particles used in the experiments, we adopt the standard dueling network architecture \cite{wang2016dueling}, where one value network and an advantage network are maintained to produce the estimated Q-values. 
For a fair comparison between FSAF and MetaBO, both the value network and the advantage network of our FSAF are configured to have 4 fully-connected hidden layers with ReLU activation functions and 200 hidden units per layer.
As described in Section \ref{section:alg:DQN as AF}, the input of an advantage network consists of a four-tuple, namely the posterior mean $\mu_t(x)$, the posterior standard deviation $\sigma_t(x)$, the best observation so far $y_t^*$, and the ratio between the current timestamp and total sampling budget $\frac{t}{T}$. 
Accordingly, the input of a value network consists of $y_t^*$ and $\frac{t}{T}$. 
\vspace{-1mm}

\textbf{Computing Resources.}
All the training and testing processes are run on a Linux server with (i) an Intel Xeon Gold 6136 CPU operating at a maxinum clock rate of 3.7 GHz, (ii) a total of 256 GB memory, and (iii) an RTX 3090 GPU. 

Table \ref{tab:hyp} summarizes the hyperparameter configuration of the training of FSAF.
\begin{table}[!htbp]
\centering
  \caption{FSAF training hyperarameters.}
  \label{tab:hyp}
  \begin{tabular}{ll}
    \toprule
    \bf Description & \bf Value\\
    \midrule
    Batch size (i.e., $\lvert\cD^{\text{tr}}_{\tau,k}\rvert$ in Algorithm \ref{alg:FSAF})& 128 \\
    Target update interval (in terms of iterations)& 5 \\
    Lower-level learning rate (i.e., $\eta$ in (\ref{eq:empirical SVGD})) & 0.01 \\
    Upper-level learning rate (i.e., $\beta$ in Algorithm \ref{alg:FSAF}) & 0.001 \\
    Agent/demo replay buffer size & 1000 \\
    Discount factor $\gamma$ & 0.98 \\
    Number of DQN particles & 5 \\
    Total sampling budget & 100\\
    Cardinality of the Sobol grid & 200\\
    
  \bottomrule
\end{tabular}
\end{table}

\section{Experiment Details}
\label{app:exp details}

\subsection{Experiment Details of Figure \ref{Fig:overfitting}}
\label{app:exp details:overfitting}
Recall that Figure \ref{Fig:overfitting} illustrates the overfitting issue of DQN+MAML.
Regarding the DQN used for Figure \ref{Fig:overfitting}, we use the same design, network architecture, and training tasks as those described in Appendix \ref{app:training details}.
Regarding the vanilla MAML for Figure \ref{Fig:overfitting}, we use the squared TD error in (\ref{eq:C theta}) as the loss function for both the lower-level and upper-level updates.
For both training and the fast adaptation during testing, we apply 5-shot adaptation (i.e., 5 black-box functions are used for adaptation) and set the number of few-shot gradient updates to be 5.
In Figures \ref{Fig:1particle_EI.0}-\ref{Fig:1particle_EI.2}, we report the empirical average simple regret as well as the empirical standard deviation over 100 independent trials.

\subsection{Experiment Details of Figure \ref{Fig:blackbox}}
\label{app:exp details:blackbox}
Recall that Figure \ref{Fig:blackbox} shows the simple regret performance of the AFs under various standard optimization benchmark functions.
The input domain of each benchmark function is a hypercube in the form of $[-x_{\lim},x_{\lim}]^{d}$ (e.g., $x_{\lim}=5$ and $d=2$ for the Ackley function).
Moreover, as the function values of these benchmark functions can be one or more orders of magnitude different from each other, for ease of comparison, we scale all the values of the functions to the range $[-2,2]$.
For the posterior inference required by all the AFs, we use a GP with an RBF kernel as the surrogate model. 
For each benchmark function, the lengthscale parameter of the RBF kernel is estimated via marginal likelihood maximization, which is a commonly-used Bayesian model selection framework.

\textbf{Validation datasets and testing datasets.}
As mentioned in Section \ref{section:exp}, we construct the validation datasets (mainly for the few-shot adaptation of FSAF and the fine-tuning of MetaBO-T) and the testing datasets by applying random translations and re-scalings to the $x$ and the $y$ values, respectively.
Specifically, the amount of translation added to each dimension of $x$ is selected from the range $[-0.1 x_{\lim},0.1 x_{\lim}]$ uniformly at random.
Similarly, the re-scaling factor applied to each $y$ value is chosen uniformly at random from the range $[0.9,1.1]$.



\textbf{Maximization of the AFs.} 
Recall that the input domain of each optimization benchmark function is a hypercube.
To address the continuous input domains and achieve a fair comparison between FSAF and MetaBO, we leverage the hierarchical gridding method similar to that in \cite{volpp2020metalearning} for the maximization procedure of the AFs. 
Specifically, we first construct a coarse Sobol grid of $N_{\text{coarse}}$ points that span over the entire domain and then evaluate the AF on this grid.
Next, we find the $N_{\text{m}}$ maximal evaluations on this coarse grid and build a finer local Sobol grid of $N_{\text{local}}$ points for each of the $N_{\text{m}}$ maximal points.
Then, the maximum of the AF is approximated by the maximum value among these $N_{\text{m}}N_{\text{local}}$ AF evaluations. 
To finish the testing process within a reasonable amount of time, we choose $N_{\text{coarse}}=2000$, $N_{\text{m}}=10$, and $N_{\text{local}}=1000$.

\subsection{Experiment Details of Figure \ref{Fig:real}}
\label{app:exp details:real}

Recall that Figure \ref{Fig:real} demonstrates the regret performance under the test functions obtained from a variety of real-world datasets.
Below we describe these real-world datasets in more detail.
\begin{itemize}[leftmargin=*]
\item \textbf{XGBoost Hyperparameter Optimization.}
We use the HPOBench dataset\footnote{Created by AutoML and available at \url{https://github.com/automl/HPOBench}.} for the hyperparameter optimization for the XGBoost algorithm.
Specifically, we use the pre-computed results of XGBoost with six tunable hyperparameters (e.g., learning rate, $L_1$ and $L_2$ regularization terms, and subsampling ratios) on 48 classification datasets, each of which is associated with 1000 randomly selected hyperparameter configurations.
Hence, these 48 subsets of pre-computed results naturally provide 48 black-box test functions for BO. 
In the 1-shot setting, we use 1 out of the 48 black-box functions for fast adaptation of FSAF and finding the lengthscale parameter of the GP surrogate model via marginal likelihood maximization.
The remaining 47 black-box functions are used only for testing.



\item \textbf{Electrical Grid Stability Dataset.}
This dataset corresponds to an augmented version of the \textit{Electrical Grid Stability Simulated Dataset}\footnote{Created by Vadim Arzamasov (Karlsruher Institut für Technologie, Karlsruhe, Germany) and available at \url{https://www.kaggle.com/pcbreviglieri/smart-grid-stability}.} \cite{arzamasov2018towards}. 
This dataset records total 12 features, such as the reaction times of the producer and the consumer, power balance, and price elasticity coefficient, and the objective is to maximize the grid stability. 
This dataset contains 60000 parameter configurations, and we divide them into 39 testing sets and 1 validation set.
The validation set is used for both few-shot adaptation of FSAF as well as finding the lengthscale parameter of the GP surrogate model for posterior inference.

\item \textbf{Air Quality Prediction.}
We use the meteorological monitoring data of air quality collected in northern Taiwan in 2015\footnote{Created by Environmental Protection Administration, Executive Yuan, R.O.C. (Taiwan), available at \url{https://airtw.epa.gov.tw/ENG/default.aspx}}. 
We select 14 air quality features, such as the amount of Sulfur dioxide, Carbon monoxide, and ozone, and set the amount of $\text{PM}_{2.5}$ as the objective function.
After cleaning up all the NaN entries and missing data entries, we get about 73000 parameter configurations and split them into 29 testing sets and 1 validation set.
Again, the validation set is used for both few-shot adaptation of FSAF as well as finding the lengthscale parameter of the GP surrogate model for posterior inference.

\item \textbf{Oil Well Dataset.}
We use the \textit{NYS Oil, Gas, Other Regulated Wells datasets}\footnote{Hosted by State of New York, available at \url{https://www.kaggle.com/new-york-state/nys-oil,-gas,-other-regulated-wells?select=oil-gas-other-regulated-wells-beginning-1860.csv}.} to find the deepest drilled depth among the oil wells.
For each data entry, we use the longitude and latitude of both the surface as well as the bottom of the oil well as the input features.
This dataset contains about 41000 parameter configurations, and we divide the dataset into 29 subsets of testing data and 1 set of validation data for few-shot adaptation of FSAF and finding the lengthscale parameter of the GP surrogate model for posterior inference.

\item\textbf{Asteroid Dataset.}
This dataset\footnote{Provided by the paper titled ``Prediction of Asteroid Diameter with the help of Multi-layer Perceptron Regressor" in \textit{International Conference on Computer Science, Industrial Electronics} in 2019 \cite{Asteroiddata} and available at \url{https://www.kaggle.com/basu369victor/prediction-of-asteroid-diameter}.} contains a variety features of the asteroids, and our goal is to find the maximum asteroid diameter based on all of its 12 numerical features.
Since the range of the asteroid diameters is too large, we apply the commonly-used input warping technique in BO and use the logarithm of the diameter values as the objective.
After cleaning up all the NaN entries and missing entries, we get about 136000 parameter configurations and split them into 39 testing sets and 1 validation set.

\end{itemize}

Regarding the maximization of AFs, since the input domains of the real-world test functions are all discrete,
the maximum value of each AF can be found exactly and therefore the hierarchical gridding procedure is not required in this case.

\section{Pseudo Code and Architecture of the FSAF Training Algorithm}
\label{app:pseudo code}
\begin{figure}[!htbp]
  \centering
  \includegraphics[width=6cm]{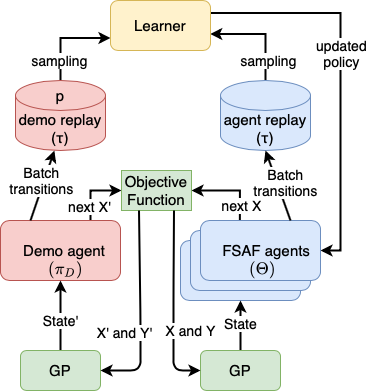}\\
  \caption{Architecture of the proposed FSAF.}
  \label{Fig:architecture}
\end{figure}

For completeness, we provide the pseudo code of the training algorithm of FSAF in the following Algorithm \ref{alg:FSAF}. 
Figure \ref{Fig:architecture} further illustrates the demo policy and the replay buffers used in FSAF.

\begin{algorithm}[!htbp]
\caption{FSAF Training Algorithm}
\label{alg:FSAF}
\begin{algorithmic}[1]
    \State {\bfseries Initialize:} $N$ instances of Q-networks with parameters $\Theta = \{\theta^{(n)}\}^{N}_{n=1}$, demo policy $\pi_{\text{D}}$, a collection of candidate tasks $\cT$, and the parameters $K,S$ for computing the meta-loss\;
    \For{each iteration $i = 0,1,\cdots$}
    \State Sample a batch of tasks $\mathcal{T}_i$ from $\cT$ \label{pseudo:sample tasks}
        \For{each task $\tau\in \mathcal{T}_{i}$}
            \State Generate black-box functions of task $\tau$ from GP and set $\Theta_{\tau,0}=\Theta$
            \For{$k=0,\cdots,K-1$}
            \State Collect trajectories $\{(s_0,a_1,r_1,\cdots)\}$ using $\pi_{\Theta_{\tau,k}}$ and store transitions in $\cR_Q$
            \State Collect trajectories $\{(s_0,a_1,r_1,\cdots)\}$ using $\pi_{\text{D}}$ and store transitions in $\cR_{\text{D}}$ 
            \State Sample a mini-batch of transitions $\cD_{\tau,k}^{\text{tr}}$ from the replay buffers $\cR_Q, \cR_{\text{D}}$\;
            \State $\Theta_{\tau,k+1}=\cM_{\text{SVGD}}(\Theta_{\tau,k}, \cD_{\tau,k}^{\text{tr}})$
            \EndFor
            \State Set $\Theta^{*}_{\tau,0}=\Theta_{\tau,K}$
            \For{$s=0,\cdots, S-1$}
            \State Sample a mini-batch of transitions $\cD_{\tau,s}^{\text{val}}$ from the replay buffers $\cR_Q, \cR_{\text{D}}$
            \State $\Theta_{\tau,s+1}^{*}=\cM_{\text{SVGD}}(\Theta_{\tau,s}^{*}, \cD_{\tau,s}^{\text{val}})$
            \EndFor
        \EndFor
        \State ${\Theta} \leftarrow {\Theta} - \beta\nabla_{\Theta} \big(\sum_{\tau\in \mathcal{T}_i}\cL_{\text{meta}}(\Theta;\tau)\big)$\;
    \EndFor
\end{algorithmic}
\end{algorithm}

\section{Additional Experimental Results}
\label{app:results}
In this section, we provide additional experimental results regarding the synthetic GP test functions, the combination of FSAF and FSBO, and the percentiles associated with Figure \ref{Fig:real} in the main text.

\begin{figure}[!htbp]
\centering 
\subfigure[]{
\label{Fig.kernel.0}
\includegraphics[width=0.255\textwidth]{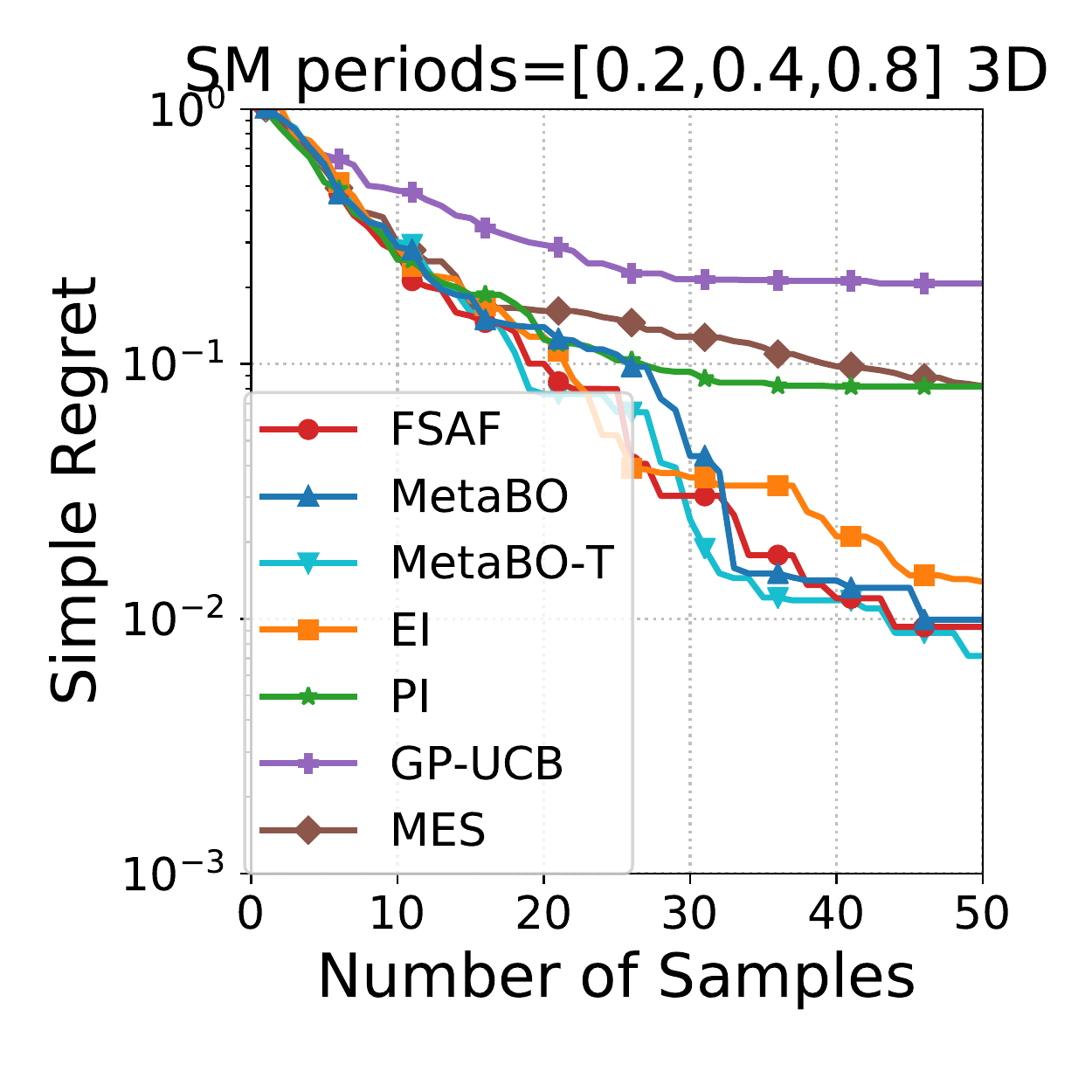}}
\hspace{-4mm}
\subfigure[]{
\label{Fig.kernel.1}
\includegraphics[width=0.255\textwidth]{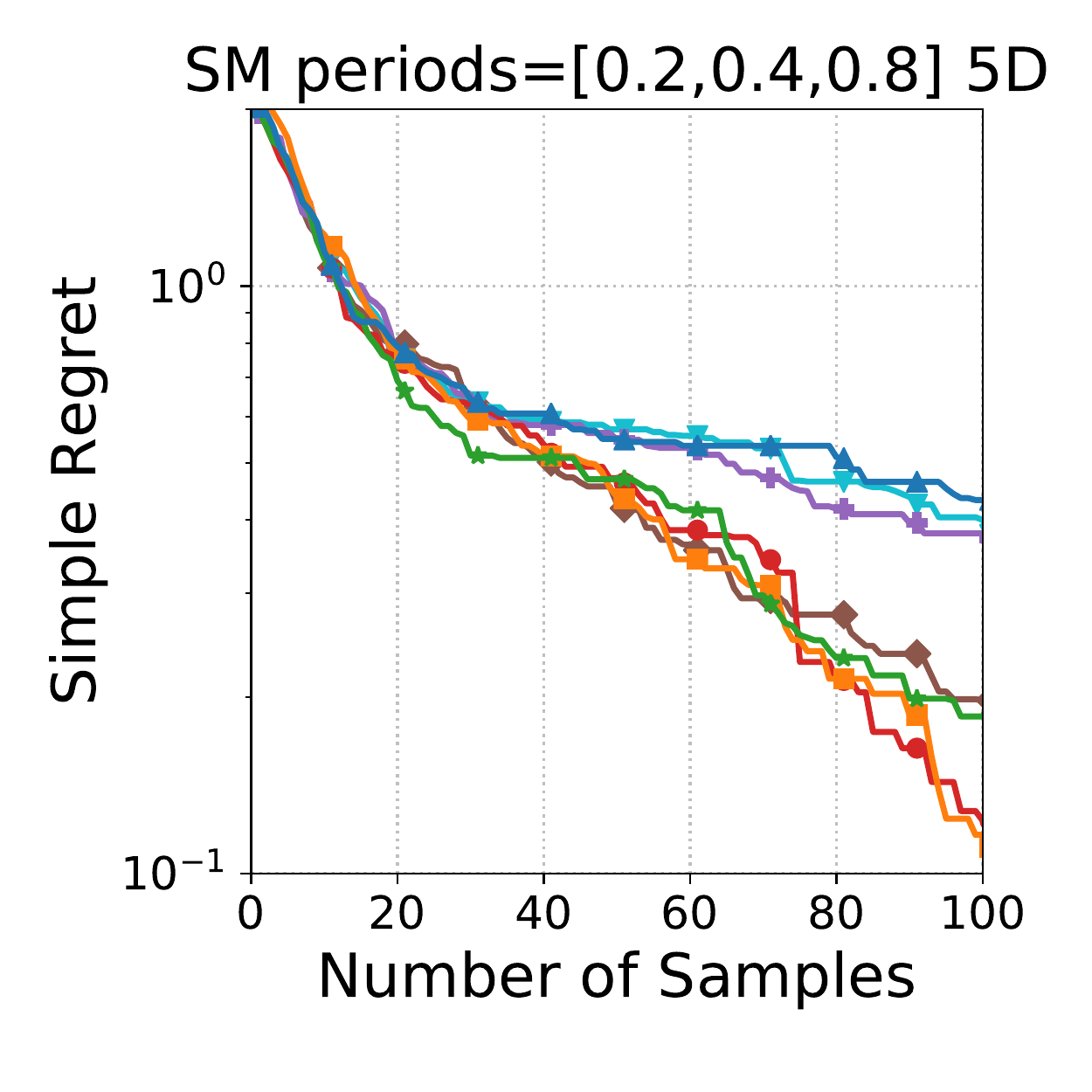}}
\hspace{-4mm}
\subfigure[]{
\label{Fig.kernel.2}
\includegraphics[width=0.255\textwidth]{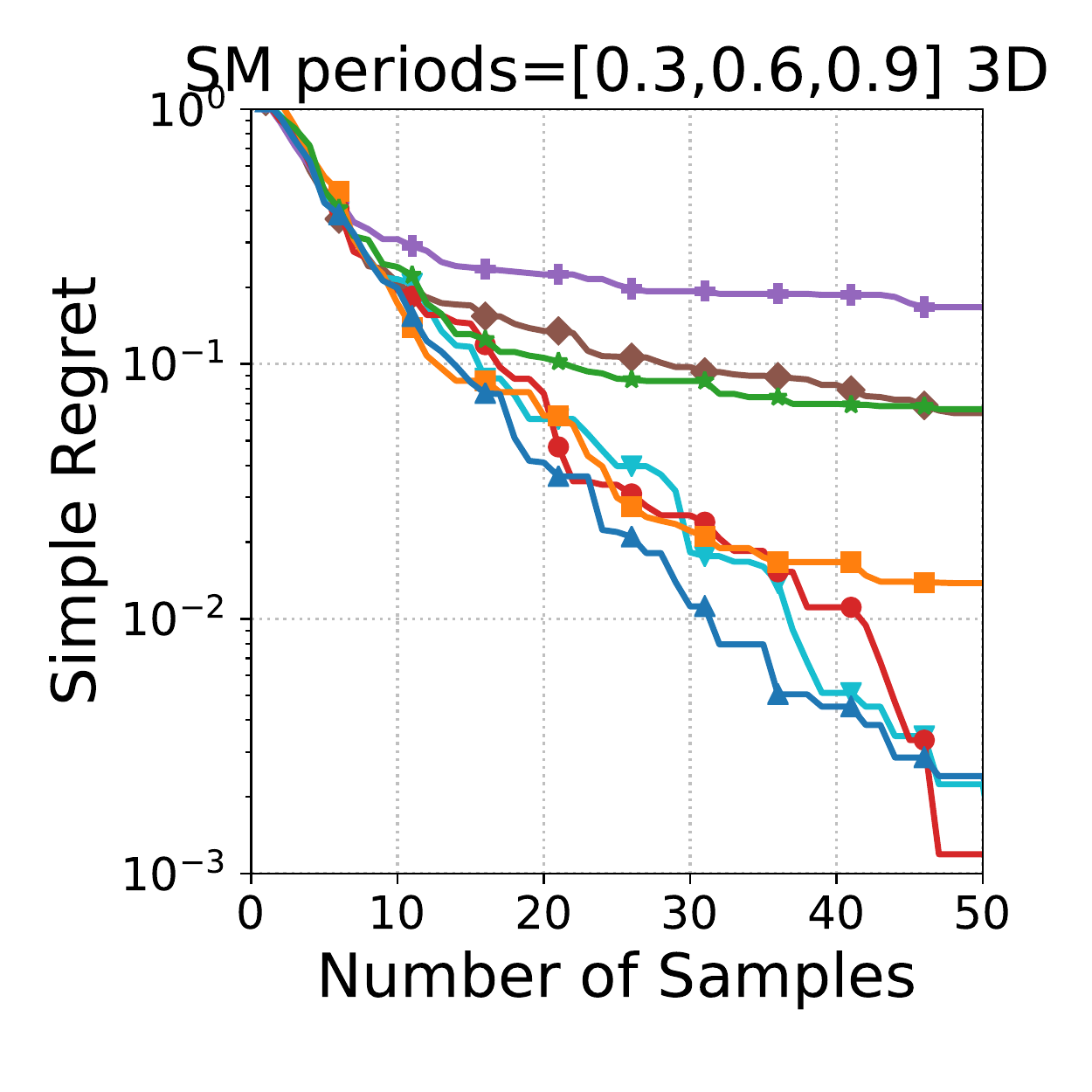}}
\hspace{-4mm}
\subfigure[]{
\label{Fig.kernel.3}
\includegraphics[width=0.255\textwidth]{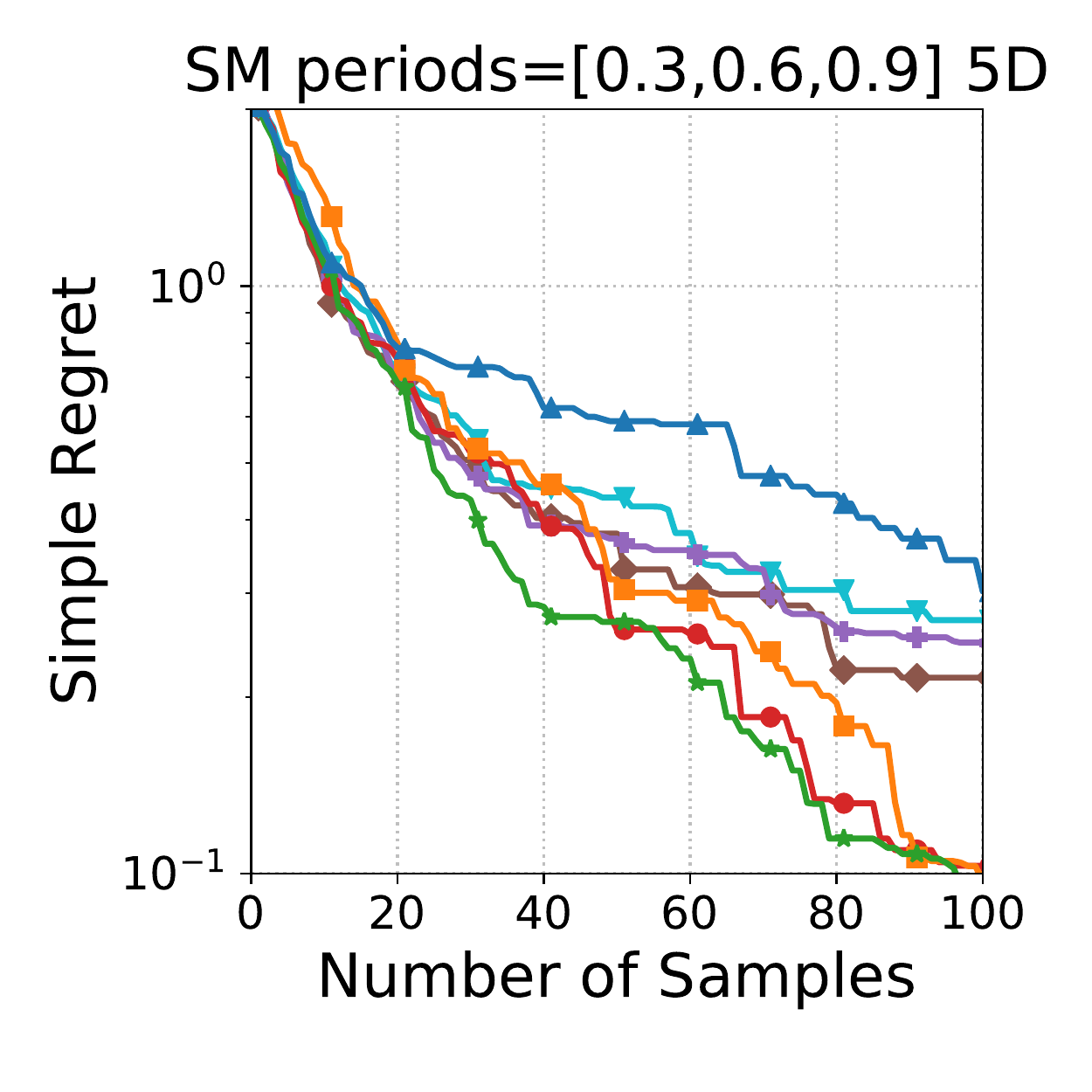}}
\subfigure[]{
\label{Fig.kernel.4}
\includegraphics[width=0.255\textwidth]{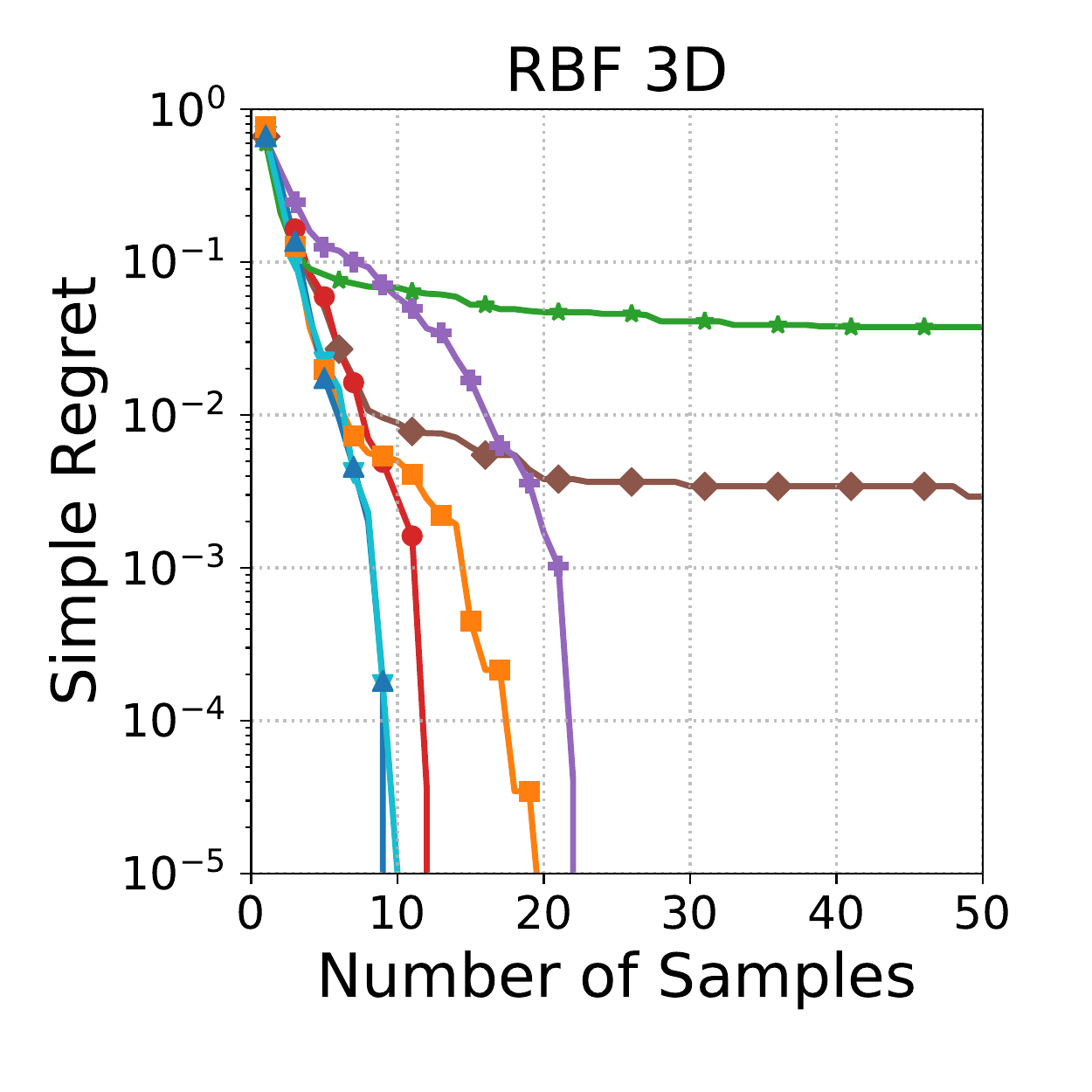}}
\hspace{-4mm}
\subfigure[]{
\label{Fig.kernel.5}
\includegraphics[width=0.255\textwidth]{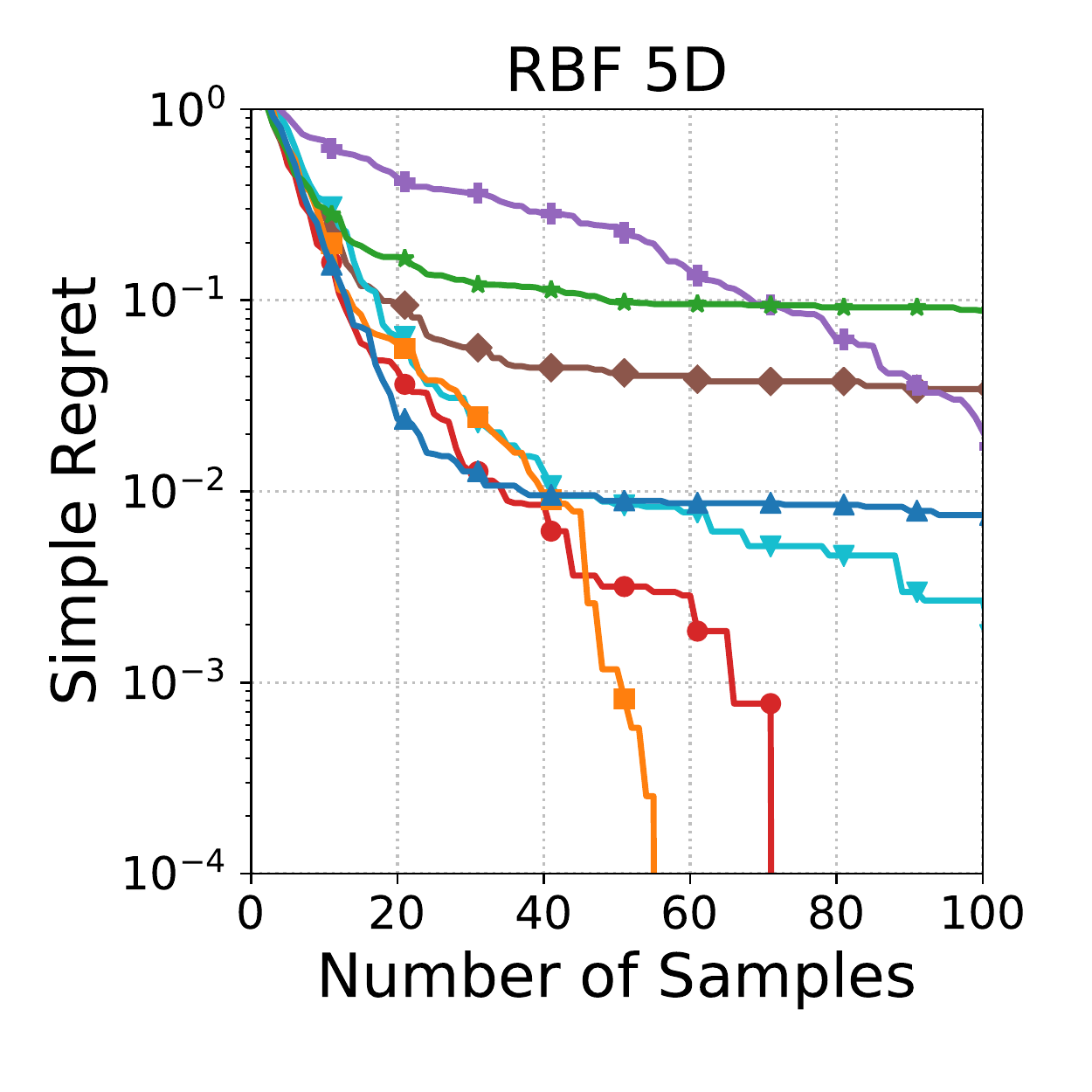}}
\hspace{-4mm}
\subfigure[]{
\label{Fig.kernel.6}
\includegraphics[width=0.255\textwidth]{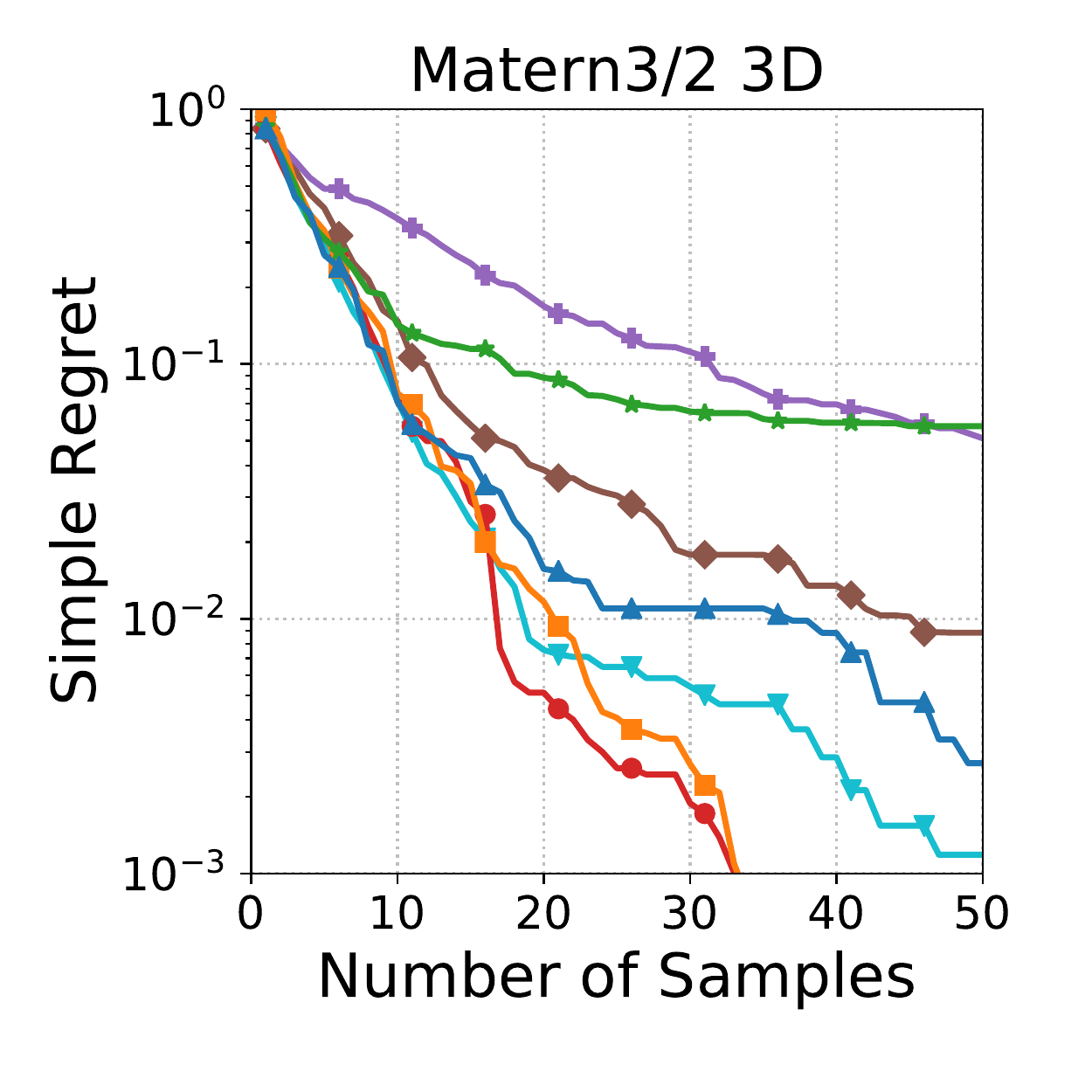}}
\hspace{-4mm}
\subfigure[]{
\label{Fig.kernel.7}
\includegraphics[width=0.255\textwidth]{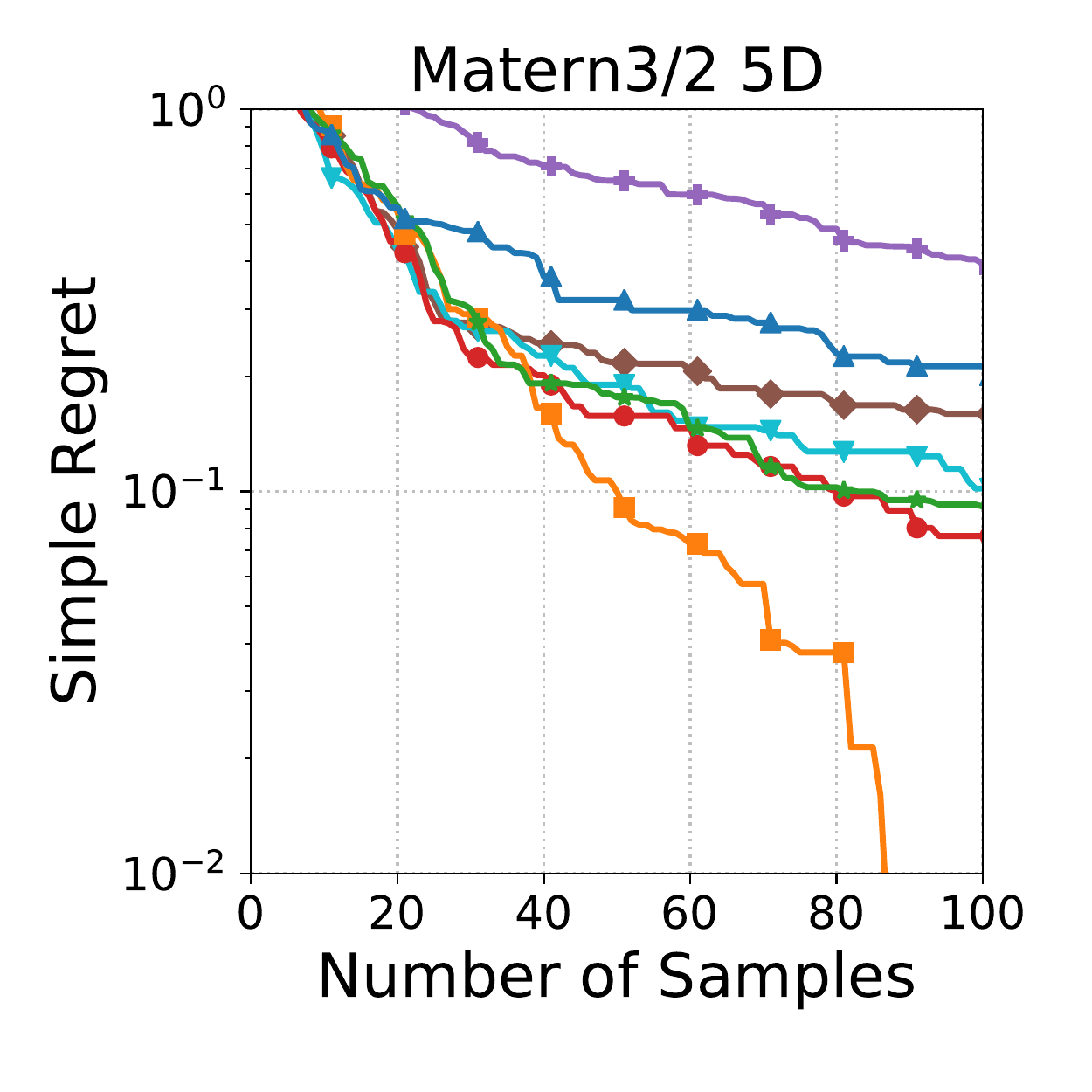}}
    \caption{Median simple regrets of FSAF and other benchmark methods under synthetic GP test functions. Other percentiles are provided in Table \ref{table:percentiles GP functions} for visual clarity.}
  \label{Fig.kernel}
\end{figure}

\subsection{Synthetic GP Test Functions}
\label{app:results:GP}
To further validate the effectiveness of FSAF, we evaluate FSAF and the benchmark AFs on GP functions drawn from various kernels.
Specifically, to generate the validation and testing datasets, we consider four different types of GP kernels for both the input domains of $[0,1]^3$ and $[0,1]^5$, including: (i) Spectral mixture kernel with three Gaussian components of periods $0.2$, $0.4$, and $0.8$ as well as a lengthscale range $[0.5,0.55]$; (ii) Spectral mixture kernel with three Gaussian components of periods $0.3$, $0.6$, and $0.9$ as well as a lengthscale range $[0.5,0.55]$ (iii) RBF kernel with a lengthscale range $[0.5, 0.55]$; (iv) Mat\`ern kernel with a lengthscale range $[0.5, 0.55]$.
Notably, all the above kernels have never been seen by FSAF during training.
To address the continuous input domains, we use the same hierarchical gridding method as described in Appendix \ref{app:exp details:blackbox} to maximize the AFs.
Similar to the case of Figure \ref{Fig:blackbox}, we consider 5-shot adaptation for FSAF and use the same amount of meta-data for MetaBO-T.

Figure \ref{Fig.kernel} shows the simple regrets of all the AFs under eight different types of GP functions.
Again, we can see that FSAF is constantly among the best of all the methods under most of the test functions.
We also observe that MetaBO and MetaBO-T perform quite well in the 3-dimensional tasks.
We conjecture that this is due to the fact that the pre-trained MetaBO model was originally trained on 3-dimensional GP functions, as described in \cite{volpp2020metalearning}. 
However, MetaBO and MetaBO-T do not adapt well to the 5-dimensional GP functions, as shown in Figures \ref{Fig.kernel.1}, \ref{Fig.kernel.3}, \ref{Fig.kernel.5}, and \ref{Fig.kernel.7}.
By contrast, despite that FSAF is also trained on 3-dimensional GP functions, FSAF adapts more effectively to GP functions of a higher input dimension, as shown in \ref{Fig.kernel.1}, \ref{Fig.kernel.3}, \ref{Fig.kernel.5}.
On the other hand, Figure \ref{Fig.kernel.7} shows that FSAF does not catch up well with EI after 40 samples under the 5-dimensional Mat\`ern-2/3 functions. 
We conjecture that this is due to the fact that 5-dimensional Mat\`ern-2/3 functions have even more local variations that could not be captured well by only a small amount of meta-data.
Moreover, among the benchmark methods, the best-performing AF still varies under different types of functions. 
This again corroborates the commonly-seen phenomenon and our motivation.


\begin{table*}[!htbp]
\centering
\linespread{0.1}
\renewcommand\arraystretch{0.1}
  \caption{The percentiles of simple regrets over 100 independent trials for all the AFs under synthetic GP test functions. For 3-dimensional kernel functions, we report the percentiles measured at steps 25 and 50. For 5-dimensional kernel functions, we report the percentiles measured at steps 50 and 100.
  All the values displayed here are scaled by 100 for more compact notations.
  The best of each percentile under each type of functions is highlighted in \underline{\bf{bold and underline}}.}
  \label{tab:exp_value:kernel}

  \begin{tabular}{lllllllllll}
\midrule
\midrule
AF &  P  &  SM periods= & SM periods=& RBF 3D & Mat\`ern-3/2 3D \\
 &    &  [0.2,0.4,0.8] 3D & [0.3,0.6,0.9] 3D &  &  \\
\midrule
FSAF & 25\%   &  0.03, \underline{\bf{0.00}} & 0.01, \underline{\bf{0.00}} & \underline{\bf{0.00}}, \underline{\bf{0.00}} & \underline{\bf{0.00}}, \underline{\bf{0.00}}  \\ 
 & 50\%   &  7.80, 0.91 & 3.28, \underline{\bf{0.12}} & \underline{\bf{0.00}}, \underline{\bf{0.00}} & \underline{\bf{0.25}}, 0.03  \\ 
 & 75\%   &  41.62, 38.81 & \underline{\bf{24.78}}, 16.74 & \underline{\bf{0.00}}, \underline{\bf{0.00}} & \underline{\bf{3.87}}, 0.54  \\ 
 & 90\%   &  \underline{\bf{70.97}}, 67.15 & 64.65, 62.63 & \underline{\bf{0.55}}, 0.52 & 16.75, 9.67  \\ 
\midrule
MetaBO & 25\%   &  \underline{\bf{0.00}}, \underline{\bf{0.00}} & \underline{\bf{0.00}}, \underline{\bf{0.00}} & \underline{\bf{0.00}}, \underline{\bf{0.00}} & 0.02, \underline{\bf{0.00}}  \\ 
 & 50\%   &  10.88, 0.99 & \underline{\bf{2.19}}, 0.24 & \underline{\bf{0.00}}, \underline{\bf{0.00}} & 1.10, 0.27  \\ 
 & 75\%   &  46.07, 29.76 & 28.96, 17.16 & 0.19, 0.13 & 11.51, 5.77  \\ 
 & 90\%   &  81.47, 75.46 & 72.62, 61.25 & 0.92, 0.84 & 25.12, 22.52  \\ 
\midrule
MetaBO-T & 25\%   &  0.09, \underline{\bf{0.00}} & 0.15, \underline{\bf{0.00}} & \underline{\bf{0.00}}, \underline{\bf{0.00}} & \underline{\bf{0.00}}, \underline{\bf{0.00}}  \\ 
 & 50\%   &  6.47, \underline{\bf{0.72}} & 3.97, 0.22 & \underline{\bf{0.00}}, \underline{\bf{0.00}} & 0.65, 0.12  \\ 
 & 75\%   &  46.25, 16.45 & 41.52, 18.71 & 0.25, 0.02 & 7.04, 4.93  \\ 
 & 90\%   &  82.98, 51.48 & 78.32, 59.29 & 1.41, 0.84 & 24.48, 23.85  \\ 
\midrule
EI & 25\%   &  1.02, 0.50 & 0.93, 0.34 & \underline{\bf{0.00}}, \underline{\bf{0.00}} & 0.00, \underline{\bf{0.00}}  \\ 
 & 50\%   &  \underline{\bf{4.57}}, 1.22 & 2.60, 1.20 & \underline{\bf{0.00}}, \underline{\bf{0.00}} & 0.36, \underline{\bf{0.00}}  \\ 
 & 75\%   &  49.63, \underline{\bf{6.27}} & 29.60, \underline{\bf{4.19}} & 0.33, 0.02 & 5.31, \underline{\bf{0.43}}  \\ 
 & 90\%   &  88.47, 48.41 & 83.11, 38.15 & 1.22, 0.99 & 14.93, \underline{\bf{1.38}}  \\ 
\midrule
PI & 25\%   &  5.93, 3.96 & 4.55, 3.54 & 1.06, 1.01 & 3.94, 3.21  \\ 
 & 50\%   &  10.34, 8.16 & 8.76, 6.65 & 4.59, 3.76 & 7.26, 5.70  \\ 
 & 75\%   &  40.42, 14.74 & 28.49, 12.45 & 8.04, 6.71 & 12.61, 9.13  \\ 
 & 90\%   &  79.30, 42.57 & 74.08, 33.11 & 11.45, 8.93 & 19.77, 13.93  \\ 
\midrule
GP-UCB & 25\%   &  13.04, 11.71 & 11.18, 10.80 & \underline{\bf{0.00}}, \underline{\bf{0.00}} & 6.06, 1.79  \\ 
 & 50\%   &  23.83, 20.70 & 20.46, 16.71 & \underline{\bf{0.00}}, \underline{\bf{0.00}} & 13.20, 5.11  \\ 
 & 75\%   &  47.45, 33.06 & 35.04, 27.53 & 0.21, \underline{\bf{0.00}} & 26.47, 9.42  \\ 
 & 90\%   &  73.17, 47.62 & \underline{\bf{59.32}}, 34.97 & 1.11, \underline{\bf{0.44}} & 41.64, 16.09  \\ 
\midrule
MES & 25\%   &  5.70, 3.70 & 5.72, 2.66 & 0.03, 0.02 & 0.69, 0.21  \\ 
 & 50\%   &  14.97, 8.18 & 10.69, 6.42 & 0.37, 0.29 & 3.05, 0.88  \\ 
 & 75\%   &  \underline{\bf{37.36}}, 16.17 & 28.17, 10.70 & 1.81, 1.18 & 7.59, 2.88  \\ 
 & 90\%   &  75.31, \underline{\bf{36.15}} & 67.86, \underline{\bf{17.39}} & 5.03, 3.75 & \underline{\bf{11.78}}, 6.27  \\ 
 \midrule
 \midrule
 AF &  P  &  SM periods= & SM periods= & RBF 5D & Mat\`ern-3/2 5D \\
   &    &  [0.2,0.4,0.8] 5D & [0.3,0.6,0.9] 5D &  &  \\
\midrule
FSAF & 25\%   &  8.10, 0.88 & 5.51, 1.75 & \underline{\bf{0.00}}, \underline{\bf{0.00}} & 2.67, \underline{\bf{0.00}}  \\ 
 & 50\%   &  46.04, 12.03 & \underline{\bf{25.44}}, 10.08 & 0.31, \underline{\bf{0.00}} & 15.40, 7.47  \\ 
 & 75\%   &  \underline{\bf{67.89}}, \underline{\bf{46.69}} & 74.85, 41.58 & \underline{\bf{2.67}}, \underline{\bf{1.32}} & 45.85, 23.46  \\ 
 & 90\%   &  \underline{\bf{104.46}}, \underline{\bf{71.09}} & \underline{\bf{98.25}}, 78.89 & \underline{\bf{5.52}}, \underline{\bf{4.00}} & 72.67, 52.40  \\ 
\midrule
MetaBO & 25\%   &  22.55, 13.76 & 14.32, 6.84 & \underline{\bf{0.00}}, \underline{\bf{0.00}} & 9.18, 3.44  \\ 
 & 50\%   &  54.93, 43.24 & 58.90, 30.21 & 0.89, 0.75 & 31.69, 21.27  \\ 
 & 75\%   &  80.30, 69.47 & 101.54, 78.60 & 7.02, 4.92 & 69.30, 48.99  \\ 
 & 90\%   &  128.87, 96.96 & 156.15, 124.84 & 30.42, 17.31 & 86.73, 80.38  \\ 
\midrule
MetaBO-T & 25\%   &  23.23, 15.84 & 12.31, 5.69 & \underline{\bf{0.00}}, \underline{\bf{0.00}} & 4.82, 0.61  \\ 
 & 50\%   &  57.05, 39.99 & 43.67, 27.00 & 0.85, 0.27 & 19.01, 10.17  \\ 
 & 75\%   &  88.13, 69.88 & 75.39, 65.34 & 4.58, 3.12 & 55.67, 28.54  \\ 
 & 90\%   &  123.23, 91.40 & 107.24, 92.93 & 23.03, 8.77 & 77.27, 55.63  \\ 
\midrule
EI & 25\%   &  \underline{\bf{6.35}}, \underline{\bf{0.00}} & \underline{\bf{3.12}}, \underline{\bf{0.00}} & \underline{\bf{0.00}}, \underline{\bf{0.00}} & \underline{\bf{0.51}}, \underline{\bf{0.00}}  \\ 
 & 50\%   &  \underline{\bf{38.09}}, \underline{\bf{10.12}} & 27.56, \underline{\bf{8.42}} & \underline{\bf{0.10}}, \underline{\bf{0.00}} & \underline{\bf{8.68}}, \underline{\bf{0.00}}  \\ 
 & 75\%   &  79.77, 51.62 & 86.52, 57.03 & 3.39, 1.59 & \underline{\bf{28.38}}, \underline{\bf{7.36}}  \\ 
 & 90\%   &  143.68, 92.64 & 123.61, 102.67 & 8.09, 5.56 & 68.32, 30.38  \\ 
\midrule
PI & 25\%   &  12.74, 4.36 & 6.76, 2.45 & 5.81, 4.80 & 8.54, 4.42  \\ 
 & 50\%   &  46.90, 18.52 & 26.81, 9.55 & 9.76, 8.80 & 17.61, 9.13  \\ 
 & 75\%   &  76.23, 53.18 & 74.40, \underline{\bf{40.77}} & 14.47, 12.75 & 30.70, 17.44  \\ 
 & 90\%   &  110.59, 89.79 & 107.18, \underline{\bf{70.79}} & 20.94, 16.77 & \underline{\bf{53.25}}, \underline{\bf{27.73}}  \\ 
\midrule
GP-UCB & 25\%   &  24.86, 17.37 & 21.25, 15.13 & 12.79, \underline{\bf{0.00}} & 41.12, 19.84  \\ 
 & 50\%   &  54.77, 37.94 & 37.17, 24.71 & 24.31, 2.06 & 64.86, 39.35  \\ 
 & 75\%   &  88.69, 64.84 & \underline{\bf{71.08}}, 46.74 & 37.09, 7.72 & 84.72, 56.06  \\ 
 & 90\%   &  127.81, 92.97 & 105.94, 71.43 & 46.44, 13.06 & 111.00, 68.07  \\ 
\midrule
MES & 25\%   &  19.37, 4.40 & 11.08, 5.16 & 0.84, 0.43 & 9.31, 4.46  \\ 
 & 50\%   &  42.62, 19.73 & 37.91, 21.55 & 4.17, 3.43 & 21.72, 15.95  \\ 
 & 75\%   &  74.29, 46.93 & 73.01, 50.09 & 11.76, 9.91 & 40.49, 28.11  \\ 
 & 90\%   &  109.83, 72.50 & 98.32, 85.27 & 18.17, 16.03 & 67.14, 59.04  \\ 
  \end{tabular}
  \label{table:percentiles GP functions}
\end{table*}

\vspace{-2mm}
\begin{figure*}[!htbp]
\centering 
  \begin{minipage}{0.65\textwidth}
\subfigure[]{
\label{Fig.fsbo.0}
\includegraphics[width=0.5\textwidth]{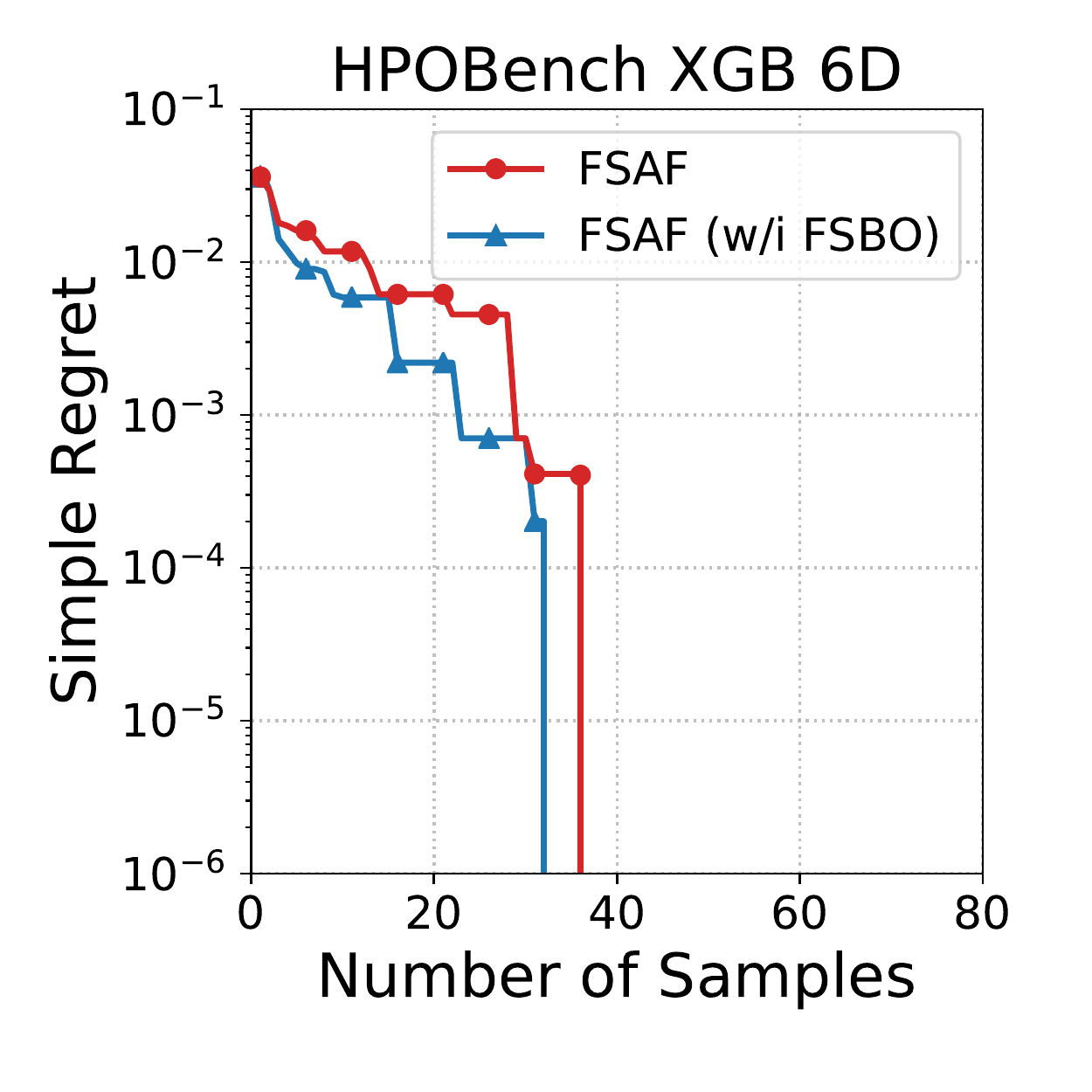}}
\hspace{-5mm}
\subfigure[]{
\label{Fig.fsbo.1}
\includegraphics[width=0.5\textwidth]{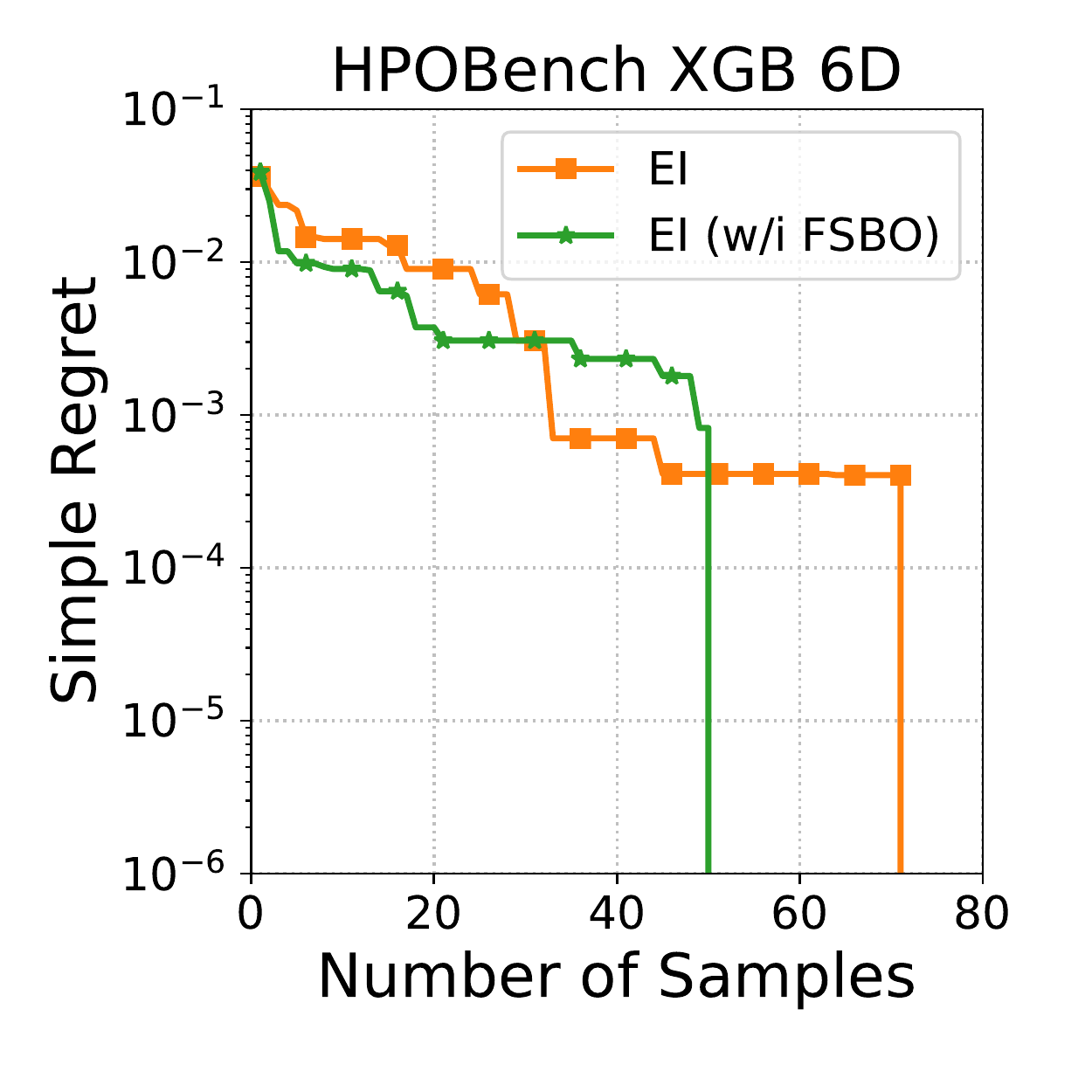}}
    \caption{Median simple regrets of FSAF and EI, with and without the integration with FSBO. Other percentiles are in Table \ref{tab:exp_fsbo}.}
    \label{Fig.growth2}
    \label{Fig.fsbo}
  \end{minipage}
\end{figure*}
\vspace{-2mm}

\subsection{Combining FSAF with FSBO}
\label{app:results:FSBO}
Recall from Remark \ref{remark:FSAF and FSBO} that \cite{wistuba2021fewshot} proposes FSBO, which leverages meta-data to fine-tune the initialization of the GP kernel parameters for the off-the-shelf AFs.
As FSBO can provide better GP kernel parameters for posterior inference than the conventional approaches (e.g., marginal likelihood maximization), it appears feasible to let FSAF and FSBO complement each other and even combine other off-the-shelf AFs with FSBO for better regret performance.
In this section, we provide experimental results to validate the above argument.
Since \cite{wistuba2021fewshot} did not release their source code, we need to re-implement FSBO by ourselves\footnote{We contacted the authors of \cite{wistuba2021fewshot} via email and got the reply that they still needed a bit more time before making the code publicly available. 
Moreover, we also confirmed the important design choices with the authors to better reproduce FSBO.}.
Specifically, for the meta-learning part of FSBO, we leverage Reptile \cite{nichol2018first} with a spectral mixture kernel as a lightweight variant of \cite{patacchiola2020bayesian}.
Moreover, we use cosine annealing outer-loop learning rate from $10^{-3}$ to $10^{-5}$ and set the inner-loop learning rate to be $10^{-2}$.
The deep kernel is represented by a neural network with two hidden layers (with 128 hidden units per layer), and the degree of few-shot deep kernel learning is configured to be 4.
The spectral mixture kernel is configured to have 10 components.
We test (i) the combination of FSAF and FSBO as well as (ii) the combination of EI and FSBO on the XGBoost hyperparameter optimization task described in Appendix \ref{app:exp details:real}.
As FSBO requires some data for obtaining an initial model, we use 5 out of the 48 subsets of the HPOBench dataset to obtain an initial FSBO model and evaluate the regret performance on the other 43 subsets. 
From Figure \ref{Fig.fsbo} and Table \ref{tab:exp_fsbo}, we see that FSBO can slightly improve the regret performance of the two AFs.


\begin{table*}[!htbp]
\centering
\renewcommand\arraystretch{0.1}
  \caption{The percentiles of simple regrets under the HPOBench dataset at steps 30 and 60. All the values displayed here are scaled by 100 for more compact notations.
  The best of each percentile under each type of functions is highlighted in \underline{\bf{bold and underline}}.}
  \vspace{2mm}
  \label{tab:exp_fsbo}
  \begin{tabular}{llll}

Dataset &  Percentile  &  FSAF & FSAF (w/i FSBO) \\
\midrule
HPOBench XGB & 25\%   &  \underline{\bf{0.00}}, \underline{\bf{0.00}} & \underline{\bf{0.00}}, \underline{\bf{0.00}}  \\ 
 & 50\%   &  \underline{\bf{0.07}}, \underline{\bf{0.00}} & \underline{\bf{0.07}}, \underline{\bf{0.00}}  \\ 
 & 75\%   &  1.79, \underline{\bf{0.90}} & \underline{\bf{0.99}}, 0.99  \\ 
 & 90\%   &  4.51, \underline{\bf{3.31}} & \underline{\bf{3.37}}, 3.37  \\ 
\midrule
\midrule
Dataset & Percentile  &  EI & EI (w/i FSBO) \\
\midrule
HPOBench XGB & 25\%   &  \underline{\bf{0.00}}, \underline{\bf{0.00}} & \underline{\bf{0.00}}, \underline{\bf{0.00}}  \\ 
 & 50\%   &  \underline{\bf{0.31}}, 0.04 & \underline{\bf{0.31}}, \underline{\bf{0.00}}  \\ 
 & 75\%   &  2.50, 1.99 & \underline{\bf{1.57}}, \underline{\bf{1.13}}  \\ 
 & 90\%   &  6.36, 6.27 & \underline{\bf{5.18}}, \underline{\bf{4.62}}  \\ 
 
 \end{tabular}
 \label{table:percentiles of FSBO}
\end{table*}

\subsection{Additional Percentiles for Figure \ref{Fig:real}}
In this section, we provide more detailed percentiles associated with the experiments of Figure \ref{Fig:real} in Table \ref{table:percentiles of real datasets}. 
We can see that FSAF is still among the best in terms of either lower or higher percentiles under most of the real-world test functions.

\begin{table*}[!htbp]
\centering
\linespread{0.1}
\renewcommand\arraystretch{0.1}
  \caption{The percentiles of simple regrets for all the AFs under different real-world test functions. 
  For PM2.5, and Electrical Grid Stability, we report the percentiles measured at steps 60 and 120. For HPOBench XGB, Oil, and Asteroid,  we report the percentiles measured at steps 30 and 60.
  All the values displayed here are scaled by 100 for more compact notations.
  The best of each percentile under each type of functions is highlighted in \underline{\bf{bold and underline}}.}
  \vspace{2mm}
  \label{tab:exp_value:Real}
  \begin{tabular}{lllllllll}

\midrule
AF &  P  &  Electrical  & PM2.5 & HPOBench  & Oil & Asteroid \\
 & & Grid Stability & & XGB &  &  &  \\
\midrule
FSAF & 25\%   &  \underline{\bf{0.00}}, \underline{\bf{0.00}} & \underline{\bf{0.00}}, \underline{\bf{0.00}} & \underline{\bf{0.00}}, \underline{\bf{0.00}} & \underline{\bf{0.00}}, \underline{\bf{0.00}} & \underline{\bf{0.00}}, \underline{\bf{0.00}}  \\ 
 & 50\%   &  \underline{\bf{5.28}}, 0.60 & \underline{\bf{11.00}}, \underline{\bf{6.00}} & \underline{\bf{0.14}}, \underline{\bf{0.00}} & \underline{\bf{16.59}}, \underline{\bf{0.00}} & 2.85, \underline{\bf{0.00}}  \\ 
 & 75\%   &  12.16, 7.00 & \underline{\bf{19.00}}, \underline{\bf{15.00}} & 3.16, 2.03 & 40.82, 16.59 & 11.16, \underline{\bf{0.00}}  \\ 
 & 90\%   &  16.76, \underline{\bf{12.20}} & 37.00, \underline{\bf{27.60}} & 8.42, 6.29 & 50.89, \underline{\bf{22.89}} & 18.90, \underline{\bf{0.00}}  \\ 
\midrule
MetaBO & 25\%   &  2.42, \underline{\bf{0.00}} & 20.00, 9.00 & \underline{\bf{0.00}}, \underline{\bf{0.00}} & \underline{\bf{0.00}}, \underline{\bf{0.00}} & 16.66, \underline{\bf{0.00}}  \\ 
 & 50\%   &  7.43, 2.70 & 26.00, 17.00 & 1.38, 0.28 & 20.28, 1.96 & 40.58, 10.54  \\ 
 & 75\%   &  11.15, \underline{\bf{6.86}} & 39.00, 26.00 & \underline{\bf{3.15}}, \underline{\bf{1.51}} & 43.34, 25.69 & 50.32, 19.76  \\ 
 & 90\%   &  17.59, 13.61 & 44.40, 35.00 & \underline{\bf{6.62}}, 5.79 & 60.52, 37.25 & 58.37, 35.71  \\ 
\midrule
MetaBO-T & 25\%   &  2.81, \underline{\bf{0.00}} & 23.00, 12.00 & \underline{\bf{0.00}}, \underline{\bf{0.00}} & 15.96, \underline{\bf{0.00}} & \underline{\bf{0.00}}, \underline{\bf{0.00}}  \\ 
 & 50\%   &  6.86, 3.74 & 30.00, 16.00 & 1.38, 0.61 & 31.71, \underline{\bf{0.00}} & 2.58, \underline{\bf{0.00}}  \\ 
 & 75\%   &  12.17, 9.72 & 39.00, 28.00 & 3.46, 2.47 & 49.05, 15.75 & 22.93, \underline{\bf{0.00}}  \\ 
 & 90\%   &  14.76, 13.89 & 43.60, 36.00 & 8.27, 6.62 & 63.61, 31.67 & 32.49, \underline{\bf{0.00}}  \\ 
\midrule
EI & 25\%   &  0.30, \underline{\bf{0.00}} & 12.00, 5.00 & \underline{\bf{0.00}}, \underline{\bf{0.00}} & 15.10, \underline{\bf{0.00}} & \underline{\bf{0.00}}, \underline{\bf{0.00}}  \\ 
 & 50\%   &  5.72, \underline{\bf{0.00}} & 16.00, 15.00 & 0.60, 0.04 & 25.60, \underline{\bf{0.00}} & \underline{\bf{0.00}}, \underline{\bf{0.00}}  \\ 
 & 75\%   &  \underline{\bf{9.84}}, 7.12 & 29.00, 21.00 & 3.16, 2.70 & 43.43, \underline{\bf{15.63}} & 8.73, \underline{\bf{0.00}}  \\ 
 & 90\%   &  \underline{\bf{14.66}}, 12.72 & \underline{\bf{36.00}}, 33.60 & 8.03, 8.03 & 63.95, 25.83 & 21.64, \underline{\bf{0.00}}  \\ 
\midrule
PI & 25\%   &  \underline{\bf{0.00}}, \underline{\bf{0.00}} & 6.00, \underline{\bf{0.00}} & \underline{\bf{0.00}}, \underline{\bf{0.00}} & 0.27, \underline{\bf{0.00}} & \underline{\bf{0.00}}, \underline{\bf{0.00}}  \\ 
 & 50\%   &  5.72, 1.30 & 17.00, 8.00 & 0.60, 0.03 & 18.76, 1.66 & 2.85, \underline{\bf{0.00}}  \\ 
 & 75\%   &  \underline{\bf{9.84}}, 7.30 & 33.00, \underline{\bf{15.00}} & 3.16, 2.14 & 29.18, 15.75 & 17.60, \underline{\bf{0.00}}  \\ 
 & 90\%   &  15.31, 12.48 & 37.40, 32.20 & 8.03, \underline{\bf{4.92}} & 53.95, 27.49 & 23.13, \underline{\bf{0.00}}  \\ 
\midrule
GP-UCB & 25\%   &  1.49, \underline{\bf{0.00}} & 11.00, 8.00 & \underline{\bf{0.00}}, \underline{\bf{0.00}} & 14.11, \underline{\bf{0.00}} & \underline{\bf{0.00}}, \underline{\bf{0.00}}  \\ 
 & 50\%   &  8.05, 2.87 & 18.00, 15.00 & 0.60, 0.04 & 21.54, 13.27 & \underline{\bf{0.00}}, \underline{\bf{0.00}}  \\ 
 & 75\%   &  14.55, 9.84 & 28.00, 20.00 & 3.16, 2.53 & 30.98, 21.83 & \underline{\bf{4.84}}, \underline{\bf{0.00}}  \\ 
 & 90\%   &  17.93, 12.83 & 38.20, 30.40 & 8.03, \underline{\bf{4.92}} & 49.56, 30.39 & 20.66, 1.97  \\ 
\midrule
MES & 25\%   &  1.49, \underline{\bf{0.00}} & 11.00, 8.00 & \underline{\bf{0.00}}, \underline{\bf{0.00}} & 1.95, \underline{\bf{0.00}} & \underline{\bf{0.00}}, \underline{\bf{0.00}}  \\ 
 & 50\%   &  6.81, 2.89 & 15.00, 14.00 & 0.61, 0.04 & 20.28, 1.95 & 1.36, \underline{\bf{0.00}}  \\ 
 & 75\%   &  13.61, 9.17 & 32.00, 16.00 & 3.16, 2.73 & \underline{\bf{28.83}}, 20.93 & 7.62, \underline{\bf{0.00}}  \\ 
 & 90\%   &  16.83, 13.22 & 37.00, 37.00 & 8.03, 6.29 & \underline{\bf{33.85}}, 30.42 & \underline{\bf{14.82}}, 3.12  \\

 \end{tabular}
 \label{table:percentiles of real datasets}
\end{table*}